\documentclass[9.5pt,journal,compsoc]{IEEEtran}

\usepackage[pagebackref=true,hidelinks,breaklinks=true,bookmarks=false,linkcolor={black},citecolor={black},urlcolor={black}]{hyperref}

\usepackage[nocompress]{cite}

\usepackage{amsmath,amsfonts}
\usepackage{array}
\usepackage{textcomp}
\usepackage{stfloats}
\usepackage{url}
\usepackage{verbatim}
\usepackage{graphicx}
\usepackage{cite}
\usepackage{xspace}

\usepackage[table,dvipsnames]{xcolor}
\usepackage{enumitem}
\usepackage{amssymb}
\usepackage{amsthm}
\usepackage{mathrsfs}
\usepackage{bbding}
\usepackage{bbm}
\renewcommand{\paragraph}[1]{\vspace{2mm}\noindent \textbf{#1}}

\usepackage{subcaption}
\usepackage{nicefrac}
\usepackage{multirow}
\usepackage{booktabs}
\usepackage{soul}
\usepackage{color}
\usepackage[percent]{overpic}
\usepackage{ragged2e}
\usepackage[ruled,vlined]{algorithm2e}

\newcommand{\InputSet}{\mathcal{I}}
\newcommand{\Image}[1]{\mathbf{I}_{#1}} 
\newcommand{\Kmat}[1]{K_{#1}}           
\newcommand{\Pose}[1]{\xi_{#1}}         
\newcommand{\Depth}[1]{\mathbf{D}_{#1}} 
\newcommand{\Normal}[1]{\mathbf{N}_{#1}}
\newcommand{\LineSet}[1]{\mathbf{L}_{#1}} 
\newcommand{\LineDet}[1]{\mathbf{l}^{\rm det}_{#1}} 
\newcommand{\LineDetPt}[2]{\mathbf{x}^{#2}_{#1}} 

\newcommand{\LineMap}{\mathbb{L}} 
\newcommand{\LineMapSeg}[1]{\mathbf{l}^{3d}_{#1}} 
\newcommand{\LineThreeDPt}[2]{\mathbf{#1}^{3d}_{#2}} 

\newcommand{\PlaneEdges}[1]{\mathbf{E}_{#1}}

\newcommand{\methodname}{LiP-Map}
\newcommand{\method}{\textit{\methodname}\xspace}

\newif\ifshowrevision
\showrevisionfalse

\ifshowrevision
  \newcommand{\rev}[1]{\textcolor{blue}{#1}}
\else
  \newcommand{\rev}[1]{#1}
\fi

\ifCLASSOPTIONcompsoc
  \usepackage[nocompress]{cite}
\else
  \usepackage{cite}
\fi

\begin{document}

\title{Interacted Planes Reveal 3D Line Mapping}

\author{Zeran~Ke$^{*}$,
        Bin~Tan$^{*}$,
        Gui-Song Xia,
        Yujun Shen,
        Nan~Xue$^{\dag}$
\thanks{($^{*}$: equal contribution. $^{\dag}$: corresponding author)}
\IEEEcompsocitemizethanks{
\IEEEcompsocthanksitem Z.-R. Ke is with the School of Computer Science, Wuhan University, Wuhan 430072, China. He is currently a research intern at Ant Group. \protect
\IEEEcompsocthanksitem Gui-Song Xia is with the School of Computer Science and the School of Artificial Intelligence, Wuhan University, Wuhan 430072, China. \protect
\IEEEcompsocthanksitem B. Tan, Y. Shen, and N. Xue are with Ant Group, Hangzhou 310000, China. \protect
}
}

\markboth{IEEE Trans. on Pattern Analysis and Machine Intelligence}{Manuscript}

\IEEEtitleabstractindextext{%
\justifying

\begin{abstract}
3D line mapping from multi-view RGB images provides a compact and structured visual representation of scenes.
We study the problem from a physical and topological perspective: a 3D line most naturally emerges as the edge of a finite 3D planar patch.
We present LiP-Map, a line–plane joint optimization framework that explicitly models learnable line and planar primitives. This coupling enables accurate and detailed 3D line mapping while maintaining strong efficiency (typically completing a reconstruction in 3 to 5 minutes per scene).
LiP-Map pioneers the integration of planar topology into 3D line mapping, not by imposing pairwise coplanarity constraints but by explicitly constructing interactions between plane and line primitives, thus offering a principled route toward structured reconstruction in man-made environments.
On more than 100 scenes from ScanNetV2, ScanNet++, Hypersim, 7Scenes, and Tanks\&Temple, LiP-Map improves both accuracy and completeness over state-of-the-art methods. Beyond line mapping quality, LiP-Map significantly advances line-assisted visual localization, establishing strong performance on 7Scenes.
%
\rev{Our code is released at \url{https://github.com/calmke/LiPMAP} for reproducible research.}
\end{abstract}

\begin{IEEEkeywords}
3D Line Mapping, 3D Planar Primitives, Interacted Planes, Structured Visual Geometry, Camera Relocalization
\end{IEEEkeywords}
}

\maketitle

\section{Introduction}\label{sec:intro}

Structured 3D reconstruction from multi-view images using simple primitives~\cite{limap, LiPYLPYP24, Line3D++, NEAT-TXDX0S24, planeAE-YuZLZG19, PlaneTR-Tan0B0X21, HoW3D-MaTXWZX22, PlanarSplatting2024} is compelling, as it parallels how humans intuitively build complex 3D geometries from fundamental elements such as points, lines, curves, and planes, leveraging the parsimony of geometric composition.
The symbolic nature of structured 3D scene representations could arguably enhance spatial intelligence, enabling a more efficient understanding of our 3D world. Yet this perspective is controversial, since many in the community remain unconvinced, citing the limited reconstruction quality of current methods.
In this paper, we focus on two key primitives, {\em rectangular planes and line segments}, with the goal of achieving accurate, complete, and detailed {\bf 3D line mapping}\footnote{This terminology is borrowed from LIMAP~\cite{limap}, and this task is also referred to as 3D line reconstruction.} for man-made environments.

The problem of 3D line mapping has been extensively studied through two-view matching~\cite{lbd, gluestick} and multi-view tracking of 2D line segments, using a pipeline analogous to 3D point mapping.
However, due to the fundamental differences between line segments and points, establishing reliable line correspondences remains challenging.
A recent approach, LIMAP~\cite{limap}, attempts to mitigate this issue by considering the top $K$ line matches instead of a single match, thereby improving the chance of correct correspondences while pruning incorrect ones.
Although LIMAP~\cite{limap} significantly outperforms Line3D++\cite{Line3D++}, it is limited by a strong dependence on dense viewpoints to achieve successful line triangulation.
Other studies~\cite{LiPYLPYP24,NEAT-TXDX0S24} aim to bypass 2D line matching by leveraging neural implicit fields, demonstrating the potential for matching-free formulations. However, they suffer from the slow optimization process due to their reliance on neural fields.

\begin{figure*}[!t]
    \centering
    \includegraphics[width=\linewidth]{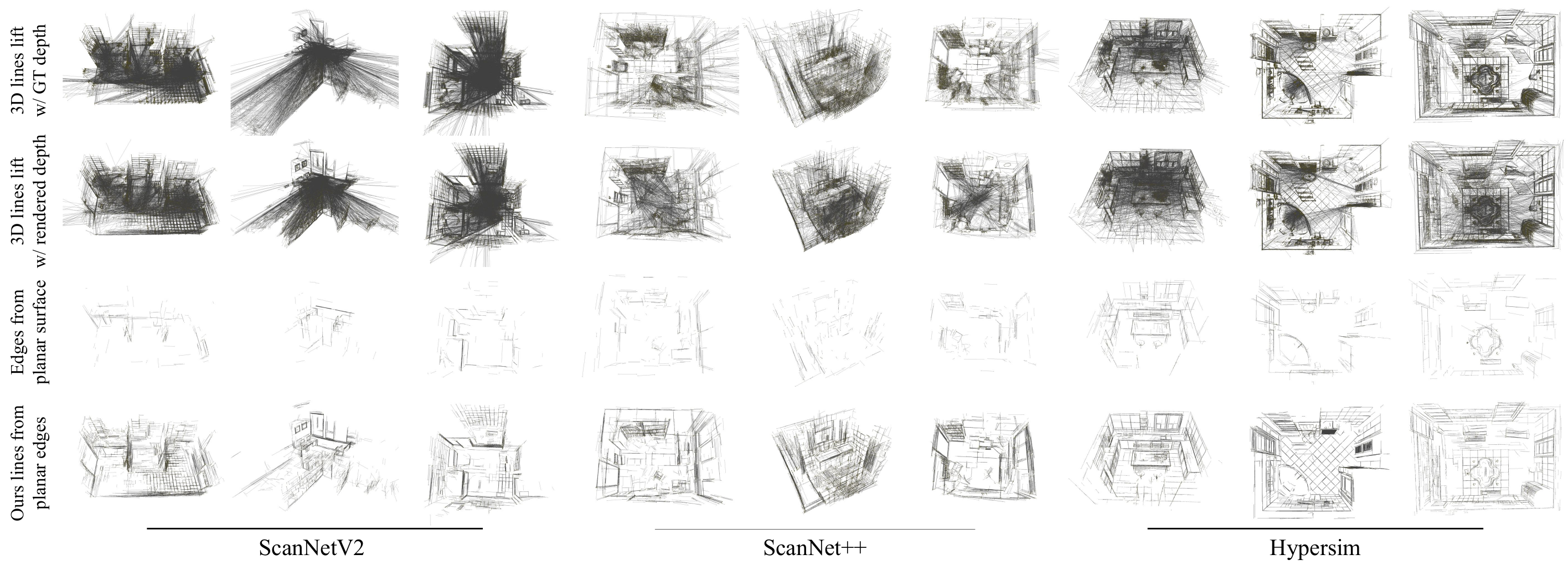}
    \caption{
    Comparison of different 3D line maps on scenes of three public datasets. First row: lift 2D detected lines into 3D lines using depths rendered from the GT mesh. Second row: lift 2D detected lines into 3D lines using depths rendered by PlanarSplatting~\cite{PlanarSplatting2024}. Third row: extract planar edges from PlanarSplatting~\cite{PlanarSplatting2024}. Last row: our 3D line maps.
    2D line segments are detected by DeepLSD~\cite{DeepLSD}.
    \rev{``lift'' means back-project 2D detected lines into 3D using the sensor/predicted depth maps.}
    }
    \label{fig:3d_line_maps}
\end{figure*}

To construct a reliable 3D line map from 2D line detections across multiple views, two key ingredients are required: robust cross-view correspondences and accurate spatial placement of the reconstructed 3D lines.
A common strategy is to utilize an existing surface reconstruction from multi-view inputs with known camera poses and to project detected 2D lines onto the surface using depth maps. However, this approach often fails in practice, as illustrated in Fig.~\ref{fig:3d_line_maps}.
On one hand, view-dependent parallax can cause a single image line to correspond to multiple 3D line segments at different depths. On the other hand, inconsistencies in the depth maps can introduce spurious structural lines.
Because true 3D lines are inherently inseparable from structural surface edges, joint recovery of surfaces and edges is indispensable, motivating a novel formulation of 3D line mapping that explicitly models the geometric relationship and interaction between planes and line segments.

{Planar representations offer a promising opportunity.}
PlanarSplatting~\cite{PlanarSplatting2024} reconstructs structured scene surfaces by optimizing a set of learnable planar primitives.
This formulation is attractive because plane boundaries are naturally formed by lines, revealing a direct geometric and topological link between surfaces and 3D line structures.
However, PlanarSplatting is designed for surface reconstruction only, without explicitly connecting multi-view 2D line detections to the learned primitives.
Thus, we explore how to associate 2D line segments with 3D planar edges and leverage these associations to reconstruct reliable, consistent 3D line maps.
We therefore propose LiP-Map, which explicitly exploits the interaction between planes and lines to recover reliable and consistent 3D line maps. The name stands for Line–Plane Joint Optimization for 3D line mapping.

\begin{figure*}[!t]
    \centering
    \includegraphics[width=\linewidth]{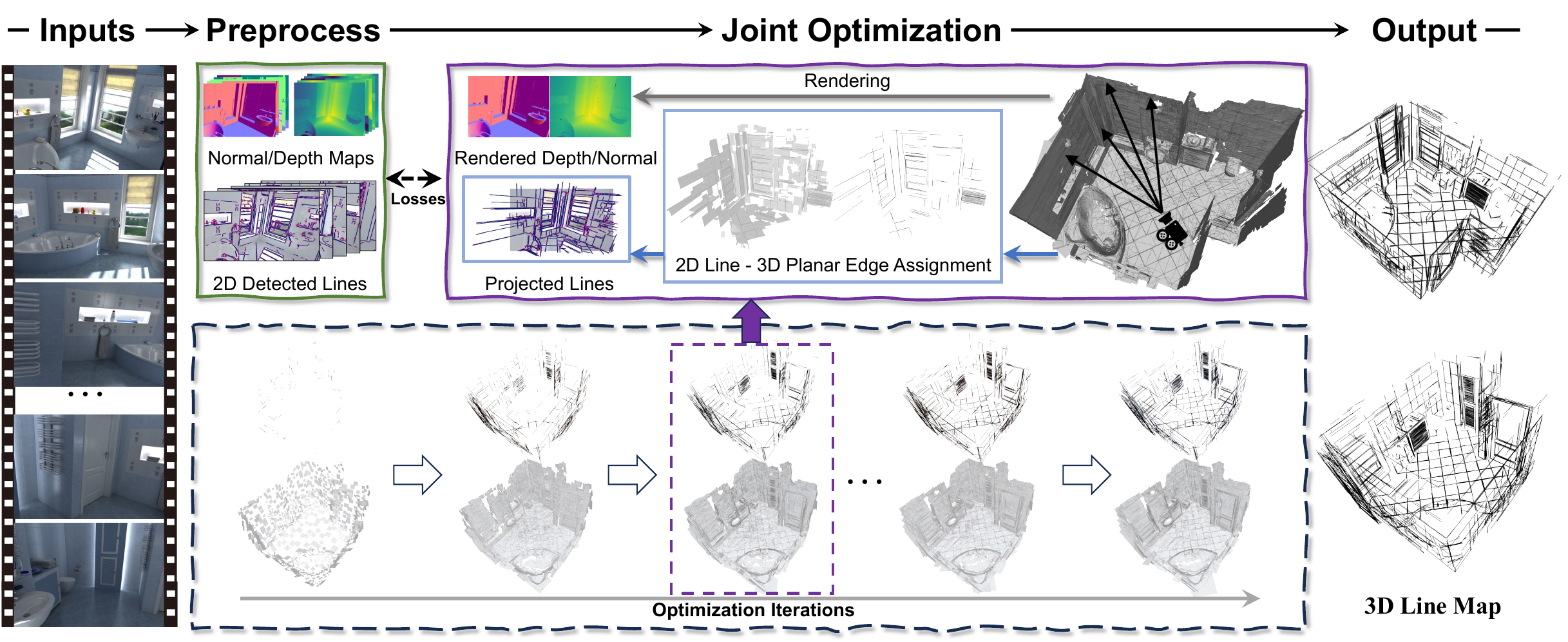}
    \caption{
    Overview of our proposed reconstruction pipeline, a novel 3D line mapping method by exploring the structural synergies between 3D planes and lines.
    }
    \label{fig:teaser}
\end{figure*}

Our approach leverages the geometric synergy between line segments and the edges of 3D planar primitives. By applying any existing 2D line segment detector~\cite{DeepLSD,LSD,HAWPv3,ScaleLSD} to multi-view input images, a set of 3D planes is optimized with two key learning objectives:
\begin{itemize}
\item The 3D planes should align as closely as possible with the 2.5D depth/normal maps from the input images.
\item The edges (or boundaries) of the 3D planes should be consistent with the observed 2D line segments.
\end{itemize}
As illustrated in Fig.~\ref{fig:teaser}, given a set of posed multi-view images, our \method can reconstruct the planar surface of the scene by optimizing a set of learnable 3D planar primitives, with the supervision of depth and normal maps.
Meanwhile, the geometric structure of the scene's planar surface model is progressively refined under the supervision of the multi-view detected 2D line segments.

By formulating a joint optimization that enforces the two above objectives, LiP-Map achieves state-of-the-art accuracy and completeness in 3D line mapping. Extensive experiments on public benchmarks, including 50 scenes from the ScanNetV2~\cite{scannet-DaiCSHFN17} dataset, 30 scenes from the ScanNet++~\cite{scannetpp-YeshwanthLND23} dataset, and 10 scenes from the Hypersim~\cite{Hypersim} dataset, demonstrate its advantages. Furthermore, qualitative comparisons on 7Scenes~\cite{7scene} and Tanks\&Temples~\cite{tanks_temples} using camera poses from VGGT~\cite{wang2025vggt} highlight the robustness of LiP-Map for pose-free inputs. Beyond reconstruction, our 3D line maps provide practical gains in line-assisted visual localization: the lifted 3D primitives offer stable, geometrically meaningful correspondences that complement point features and remain robust in textureless or repetitive regions. On the 7Scenes~\cite{7scene} dataset, integrating them into a standard point-only localization pipeline yields substantially improved pose accuracy and robustness over both point-only baselines and existing point–line joint methods. In addition, LiP-Map also enhances the quality of planar 3D reconstruction, further underscoring its versatility.

\section{Related Work}\label{sec:related}
\paragraph{3D Line Mapping.}
Mapping 2D features from multi-view images to 3D is a fundamental task with numerous applications in 3D vision, augmented reality, and robotics~\cite{COLMAP-SchonbergerF16, NinjaPlane, DUSt3R, USEEK}. However, in indoor scenes, where structures dominate and surfaces often lack texture, traditional 2D keypoint-based mapping methods~\cite{SIFT, ASIFT, SuperGlue, SURF} tend to be suboptimal. Consequently, 3D line mapping has emerged as a compelling alternative to conventional 3D point mapping. Recent advances in 2D line segment detection~\cite{HAWP, HAWPv3, DeepLSD, SOLD2, TP-LSD, ScaleLSD} and matching~\cite{gluestick, LiC0WKLL21} have enabled the adaptation of 3D point mapping pipelines to 3D line mapping~\cite{Line3D++, limap, elsr}. However, because of the fundamental differences between 2D line features and point features, the matching process often introduces inherent uncertainties, sometimes leading to failures. To address this, several studies~\cite{Line3D++, limap, Bai_2024_CLMAP} leverage geometric constraints in the search for line correspondence to improve reconstruction. More recently, neural field-based approaches~\cite{LiPYLPYP24, NEAT-TXDX0S24} have sought to eliminate explicit line matching, achieving improved mapping results.

\paragraph{Structured 3D Surface Reconstruction.}
In addition to 3D line mapping, many studies explore 3D reconstruction using different types of primitives, such as surfels~\cite{Surfels, DSS-WangSWOS19, TwoDGS-HuangYC0G24}, planes~\cite{PlanarSplatting2024, NOPE-SAC-TanXWX23, PlaneTR-Tan0B0X21, planercnn-0012KGFK19, planeAE-YuZLZG19, PlaneRecTR-ShiZ023, Sparseplanes-Jin0OF21, PlaneFormers-AgarwalaJRF22, PlaneMVS}, polygonal meshes~\cite{dmtet, BSP-Net, NeuraldualCont, DBW-abs-2307-05473}, \rev{sketches~\cite{SketchSampler}}, and implicit surfaces~\cite{volsdf, monosdf, manhattanSDF}, among others. Among these, planar 3D reconstruction via PlanarSplatting~\cite{PlanarSplatting2024} shows great potential for indoor scenes, as it explicitly optimizes surface geometry with a set of 3D planes derived from 2.5D cues at ultrafast speed. Explicit modeling and optimization of planar surface geometry bring several benefits over other methods: (1) the expected geometry is always accessible in a differentiable manner without the need for proxies, and (2) planar surfaces have deterministic geometric meaning and approximate scene geometry with linear structures.

Our study falls into the category of matching-free solutions, and we adopt 3D planar splatting~\cite{PlanarSplatting2024} for its computational efficiency and the intrinsic relationship between plane and line representations. To the best of our knowledge, this is the first work to explicitly optimize the synergy between 3D line segments and planes, offering a novel perspective on structured 3D reconstruction.

\section{\method: Line-Plane Joint Mapping}\label{sec:method}
Given a set of posed multi-view images with monocular 2D and 2.5D sketches,
\begin{equation}
\InputSet = \{ (\Image{i}, \Kmat{i}, \Pose{i}, \Depth{i}, \Normal{i}, \LineSet{i}) \}_{i=1}^N
\end{equation}
where $\Image{i}$ is the $i$-th image, $\Kmat{i}$ and $\Pose{i}$ are the intrinsic and extrinsic matrices, $\Depth{i}$ and $\Normal{i}$ are the depth and normal maps (either provided as ground-truth or predicted by pretrained 3D foundation models~\cite{omnidata-EftekharSMZ21, Metric3Dv2}), and $\LineSet{i}$ is the set of 2D line segments detected by
any line segment detectors~\cite{LSD,DeepLSD,HAWPv3,ScaleLSD}, denoted by
\begin{equation}
    \LineSet{i} = \left\{
    \LineDet{i,j} = (\LineDetPt{i,j}{1},\LineDetPt{i,j}{2}) \in \mathbb{R}^{2}\times \mathbb{R}^2
    \right \}.
\end{equation}
Our goal is to reconstruct the 3D line map of the corresponding scene using these 2D/2.5D geometric cues. The final line mapping result is denoted by
\begin{equation}
\LineMap = \left\{
\LineMapSeg{i} = (\LineThreeDPt{u}{i},\LineThreeDPt{v}{i}) \in \mathbb{R}^3 \times \mathbb{R}^3 \right\}.
\end{equation}

\begin{figure}
  \centering
  \includegraphics[width=0.8\linewidth]{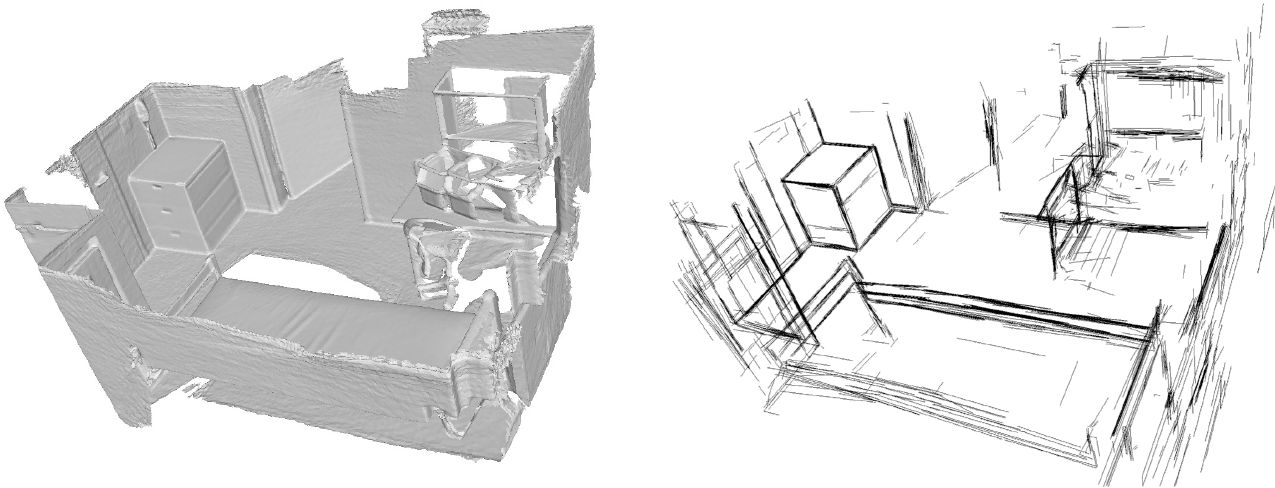}
  \caption{Illustration of surfaces (left) and 3D lines (right) in a ScanNetV2 scene~\cite{scannet-DaiCSHFN17}. The 3D lines align closely with physical surface boundaries, highlighting the strong correlation between the two structures.
  }
  \label{fig:illu_surface_line}
\end{figure}

\subsection{Representation of Planar Surface}
\label{subsec: plane definition}
We follow the recent PlanarSplatting~\cite{PlanarSplatting2024}, which reconstructs a scene using a set of 3D rectangular planes $\mathbf{\Pi} = \{ \pi_{i} \}_{i=1}^{K}$ from $\mathcal{I}$. This is motivated by the observation (see Fig.~\ref{fig:illu_surface_line}) that 3D rectangular planes naturally link surfaces and 3D lines: the interiors of the planes approximate scene surfaces, while their edges encompass the target 3D lines.

\begin{figure}
  \centering
  \includegraphics[width=0.5\linewidth]{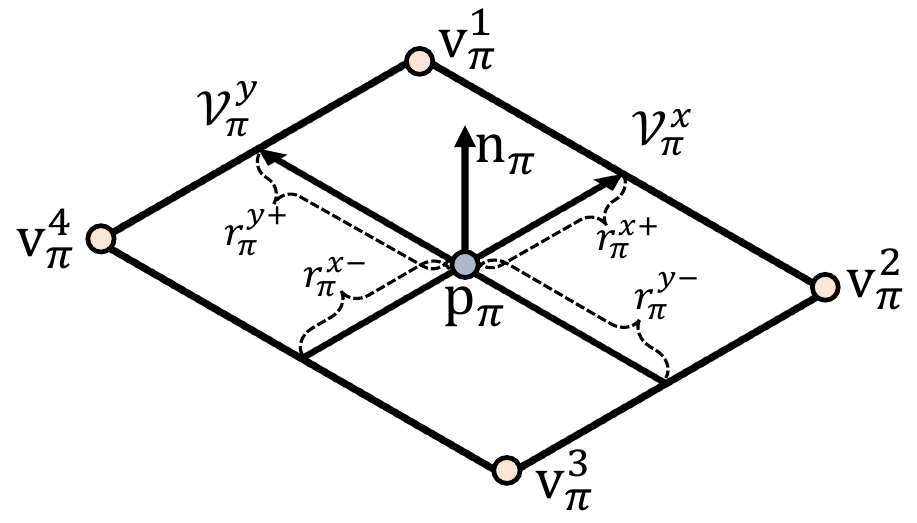}
  \caption{Representation of the 3D rectangular plane with learnable shape parameters.}
  \label{fig:plane_rep}
\end{figure}

\paragraph{3D Plane Parameterization.}
As shown in the Fig.~\ref{fig:plane_rep} and described in PlanarSplatting~\cite{PlanarSplatting2024}, each 3D rectangular plane $\pi$ is equipped with learnable parameters, including the plane center $\mathbf{p}_{\pi} \in \mathbb{R}^3$, the plane rotation $\mathbf{q}_{\pi} \in \mathbb{R}^4$ (in quaternion), and the plane radii $\mathbf{r}_{\pi} = \{ r_\pi^{x+},r_\pi^{x-},r_\pi^{y+},r_\pi^{y-} \} \in \mathbb{R}^4 $.
Here, the symbol ``$+/-$" represents the ``positive/negative" direction of the X-axis/Y-axis of the rectangle.
Then, the positive direction of the X-axis $\mathcal{V}^{x}_{\pi} \in \mathbb{R}^3$ and Y-axis $\mathcal{V}^{y}_{\pi} \in \mathbb{R}^3$ of the 3D rectangular plane can be further calculated as:
\begin{equation}
    \mathcal{V}^{x}_{\pi} = \mathbf{R}(\mathbf{q}_{\pi}){[1,0,0]}^{\top}, \quad \mathcal{V}^{y}_{\pi} = \mathbf{R}(\mathbf{q}_{\pi}){[0,1,0]}^{\top},
\end{equation}
where $\mathbf{R}(\mathbf{q}_{\pi}) \in \mathbb{R}^{3 \times 3}$ is the rotation matrix converted from $\mathbf{q}_{\pi}$. The normal of the 3D rectangular plane $\mathbf{n}_{\pi} \in \mathbb{R}^3$ can be calculated as:
\begin{equation}
    \mathbf{n}_{\pi} = \mathbf{R}(\mathbf{q}_{\pi}){[0,0,1]}^{\top}.
\end{equation}
These learnable 3D planes $\mathbf{\Pi}$ are optimized with 2.5D geometric cues (\emph{i.e.}, monocular depth and normal maps) to reconstruct the scene with the differentiable rasterization. Please refer to~\cite{PlanarSplatting2024} or our Appendix~A for more details.

\paragraph{Plane Edges as Lines.}
Based on the above representation, we further sequentially define the set of four planar vertices $\mathbf{V}_{\pi}=\{ \mathbf{v}_{\pi}^i \in \mathbb{R}^{3}\}_{i=1}^{4}$ of the 3D plane $\pi$ as:
\begin{equation}
\begin{aligned}
    \mathbf{v}_{\pi}^1=&\mathbf{p}_{\pi} + r_{\pi}^{x+}\mathcal{V}_{\pi}^{x} + r_{\pi}^{y+}\mathcal{V}_{\pi}^{y}, \\
    \mathbf{v}_{\pi}^2=&\mathbf{p}_{\pi} + r_{\pi}^{x+}\mathcal{V}_{\pi}^{x} - r_{\pi}^{y-}\mathcal{V}_{\pi}^{y}, \\
    \mathbf{v}_{\pi}^3=&\mathbf{p}_{\pi} -r_{\pi}^{x-}\mathcal{V}_{\pi}^{x} - r_{\pi}^{y-}\mathcal{V}_{\pi}^{y}, \\
    \mathbf{v}_{\pi}^4=&\mathbf{p}_{\pi} - r_{\pi}^{x-}\mathcal{V}_{\pi}^{x} + r_{\pi}^{y+}\mathcal{V}_{\pi}^{y}.
\end{aligned}
\end{equation}
Then, the set of four planar edges $\mathbf{E}_{\pi}=\{ \mathbf{e}_{\pi}^{i}\}_{i=1}^{4}$ of the 3D plane $\pi$ can be represented as:
\begin{equation}
\begin{aligned}
    {e}_{\pi}^{1} =& (\mathbf{v}_{\pi}^{1},\mathbf{v}_{\pi}^{2}),~\quad {e}_{\pi}^{2} = (\mathbf{v}_{\pi}^{2},\mathbf{v}_{\pi}^{3}),~\quad \\
    {e}_{\pi}^{3} =& (\mathbf{v}_{\pi}^{3},\mathbf{v}_{\pi}^{4}),~\quad {e}_{\pi}^{4} = (\mathbf{v}_{\pi}^{4},\mathbf{v}_{\pi}^{1}).
\end{aligned}
\end{equation}
For a scene represented with $K$ 3D planes $\mathbf{\Pi}=\{ \pi_{i}\}_{i=1}^{K}$, the target 3D lines $\mathbb{L}$ is the subset of the edges of all 3D planes:
\begin{equation}
    \mathbb{L} \subseteq \{ \mathbf{E}_{\pi_i}\}_{i=1}^{K}.
\end{equation}
This simplifies 3D line mapping into the task of optimizing plane edges aligned with 2D line detections, jointly with the optimization of plane surfaces.

\subsection{2D Line to 3D Plane Edge Assignment}
\label{subsec:3d line proposals}
\begin{figure}
    \centering
    \includegraphics[width=0.8\linewidth]{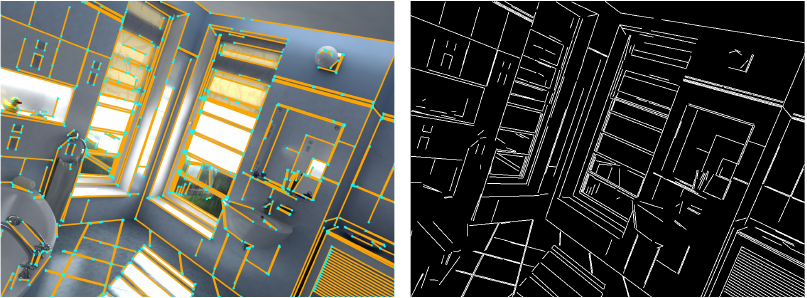}
    \caption{
    Illustration of 2D line segments detected by DeepLSD~\cite{DeepLSD} and their corresponding 1-pixel regions in a single view of the Hypersim~\cite{Hypersim} scene ``ai\_001\_001''. Left: the detected 2D lines. Right: the corresponding 1-pixel regions. }
    \label{fig:line_region}
\end{figure}

To reconstruct 3D lines $\mathbb{L}$ within the plane edges $\{ \mathbf{E}_{\pi_i}\}_{i=1}^{K}$, we have to find the plane edges that potentially correspond to 3D lines. To avoid ambiguity, we refer to these kinds of edges as {``line-edge"}, denoted as $\mathcal{E}=\{ {e}_{\pi_i}^{j}\}$, where $i\in \{1,2,...,K \}$, $j \in \{ 1,2,3,4\}$, and $\mathcal{E} \subseteq \{ \mathbf{E}_{\pi_i}\}_{i=1}^{K}$. We apply the 2D line segments $\mathbf{L}_i$ from arbitrary line detectors to find ``line-edge''. Because all planes are under optimization, directly associating any plane edges $\PlaneEdges{\pi_i}$ with 2D observations $\LineDet{i,j} \in \LineSet{i}$ would be fragile. We tackle this issue from the attraction field representations~\cite{HAWPv3} of the 2D line segments $\LineSet{i}$. Denote $\mathcal{R}_{i,j}$ as the 1-pixel region of the line segment $\LineDet{i,j}$, in which the distance of any pixel $\mathbf{p} \in \mathcal{R}_{i,j}$ to $\LineDet{i,j}$ is not greater than 1 pixel, and we pixel-wise associate the most possible planar primitive with respect to the pixel $\mathbf{p}$ and its supported line segment $\LineDet{i,j}$. Fig.~\ref{fig:line_region} shows a case of detected line segments using DeepLSD~\cite{DeepLSD} and all 1 pixel regions of the detection results.

Taking a pixel $\mathbf{p}$ from the 1-pixel region of $\LineDet{i,j}$, we cast its corresponding ray which starts from the camera center of image $\Image{i}$, and find the first hit 3D plane $\pi$ pixel-wise.
Next, to make sure which 3D edge of the 3D plane $\pi$ is the target ``line-edge'', we project all four 3D edges $\mathbf{E}_{\pi}=\{ e_{\pi}^{v} \}_{v=1}^{4}$ to the 2D image. The projected 2D edges can be defined as:
 \begin{equation}
     \mathbf{E}_{\pi}^{2d}=\left\{ e_{\pi,v}^{2d} \right\}_{v=1}^{4}.
\end{equation}

Since the 3D plane is optimized on the fly, the projected 2D edges $\mathbf{E}_{\pi}^{2d}$ may not be well aligned with the detected 2D line segment $\LineDet{i,j}$ before convergence. To address this, we first compare the angular distances between the edges in $\mathbf{E}_{\pi}^{2d}$ and the corresponding detected 2D line, and discard the two edges with the largest angular distances.
We then apply the orthogonal distance between the remaining edges and the corresponding detected 2D line to select the best candidate ${e}_{\pi,v^*}^{2d} \in \mathbf{E}_{\pi}^{2d}$.
With the best candidate ${e}_{\pi,v^*}^{2d}$ regarding to $\LineDet{i,j}$, we define the selected 2D/3D ``line-edge`` at pixel $\mathbf{p}$ as
\begin{equation}
    {e}_{\mathbf{p}\rightarrow\pi}^{2d} := {e}_{\pi,v^*}^{2d} = (\mathbf{v}_{\pi,v^*}^1,\mathbf{v}_{\pi,v^*}^2), \quad
    {e}_{\mathbf{p}\rightarrow\pi} := {e}_{\pi}^{v^*},
\end{equation}
where $(\mathbf{v}_{\pi,v^*}^1,\mathbf{v}_{\pi,v^*}^2)$ are the two vertices of the projected 2D edge ${e}_{\pi,v^*}^{2d}$.

We now obtain an assignment between the 2D line segment and 2D/3D ``line-edge'' from pixel $p$, denoted as $S(p)=(\LineDet{i,j}, {e}_{\mathbf{p}\rightarrow\pi}^{2d}, e_{\mathbf{p}\rightarrow\pi})$. After searching throughout all 1-pixel regions of all detected 2D line segments $\LineSet{i}$ on image $\Image{i}$, we finally obtain the assignments of all 2D lines to 3D plane edges with regard to meaningful pixels.
These assignments are then used to jointly optimize the 3D lines (``line-edge") and 3D planes, which will be described in the next section.

\begin{figure}
  \centering
  \includegraphics[width=1.0\linewidth]{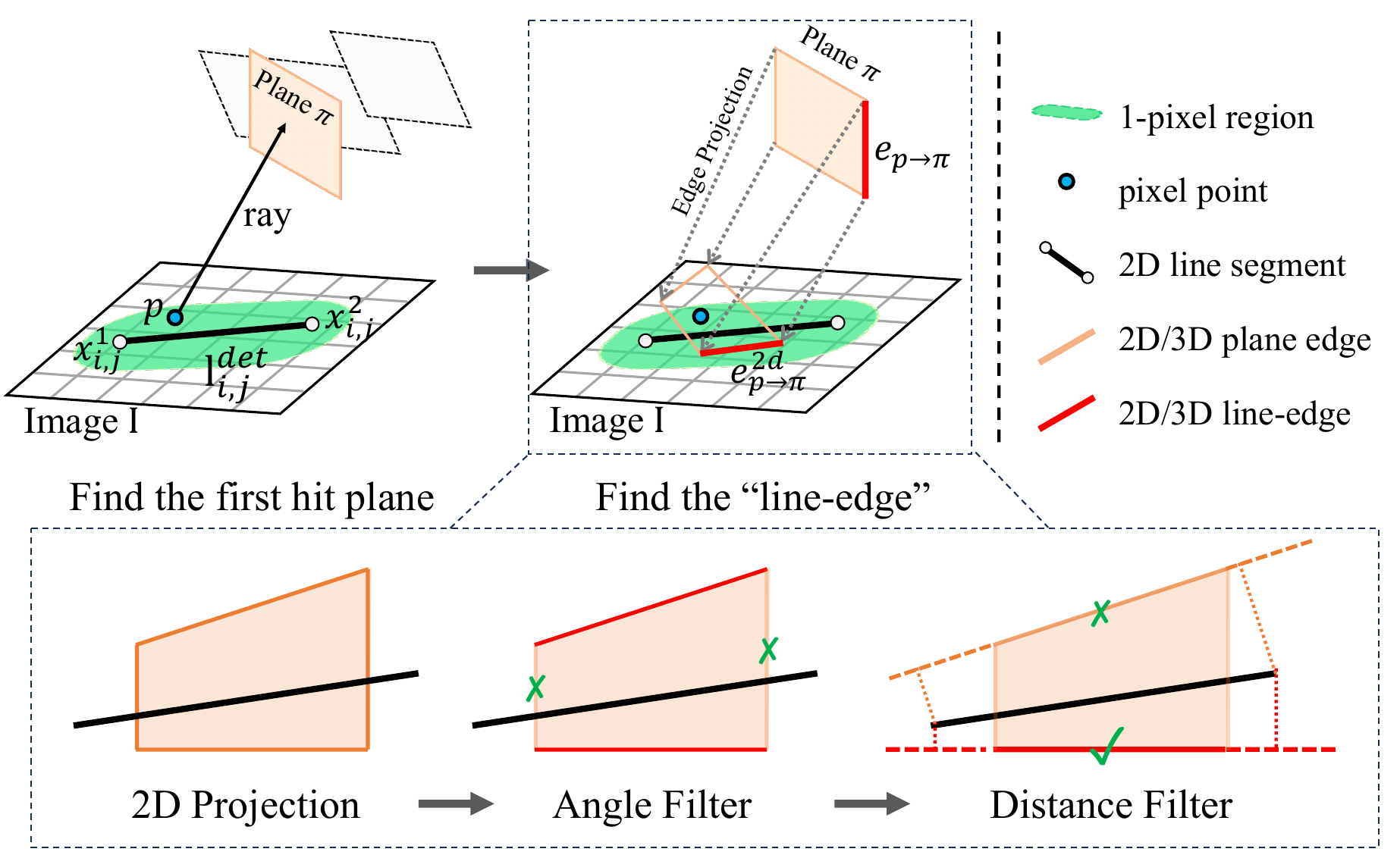}
  \caption{Illustration of 2D line to 3D plane edge assignment. The best candidate of the ``line-edge'' is filtered from the perspective of angle and distance.}
  \label{fig:2d-3d-ass}
\end{figure}

\begin{figure}
    \centering
    \includegraphics[width=\linewidth]{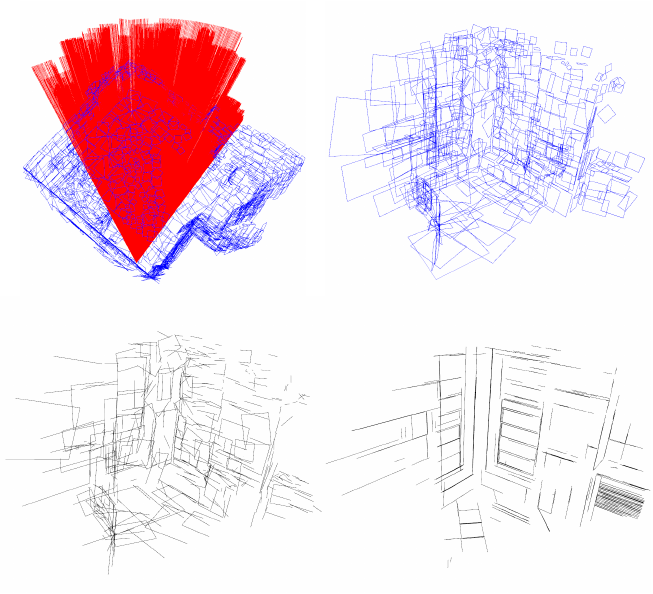}
    \caption{Illustration of the assignment process from one view. \textbf{Red}: rays. \textbf{Blue}: planar primitives. \textbf{Black}: planar edges. \textbf{Top left}: the association of rays and planar primitives. \textbf{Top right}: the selected planar primitives. \textbf{Bottom left}: the selected planar edges. \textbf{Bottom right}: the optimized planar edges.}
    \label{fig:association}
\end{figure}

For clarity, we illustrate this process in Fig.~\ref{fig:association}. As it is shown, rays are cast from all pixels in the 1-pixel regions to find the first intersected planar primitive. To achieve this, we first compute the intersection points between all rays and the planes underlying the planar primitives. We then check whether each intersection lies within the bounds of the primitive and whether its depth value is positive, thereby identifying primitives that are validly intersected by the rays. Among these valid intersections, we select the nearest planar primitive with the minimal depth value along each ray. In this way, we identify the 3D planes that potentially contain 3D line segments. Finally, the selected planar primitives are jointly optimized to reconstruct the corresponding 3D lines as described in Sec.~\ref{subsec:joint optimization}. Their edges align well with the detected 2D lines, enabling for reliable and plausible reconstruction.

\subsection{Plane-Line Optimization}
\label{subsec:joint optimization}
Given an image $\mathbf{I}$ with 2D line to 3D plane edge assignments $\mathcal{S}_{\mathbf{I}}$, our goal is to optimize the assigned 3D ``line-edge'' for line mapping. We achieve this by constraining the ``line-edge'' in both 3D and 2D space.

In 3D space, we have to ensure that these ``line-edge'' are located on the scene surface. Note that the ``line-edge'' themselves are part of the 3D planes. Thus, the on-surface constraint can be directly implemented by optimizing the 3D planes with the plane rendering loss $\mathcal{L}_{\mathbf{\Pi}}$ supervised by 2.5D monocular depths \& normals, formed as:
\begin{equation}
    \mathcal{L}_{\mathbf{\Pi}} = \alpha_{D} \mathcal{L}_{\mathbf{D_{epth}}} + \alpha_{N} \mathcal{L}_{\mathbf{N_{ormal}}},
\end{equation}
where $\alpha_{D}=5$ and $\alpha_{N}=1$. This rendering process has been introduced in PlanarSplatting~\cite{PlanarSplatting2024}, and we also make an introduction in Appendix~A.

In 2D space, we have to ensure that the ``line-edge'' are aligned with the 2D detected line segments on image $\mathbf{I}$. Specifically, given all 2D line to 3D plane edge assignments $\mathcal{S}_{\mathbf{I}}$, the 2D alignment loss can be calculated as:
\begin{equation}
    \mathcal{L}_{\rm 2d} = \mathcal{L}_{\rm euc}^{2d} + \mathcal{L}_{\rm ort}^{2d},
\end{equation}
and the 2D endpoints loss $\mathcal{L}_{\rm euc}^{2d}$ and 2D lines loss $\mathcal{L}_{\rm ort}^{2d}$ are calculated by:
\begin{equation}
    \begin{split}
    \mathcal{L}_{\rm euc}^{2d} &= \sum_{\LineDet{i,j}\in\LineSet{i}} ~\sum_{\mathbf{p} \in \mathcal{R}(\LineDet{i,j})} \mathcal{D}_{\rm euc}(\LineDet{i,j}, {e}^{2d}_{\mathbf{p}\rightarrow\pi}),  \\
   \mathcal{L}_{\rm ort}^{2d} &= \sum_{\LineDet{i,j}\in\LineSet{i}} ~\sum_{\mathbf{p} \in \mathcal{R}(\LineDet{i,j})}
   \mathcal{D}_{\rm ort}(\LineDet{i,j}, {e}^{2d}_{\mathbf{p}\rightarrow\pi}),
    \end{split}
\end{equation}
where
\begin{equation}
\begin{split}
    \mathcal{D}_{\rm euc}(\LineDet{i,j}, {e}^{2d}_{\mathbf{p}\rightarrow\pi})= \min (\|\mathbf{x}_{i,j}^1-\mathbf{v}_{\pi,v^*}^1\| + \|\mathbf{x}_{i,j}^2-\mathbf{v}_{\pi,v^*}^2\|, \\
    \|\mathbf{x}_{i,j}^1-\mathbf{v}_{\pi,v^*}^2\| + \|\mathbf{x}_{i,j}^2-\mathbf{v}_{\pi,v^*}^1\|),
\end{split}
\end{equation}
\begin{equation}
\begin{split}
    \mathcal{D}_{\rm ort}(\LineDet{i,j}, {e}^{2d}_{\mathbf{p}\rightarrow\pi}) = &  \frac{\|(\mathbf{v}_{\pi,v^*}^1-\mathbf{x}_{i,j}^2) \times (\mathbf{x}_{i,j}^1-\mathbf{x}_{i,j}^2)\|}{\|\mathbf{x}_{i,j}^1-\mathbf{x}_{i,j}^2\|}   \\
    + &
    \frac{\|(\mathbf{v}_{\pi,v^*}^2-\mathbf{x}_{i,j}^2) \times (\mathbf{x}_{i,j}^1-\mathbf{x}_{i,j}^2)\|}{\|\mathbf{x}_{i,j}^1-\mathbf{x}_{i,j}^2\|}.
\end{split}
\end{equation}
Here $(\mathbf{x}_{i,j}^1,\mathbf{x}_{i,j}^2)$ are two endpoints of the 2D line segment $\LineDet{i,j}$, and $(\mathbf{v}_{\pi,v^*}^1,\mathbf{v}_{\pi,v^*}^2)$ are two endpoints of the projected ``line-edge'' ${e}^{2d}_{\mathbf{p}\rightarrow\pi}$.

Besides the above two losses for constraining each ``line-edge'' in both 2D and 3D space individually, we also introduce a group loss $\mathcal{L}_{\rm group}$ which encourages all ``line-edge'' assigned to the same 2D line segment to be close enough in the 3D space and can be calculated as:
\begin{equation}
    \mathcal{L}_{\rm group}=\sum_{\forall\LineDet{i,j}\in\LineSet{i}} ~\sum_{\forall \mathbf{p_1},\mathbf{p_2} \in \mathcal{R}(\LineDet{i,j})} \mathcal{D}_{\rm ort}(e_{\mathbf{p_1}\rightarrow\pi}, e_{\mathbf{p_2}\rightarrow\pi}),
    \label{eq:grouping}
\end{equation}
where $p_1, p_2$ are the pixels within the ``1-pixel region'' $\mathcal{R}(\ddot{l})$ of line segment $\ddot{l}$. The final loss can then be calculated as:
\begin{equation}
    \mathcal{L} = \alpha_{\mathbf{\Pi}} \mathcal{L}_{\mathbf{\Pi}} + \alpha_{L} \mathcal{L}_{\mathbf{E}},
\end{equation}
where $\mathcal{L}_{\mathbf{E}}=(\mathcal{L}_{\rm 2d} + \mathcal{L}_{\rm group})$, $\alpha_{\mathbf{\Pi}}=10$, and $\alpha_{L}=0.1$ in our experiments.

\subsection{Line Mapping Finalization} \label{subsec:ltb}
After the optimization process converges, extracting 3D lines is straightforward, since each line corresponds to an edge of an optimized 3D plane. To ensure the accuracy of the mapping results, we filter inliers by computing both the angular distance $d_{\rm ang} = \mathcal{D}_{\rm ang}(\LineDet{i,j}, {e}^{2d}_{\mathbf{p}\rightarrow\pi})$ and the orthogonal distance $d_{\rm orth} = \mathcal{D}_{\rm ort}(\LineDet{i,j}, {e}^{2d}_{\mathbf{p}\rightarrow\pi})$ between the 2D projection ${e}_{\pi,v}^{2d}$ and its associated 2D line detection $\LineDet{i,j}$.
If $d_{\rm ang} \leq \tau_{\alpha}$ and $d_{\rm orth} \leq \tau_{d}$, then the 3D planar edge ${e}_{\pi}^v$ is considered a correct interpretation of the detected 2D line segment $\LineDet{i,j}$ and is retained in the final mapping results.
We set $\tau_d = 1$ pixel and $\tau_{\alpha} = 0.01$ in our experiments.

\paragraph{Line Track Builder.} After the final 3D line map is obtained, we then build the 3D-2D line correspondences between $\LineMap$ and all the 2D detection results over image set $\InputSet$.
Denoted by $\mathcal{T}(\mathbf{l}_i^{3d}) = \{(v_1, \imath_1), \ldots, (v_T, \imath_T)\}$ the tracked 2D line segments of $\mathbf{l}_i^{3d}$, any detected 2D line segment $\LineDet{v_k,\imath_k} \in \LineSet{v_k}$ should satisfy the following criteria between the 2D projection  $\mathbf{l}_i^{v_k}$ of $\mathbf{l}_i^{3d}$ and $\LineDet{v_k,\imath_k}$,
\begin{itemize}
    \item The angular distance $d_{\rm ang} = \mathcal{D}_{\rm ang}(\LineDet{v_k,\imath_k},\mathbf{l}_i^{v_k})$ should be smaller than $\tau_a = 0.01$;
    \item The orthogonal distance $d_{\rm dist}$ between $\LineDet{v_k,\imath_k}$ and $\mathbf{l}_i^{v_k}$ should be smaller than $\tau_d = 2$ pixels;
    \item The overlap ratio $d_{\rm overlap}$ between $\LineDet{v_k,\imath_k}$ and $\mathbf{l}_i^{v_k}$ should be greater than $\tau_o = 0.2$.
\end{itemize}
We provide the diagrams and mathematical formulas for the computations of $d_{\rm ang}$, $d_{\rm dist}$, and $d_{\rm overlap}$ in Appendix~B.

\section{Experiments}\label{sec:exp}

\begin{table*}[!t]
    \centering
    \caption{Quantitative results of 3D line mapping on the ScanNetV2~\cite{scannet-DaiCSHFN17} dataset and the ScanNet++~\cite{scannetpp-YeshwanthLND23} dataset with lines detected by 4 different detectors. All metrics are reported at 5 mm along with the average number of reconstructed lines.}
    \label{tab:scannet1}
    \resizebox{\linewidth}{!}{
    \begin{tabular}{l|l|ccccc|ccccc|c}
    \toprule
    \multicolumn{2}{c}{\textit{ScanNetV2}~\cite{scannet-DaiCSHFN17}: 50 scenes} & \multicolumn{5}{c}{Line-Level Metrics} & \multicolumn{5}{c}{Junction-Level Metrics} & \multicolumn{1}{c}{Statistics} \\
    \cmidrule(lr){1-2} \cmidrule(lr){3-7} \cmidrule(lr){8-12} \cmidrule(lr){13-13}
    Detector & Method & {ACC-L$\downarrow$} & {COMP-L$\downarrow$} & {PREC-L$\uparrow$} & {RECAL-L$\uparrow$} & {F-SCORE-L$\uparrow$} & {ACC-J$\downarrow$} & {COMP-J$\downarrow$} & {PREC-J$\uparrow$} & {RECAL-J$\uparrow$} & {F-SCORE-J$\uparrow$} & {\#Lines$\uparrow$} \\
    \midrule
    \multirow{4}{*}{LSD~\cite{LSD}}
    & LIMAP~\cite{limap} & 0.0748 & 0.4094 & 0.5912 & 0.0807 & 0.1436 & 0.0753 & 0.4183 & 0.5873 & 0.0502 & 0.0927 & 328 \\
    & LIMAP w/ depth~\cite{limap} & 0.1083 & 0.5499 & 0.4853 & 0.0253 & 0.0454 & 0.1137 & 0.5540 & 0.4904 & 0.0161 & 0.0298 & 241 \\
    & CLMAP~\cite{Bai_2024_CLMAP} & \textbf{0.0692} & 0.3326 & 0.5893 & 0.1333 & 0.2228 & \textbf{0.0702} & 0.3494 & 0.5837 & 0.0728 & 0.1302 & 529 \\
    & Ours & 0.0765 & \textbf{0.1260} & \textbf{0.7069} & \textbf{0.3271} & \textbf{0.4438} & 0.0774 & \textbf{0.1433} & \textbf{0.7068} & \textbf{0.2083} & \textbf{0.3183} & \textbf{2974} \\
    \midrule
    \multirow{4}{*}{HAWPv3}
    & LIMAP~\cite{limap} & 0.0791 & 0.4510 & 0.6699 & 0.0628 & 0.1139 & 0.0811 & 0.4636 & 0.6547 & 0.0293 & 0.0555 & 172 \\
    & LIMAP w/ depth~\cite{limap} & 0.0833 & 0.4092 & 0.5402 & 0.0553 & 0.0960 & 0.0853 & 0.4285 & 0.5326 & 0.0231 & 0.0426 & 237 \\
    & CLMAP~\cite{Bai_2024_CLMAP} & 0.0782 & 0.3955 & 0.7027 & 0.0895 & 0.1575 & 0.0799 & 0.4090 & 0.6927 & 0.0472 & 0.0873 & 389 \\
    & Ours & \textbf{0.0735} & \textbf{0.1168} & \textbf{0.7424} & \textbf{0.3521} & \textbf{0.4744} & \textbf{0.0743} & \textbf{0.1350} & \textbf{0.7398} & \textbf{0.2156} & \textbf{0.3301} & \textbf{2942} \\
    \midrule
    \multirow{4}{*}{DeepLSD}
    & LIMAP~\cite{limap} & 0.0733 & 0.3673 & 0.6451 & 0.1064 & 0.1814 & 0.0736 & 0.3816 & 0.6436 & 0.0571 & 0.1038 & 397 \\
    & LIMAP w/ depth~\cite{limap} & 0.0783 & 0.3469 & 0.5097 & 0.0804 & 0.1297 & 0.0786 & 0.3615 & 0.5165 & 0.0420 & 0.0732 & 658 \\
    & CLMAP~\cite{Bai_2024_CLMAP} & \textbf{0.0748} & 0.2963 & 0.6624 & 0.1610 & 0.2585 & \textbf{0.0758} & 0.3185 & 0.6569 & 0.0798 & 0.1407 & 613 \\
    & Ours & 0.0776 & \textbf{0.1202} & \textbf{0.7308} & \textbf{0.3504} & \textbf{0.4650} & 0.0781 & \textbf{0.1388} & \textbf{0.7310} & \textbf{0.2158} & \textbf{0.3292} & \textbf{3015} \\
    \midrule
    \multirow{4}{*}{ScaleLSD}
    & LIMAP~\cite{limap} & 0.0828 & 0.3217 & 0.6413 & 0.1294 & 0.2155 & 0.0843 & 0.3356 & 0.6326 & 0.0710 & 0.1263 & 499 \\
    & LIMAP w/ depth~\cite{limap} & 0.0829 & 0.2782 & 0.5437 & 0.1182 & 0.1840 & 0.0840 & 0.2977 & 0.5411 & 0.0589 & 0.1004 & 733 \\
    & CLMAP~\cite{Bai_2024_CLMAP} & 0.0827 & 0.2711 & 0.6939 & 0.1765 & 0.2822 & 0.0840 & 0.2831 & 0.6863 & 0.1121 & 0.1915 & 1186 \\
    & Ours & \textbf{0.0742} & \textbf{0.1384} & \textbf{0.7422} & \textbf{0.3399} & \textbf{0.4430} & \textbf{0.0743} & \textbf{0.1537} & \textbf{0.7431} & \textbf{0.2078} & \textbf{0.3204} & \textbf{2934} \\
    \midrule
        
    \multicolumn{2}{c}{\textit{ScanNet++}~\cite{scannetpp-YeshwanthLND23}: 30 scenes} & \multicolumn{5}{c}{Line-Level Metrics} & \multicolumn{5}{c}{Junction-Level Metrics} & \multicolumn{1}{c}{Statistics} \\
    \cmidrule(lr){1-2} \cmidrule(lr){3-7} \cmidrule(lr){8-12} \cmidrule(lr){13-13}
    Detector & Method & {ACC-L$\downarrow$} & {COMP-L$\downarrow$} & {PREC-L$\uparrow$} & {RECAL-L$\uparrow$} & {F-SCORE-L$\uparrow$} & {ACC-J$\downarrow$} & {COMP-J$\downarrow$} & {PREC-J$\uparrow$} & {RECAL-J$\uparrow$} & {F-SCORE-J$\uparrow$} & {\#Lines$\uparrow$} \\
    \midrule
    \multirow{4}{*}{LSD}
    & LIMAP~\cite{limap} & 0.0718 & 1.1532 & 0.5283 & 0.0281 & 0.0515 & 0.0721 & 1.1604 & 0.5396 & 0.0141 & 0.0266 & 73 \\
    & LIMAP w/ depth~\cite{limap} & 0.1084 & 1.5770 & 0.4355 & 0.0025 & 0.0050 & 0.1038 & 1.5790 & 0.4593 & 0.0014 & 0.0029 & 16 \\
    & CLMAP~\cite{Bai_2024_CLMAP} & \textbf{0.0648} & 0.8684 & 0.5308 & 0.0689 & 0.1183 & \textbf{0.0633} & 0.8821 & 0.5403 & 0.0344 & 0.0689 & 218 \\
    & Ours & 0.0759 & \textbf{0.2585} & \textbf{0.5653} & \textbf{0.2900} & \textbf{0.3833} & 0.0744 & \textbf{0.2718} & \textbf{0.5583} & \textbf{0.1432} & \textbf{0.2279} & \textbf{3795} \\
    \midrule
    \multirow{4}{*}{HAWPv3}
    & LIMAP~\cite{limap} & 0.0870 & 1.2963 & 0.4876 & 0.0194 & 0.0375 & 0.0878 & 1.3043 & 0.4902 & 0.0081 & 0.0160 & 41 \\
    & LIMAP w/ depth~\cite{limap} & 0.1337 & 0.5728 & 0.3699 & 0.0343 & 0.0617 & 0.1333 & 0.5918 & 0.3743 & 0.0144 & 0.0273 & 228 \\
    & CLMAP~\cite{Bai_2024_CLMAP} & \textbf{0.0783} & 1.0479 & 0.5578 & 0.0346 & 0.0635 & \textbf{0.0792} & 1.0561 & 0.5635 & 0.0186 & 0.0360 & 141 \\
    & Ours & 0.0890 & \textbf{0.2428} & \textbf{0.5724} & \textbf{0.2664} & \textbf{0.3636} & 0.0906 & \textbf{0.2568} & \textbf{0.5761} & \textbf{0.1741} & \textbf{0.2673} & \textbf{2691} \\
    \midrule
    \multirow{4}{*}{DeepLSD}
    & LIMAP~\cite{limap} & 0.0785 & 1.0347 & 0.4114 & 0.0416 & 0.0771 & 0.0769 & 1.0465 & 0.4278 & 0.0185 & 0.0356 & 102 \\
    & LIMAP w/ depth~\cite{limap} & 0.1145 & 0.8208 & 0.3629 & 0.0151 & 0.0282 & 0.1114 & 0.8335 & 0.3685 & 0.0071 & 0.0137 & 108 \\
    & CLMAP~\cite{Bai_2024_CLMAP} & \textbf{0.0729} & 0.8035 & 0.5034 & 0.0863 & 0.1512 & \textbf{0.0715} & 0.8220 & 0.5179 & 0.0396 & 0.0741 & 280 \\
    & Ours & 0.0828 & \textbf{0.2466} & \textbf{0.5770} & \textbf{0.3056} & \textbf{0.3996} & 0.0801 & \textbf{0.2621} & \textbf{0.5669} & \textbf{0.2181} & \textbf{0.3157} & \textbf{3948} \\
    \midrule
    \multirow{4}{*}{ScaleLSD}
    & LIMAP~\cite{limap} & 0.0825 & 1.0197 & 0.4507 & 0.0402 & 0.0752 & 0.0824 & 1.0280 & 0.4498 & 0.0195 & 0.0376 & 110 \\
    & LIMAP w/ depth~\cite{limap} & 0.1201 & 0.3798 & 0.3907 & 0.0920 & 0.1453 & 0.1191 & 0.4039 & 0.3965 & 0.0445 & 0.0782 & 740 \\
    & CLMAP~\cite{Bai_2024_CLMAP} & 0.0813 & 0.8639 & 0.5663 & 0.0708 & 0.1272 & 0.0812 & 0.8714 & 0.5677 & 0.0433 & 0.0807 & 382 \\
    & Ours & \textbf{0.0801} & \textbf{0.2520} & \textbf{0.5889} & \textbf{0.2895} & \textbf{0.3882} & \textbf{0.0811} & \textbf{0.2665} & \textbf{0.5959} & \textbf{0.2039} & \textbf{0.3038} & \textbf{2749} \\
    \midrule
    \end{tabular}
    }
\end{table*}

This section presents experimental results on public datasets and ablation studies to demonstrate the effectiveness of our method and its components. More details and additional results are provided in our Appendix.

\subsection{Implementation Details}
\paragraph{Initialization.}
We first use the provided depth maps from all views to generate a coarse scene mesh. From this mesh, we randomly sampled 2,000 points to serve as the initial centers of our 3D planar primitives. The initial radius of each primitive $\pi$ is set to half the minimum distance from $\pi$ to its nearest neighboring primitive. The plane rotation is initialized using the surface normal of the coarse mesh.

\paragraph{Optimization.}
Our \method is implemented in PyTorch~\cite{pytorch-PaszkeGMLBCKLGA19} and optimized with Adam~\cite{adam-KingmaB14}. For each scene, we jointly optimize lines and planes from scratch for 60 epochs, with each epoch traversing all available viewpoints once. The learning rates of the learnable parameters, including plane centers, radii, and rotations, are fixed at 0.001.

\begin{table}[!t]
    \centering
    \scriptsize
    \caption{Quantitative results of 3D line mapping on the ScanNetV2~\cite{scannet-DaiCSHFN17} dataset and the ScanNet++~\cite{scannetpp-YeshwanthLND23} dataset with lines detected by 4 different detectors. $R_{\tau}$ and $P_{\tau}$ are reported at 5 mm, 10 mm, 50 mm along with the average number of supporting images/lines.}
    \label{tab:scannet2}
    \resizebox{\linewidth}{!}{%
    \begin{tabular}{l|l|ccccccc}
    \toprule
    \multicolumn{9}{c}{\textit{ScanNetV2}~\cite{scannet-DaiCSHFN17}: 50 scenes} \\
    \midrule
    Detector & Method & R5 & R10 & R50 & P5 & P10 & P50 & \# supports \\
    \midrule
    \multirow{4}{*}{LSD}
    & LIMAP~\cite{limap} & 1.04 & 6.41 & 31.07 & 25.8 & \textbf{45.4} & 80.2 & 5.6 / 5.8 \\
    & LIMAP w/ depth~\cite{limap} & 0.19 & 1.24 & 13.97 & 7.3 & 14.4 & 54.9 & 4.2 / 4.5 \\
    & CLMAP~\cite{Bai_2024_CLMAP} & 1.50 & 14.20 & 85.05 & 24.2 & 44.6 & 79.6 & 8.6 / 8.8 \\
    & Ours & \textbf{6.23} & \textbf{40.23} & \textbf{240.29} & \textbf{27.6} & 43.0 & \textbf{81.7} & \textbf{10.7} / \textbf{14.1} \\
    \midrule
    \multirow{4}{*}{HAWPv3}
    & LIMAP~\cite{limap} & 0.90 & 5.61 & 28.25 & 28.5 & \textbf{48.4} & 82.0 & 7.1 / 12.1 \\
    & LIMAP w/ depth~\cite{limap} & 0.80 & 2.94 & 19.11 & 17.6 & 27.8 & 64.7 & 4.4 / 5.8 \\
    & CLMAP~\cite{Bai_2024_CLMAP} & 1.63 & 14.80 & 80.68 & 28.3 & 47.7 & 81.1 & \textbf{8.5} / \textbf{9.1} \\
    & Ours & \textbf{5.26} & \textbf{33.12} & \textbf{183.61} & \textbf{30.9} & 47.6 & \textbf{84.6} & 6.3 / 8.1 \\
    \midrule
    \multirow{4}{*}{DeepLSD}
    & LIMAP~\cite{limap} & 1.67 & 10.41 & 53.78 & 30.0 & \textbf{48.3} & 81.3 & 6.4 / 6.8 \\
    & LIMAP w/ depth~\cite{limap} & 0.96 & 6.45 & 46.18 & 10.8 & 19.2 & 60.1 & 4.3 / 4.5 \\
    & CLMAP~\cite{Bai_2024_CLMAP} & 1.97 & 18.81 & 118.22 & 24.5 & 46.2 & 81.8 & \textbf{9.4} / \textbf{9.6} \\
    & Ours & \textbf{4.63} & \textbf{29.74} & \textbf{168.16} & \textbf{31.7} & 47.1 & \textbf{85.6} & 6.2 / 6.4 \\
    \midrule
    \multirow{4}{*}{ScaleLSD}
    & LIMAP~\cite{limap} & 2.08 & 12.94 & 64.40 & \textbf{31.3} & 49.9 & 86.2 & 7.2 / 10.2 \\
    & LIMAP w/ depth~\cite{limap} & 1.78 & 11.49 & 54.33 & 15.0 & 25.3 & 64.5 & 4.5 / 4.4 \\
    & CLMAP~\cite{Bai_2024_CLMAP} & 4.09 & 38.01 & 204.94 & 27.6 & \textbf{50.7} & 85.0 & \textbf{8.4} / 9.1 \\
    & Ours & \textbf{6.74} & \textbf{42.65} & \textbf{224.81} & 30.4 & 48.6 & \textbf{87.4} & 8.2 / \textbf{12.8} \\
    \midrule
    
    \multicolumn{9}{c}{\textit{ScanNet++}~\cite{scannetpp-YeshwanthLND23}: 30 scenes} \\
    \midrule
    Detector & Method & R5 & R10 & R50 & P5 & P10 & P50 & \# supports \\
    \midrule
    \multirow{4}{*}{LSD}
    & LIMAP~\cite{limap} & 2.07 & 5.69 & 17.27 & 33.8 & 50.1 & 69.1 & 4.7 / 5.0 \\
    & LIMAP w/ depth~\cite{limap} & 0.07 & 0.23 & 1.03 & 12.3 & 18.8 & 51.6 & 4.1 / 4.2 \\
    & CLMAP~\cite{Bai_2024_CLMAP} & 5.63 & 16.29 & 49.65 & \textbf{36.8} & 52.4 & 71.5 & \textbf{5.8} / \textbf{6.2} \\
    & Ours & \textbf{12.87} & \textbf{48.64} & \textbf{61.20} & 29.1 & \textbf{64.0} & \textbf{74.1} & 4.9 / 5.4 \\
    \midrule
    \multirow{4}{*}{HAWPv3}
    & LIMAP~\cite{limap} & 1.33 & 3.78 & 11.23 & 33.5 & 58.2 & 78.9 & 5.1 / 6.2 \\
    & LIMAP w/ depth~\cite{limap} & 0.24 & 0.69 & 3.06 & 18.1 & 27.1 & 57.1 & 4.2 / 4.3 \\
    & CLMAP~\cite{Bai_2024_CLMAP} & 4.09 & 11.86 & 33.80 & 32.4 & 59.3 & \textbf{82.8} & \textbf{5.8} / 6.4 \\
    & Ours & \textbf{4.33} & \textbf{16.90} & \textbf{40.67} & \textbf{34.3} & \textbf{72.0} & 81.5 & 4.7 / \textbf{7.1} \\
    \midrule
    \multirow{4}{*}{DeepLSD}
    & LIMAP~\cite{limap} & 3.05 & 8.96 & 28.36 & 38.8 & 53.8 & 74.7 & 5.2 / 5.9 \\
    & LIMAP w/ depth~\cite{limap} & 0.41 & 1.42 & 7.11 & 15.4 & 22.6 & 58.7 & 4.0 / 4.2 \\
    & CLMAP~\cite{Bai_2024_CLMAP} & 8.37 & 24.13 & 76.47 & 38.1 & 63.6 & \textbf{79.5} & \textbf{6.3} / \textbf{6.6} \\
    & Ours & \textbf{11.04} & \textbf{45.22} & \textbf{59.61} & \textbf{39.9} & \textbf{73.8} & 76.6 & 4.9 / 5.3 \\
    \midrule
    \multirow{4}{*}{ScaleLSD}
    & LIMAP~\cite{limap} & 2.81 & 7.99 & 24.36 & 39.8 & 56.6 & 80.3 & 5.3 / 6.1 \\
    & LIMAP w/ depth~\cite{limap} & 0.61 & 1.93 & 9.50 & 16.7 & 23.1 & 59.4 & 4.1 / 4.0 \\
    & CLMAP~\cite{Bai_2024_CLMAP} & 10.04 & 28.73 & 80.88 & \textbf{40.4} & 67.3 & \textbf{83.1} & \textbf{5.9} / 6.6 \\
    & Ours & \textbf{14.31} & \textbf{43.91} & \textbf{99.01} & 36.5 & \textbf{71.7} & 81.5 & 4.6 / \textbf{6.7} \\
    \bottomrule
    \end{tabular}
    }
\vspace{-10pt}
\end{table}

\paragraph{Plane Pruning and Splitting.}
During optimization, we adopt the plane-splitting operation introduced in PlanarSplatting~\cite{PlanarSplatting2024}, guided by the gradients of the plane radii to better fit the geometry of the scene. If the average radius gradients along the X-axis ($r_\pi^{x+}$ and $r_\pi^{x-}$) exceed 0.2, the plane is split along the Y-axis. Similarly, if the radius gradients along the Y-axis ($r_\pi^{y+}$ and $r_\pi^{y-}$) are greater than 0.2, the plane is split along the X-axis. Unlike PlanarSplatting, which performs pruning and splitting every 1,000 iterations, we apply these operations once per epoch.

\subsection{Datasets, Metrics and Baselines}

\paragraph{Datasets.}
We test our method on the ScanNetV2 dataset~\cite{scannet-DaiCSHFN17}, the ScanNet++ dataset~\cite{scannetpp-YeshwanthLND23} and the Hypersim dataset~\cite{Hypersim}.
The ScanNetV2 and the ScanNet++ are two realistic indoor datasets that provide posed videos. Following the test protocol according to PlanarSplatting~\cite{PlanarSplatting2024}, we randomly sample 50 scenes from the ScanNetV2 dataset and 30 scenes from the ScanNet++ dataset, and sample one image every 10 frames from the video of each scene for evaluation.
Similarly to PlanarSplatting~\cite{PlanarSplatting2024}, we use Metric3Dv2~\cite{Metric3Dv2} for depth estimation and Ominidata~\cite{omnidata-EftekharSMZ21} for normal estimation.
The Hypersim is a photorealistic synthetic dataset for holistic indoor scene understanding. Each scene of this dataset provides a maximum of 100 posed images, depth maps, and normal maps. Following LIMAP~\cite{limap} and CLMAP~\cite{Bai_2024_CLMAP}, we take the first ten scenes for evaluation.

\begin{figure*}[!t]
    \centering
    \includegraphics[width=1.0\linewidth]{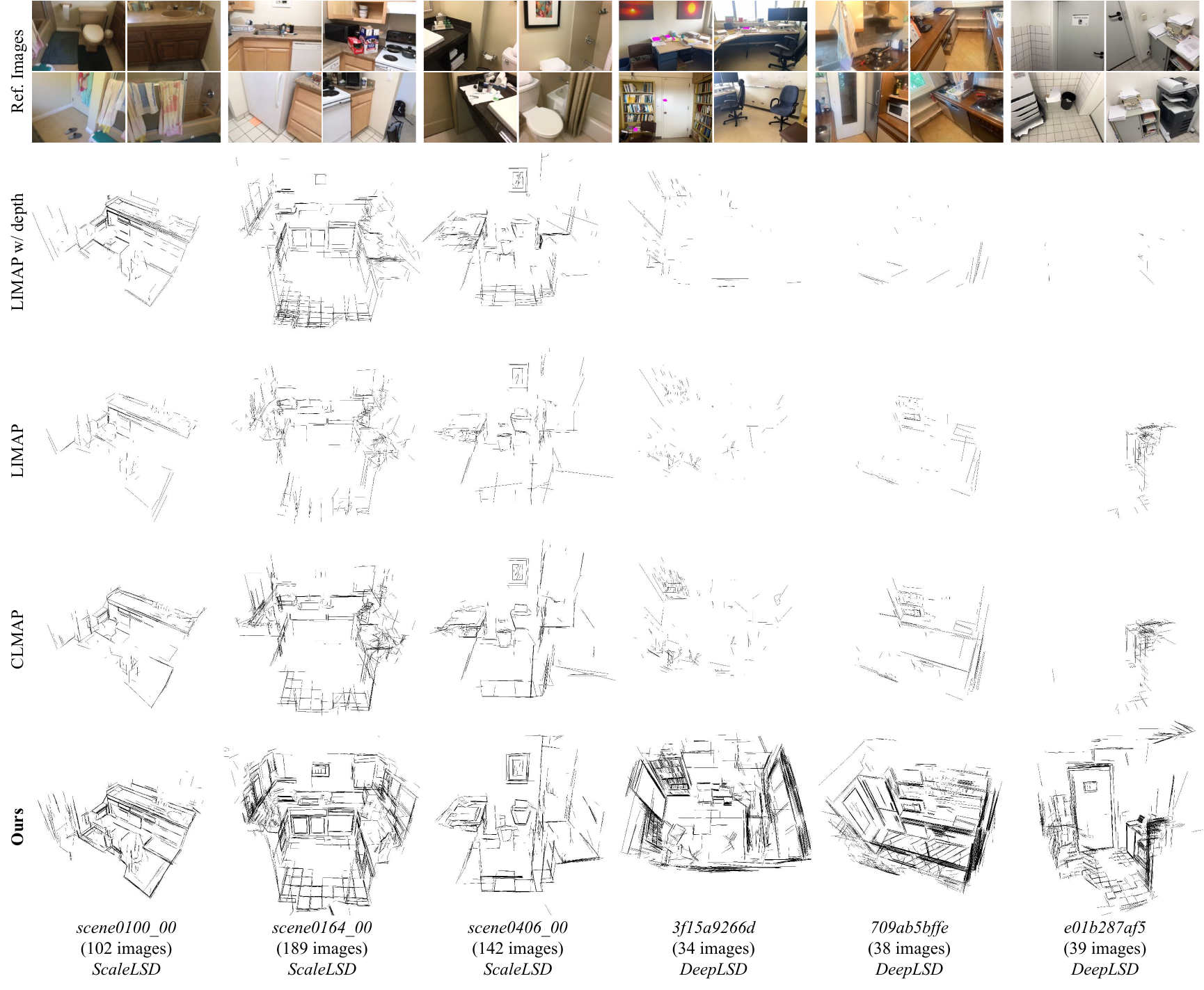}
    \caption{Qualitaitve comparison on the ScanNetV2 dataset~\cite{scannet-DaiCSHFN17} (left 3 scenes with ScaleLSD~\cite{ScaleLSD} detector) and the ScanNet++ dataset~\cite{scannetpp-YeshwanthLND23} (right 3 scenes with DeepLSD~\cite{DeepLSD} detector). First row: LIMAP w/ depth~\cite{limap}. Second row: LIMAP~\cite{limap}. Third row: CLMAP~\cite{Bai_2024_CLMAP}. Last row: Ours.}
    \vspace{-5pt}
    \label{fig:scannet}
\end{figure*}

\paragraph{Metrics.}
As there are no ground-truth (GT) 3D lines, we evaluate the 3D line mapping with either GT mesh models or point clouds.
To comprehensively evaluate the quality of the reconstructed 3D line maps, we consider two evaluation metrics: one is the widely used standard benchmark for the evaluation of reconstructed 3D models, and the other is the evaluation benchmark proposed by LIMAP~\cite{limap}.
\begin{itemize}
\item The widely used metrics (M1) adopted in 3D reconstructions~\cite{AirPlanes-WatsonASQABFV24, simplerec, NEAT-TXDX0S24}, including accuracy (ACC), completeness (COMP), precision (PREC), recall (RECAL), and F-SCORE for evaluation. We assess these five metrics at both the junction level (-J) and the line level (-L). For junction-level results, we evaluate the performance of the two endpoints (junctions) of the 3D lines. For line-level results, we uniformly sample 100 points on each line to measure performance. Additionally, we apply the number of reconstructed 3D line segments as a statistical metric for completeness and robustness.
\item LIMAP's evaluation metrics (M2) are built on the line tracking and use the following metrics: Length recall (in meters) at threshold $\tau$ ($R_{\tau}$) means the sum of the lengths of the line portions within $\tau$ mm from the GT model. Inlier percentage at threshold $\tau$ ($P_{\tau}$) means the percentage of tracks that are within $\tau$ mm from the GT model. Average supports means the average number of image supports and 2D line supports across all line tracks.
\end{itemize}
The complete formulation of the evaluation metrics is provided in the Appendix~B.

\paragraph{Baselines.}
We use the current state-of-the-art method, LIMAP~\cite{limap} as the baseline, which achieves 3D line mapping through carefully crafted scoring and tracking, as well as automatically identifying and exploiting structural priors.
Since we use depth maps for 3D line mapping and LIMAP provides optimization with depth maps, we compare with depth-based LIMAP in all experiments.
We also compare with the recent CLMAP~\cite{Bai_2024_CLMAP}, which is built upon LIMAP and is proposed to improve the consistency of the structure of the line map by jointly optimizing points, lines, and planes. Different from our learnable planar primitives, CLMAP statically applies planes detected by RSPD~\cite{PlaneDetection} for coplanar constraints of 3D lines.

\begin{figure}[t]
    \centering
    \includegraphics[width=1.0\linewidth]{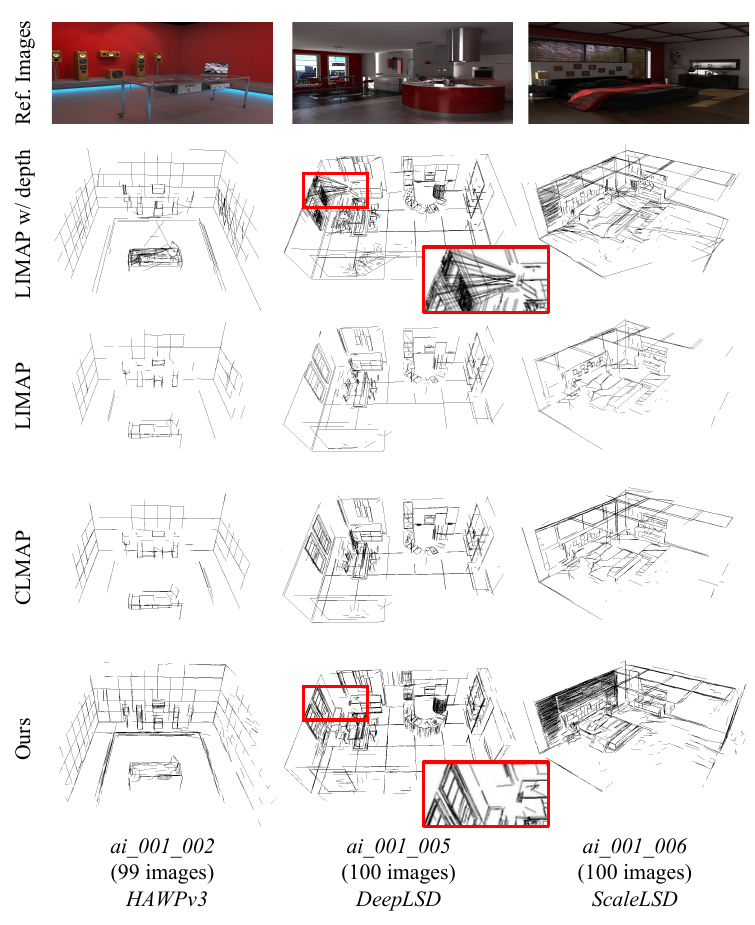}
    \caption{Qualitative comparison on the Hypersim dataset~\cite{Hypersim} using 3 different line detectors. First row: LIMAP w/ depth~\cite{limap}. Second row: LIMAP~\cite{limap}. Third row: CLMAP~\cite{Bai_2024_CLMAP}. Last row: Ours.}
    \vspace{-5pt}
    \label{fig:hypersim}
\end{figure}

\subsection{3D Line Mapping}

\paragraph{Detectors.}
To systematically investigate the impact of detected 2D line segments on the final 3D line mapping, we conduct comprehensive experiments using four distinct line detectors: LSD~\cite{LSD}, HAWPv3~\cite{HAWPv3}, DeepLSD~\cite{DeepLSD}, and ScaleLSD~\cite{ScaleLSD}, all within a unified quantitative evaluation.

\begin{table}[!t]
    \centering
    \caption{Quantitative results of 3D line mapping on the Hypersim~\cite{Hypersim} dataset with lines detected by 4 different detectors. All metrics are reported at 5 mm along with the average number of reconstructed lines.}
    \label{tab:hypersim1}
    \resizebox{\linewidth}{!}{
    \begin{tabular}{l|l|ccccc|c}
    \toprule
    \multicolumn{2}{c}{\textit{Hypersim}~\cite{Hypersim}: First 10 scenes} & \multicolumn{5}{c}{Line-Level Metrics} & \multicolumn{1}{c}{Statistics} \\
    \cmidrule(lr){1-2} \cmidrule(lr){3-7} \cmidrule(lr){8-8}
    Detector & Method & {ACC-L$\downarrow$} & {COMP-L$\downarrow$} & {PREC-L$\uparrow$} & {RECAL-L$\uparrow$} & {F-SCORE-L$\uparrow$} & {\#Lines$\uparrow$} \\
    \midrule
    \multirow{4}{*}{LSD~\cite{LSD}}
    & LIMAP~\cite{limap} & 0.0731 & 0.5336 & 0.9302 & 0.1890 & 0.3085 & 886 \\
    & LIMAP w/ depth~\cite{limap} & 0.0088 & 0.5800 & \textbf{0.9969} & 0.2024 & 0.3297 & 890 \\
    & CLMAP~\cite{Bai_2024_CLMAP} & 0.1045 & 0.5010 & 0.9136 & 0.2028 & 0.3248 & 1209 \\
    & Ours      & \textbf{0.0079} & \textbf{0.3989} & 0.9921 & \textbf{0.2908} & \textbf{0.4424} & \textbf{2538} \\
    \midrule
    \multirow{4}{*}{HAWPv3}
    & LIMAP~\cite{limap} & 0.0435 & 0.7133 & 0.9030 & 0.0912 & 0.1648 & 263 \\
    & LIMAP w/ depth~\cite{limap} & 0.0097 & 0.7907 & 0.9896 & 0.1262 & 0.2214 & 238 \\
    & CLMAP~\cite{Bai_2024_CLMAP} & 0.0591 & 0.6678 & 0.9286 & 0.1087 & 0.1935 & 641 \\
    & Ours      & \textbf{0.0077} & \textbf{0.3531} & \textbf{0.9923} & \textbf{0.2785} & \textbf{0.4295} & \textbf{2461} \\
    \midrule
    \multirow{4}{*}{DeepLSD}
    & LIMAP~\cite{limap} & 0.0861 & 0.5174 & 0.8964 & 0.2154 & 0.3412 & 665 \\
    & LIMAP w/ depth~\cite{limap} & 0.0093 & 0.7168 & \textbf{0.9934} & 0.2465 & 0.3868 & 786 \\
    & CLMAP~\cite{Bai_2024_CLMAP} & 0.1088 & \textbf{0.4745} & 0.9044 & 0.2330 & 0.3622 & 1105 \\
    & Ours      & \textbf{0.0076} & 0.5269 & 0.9924 & \textbf{0.3387} & 0.\textbf{4963} & \textbf{3355} \\
    \midrule
    \multirow{4}{*}{ScaleLSD}
    & LIMAP~\cite{limap} & 0.0520 & 0.6432 & 0.8774 & 0.1341 & 0.2296 & 458 \\
    & LIMAP w/ depth~\cite{limap} & 0.0096 & 0.5397 & \textbf{0.9901} & 0.1794 & 0.2982 & 439 \\
    & CLMAP~\cite{Bai_2024_CLMAP} & 0.0664 & 0.5602 & 0.9161 & 0.1593 & 0.2679 & 1225 \\
    & Ours      & \textbf{0.0090} & \textbf{0.3410} & 0.9857 & \textbf{0.2944} & \textbf{0.4465} & \textbf{2968} \\
    \bottomrule
    \end{tabular}
    }
\end{table}

\begin{table}[!t]
    \centering
    \scriptsize
    \caption{Quantitative results of 3D line mapping on the Hypersim~\cite{Hypersim} dataset with lines detected by 4 different detectors. $R_{\tau}$ and $P_{\tau}$ are reported at 1 mm, 5 mm, 10 mm along with the average number of supporting images/lines.}
    \label{tab:hypersim2}
    \resizebox{\linewidth}{!}{%
    \begin{tabular}{l|l|ccccccc}
    \toprule
    \multicolumn{9}{c}{\textit{Hypersim}~\cite{Hypersim}: First 10 scenes} \\
    \midrule
    Detector & Method & R1 & R5 & R10 & P1 & P5 & P10 & \# supports \\
    \midrule
    \multirow{4}{*}{LSD}
    & LIMAP~\cite{limap} & 70.5  & 282.5  & 351.5  & 65.7  & 84.0  & 89.6  & 15.6 / 17.7 \\
    & LIMAP w/ depth~\cite{limap} & \textbf{268.0} & 432.6  & 434.1  & \textbf{84.0}  & \textbf{99.9} & \textbf{100} & 16.6 / \textbf{22.2} \\
    & CLMAP~\cite{Bai_2024_CLMAP} & 112.3 & 403.0  & 492.7  & 61.9  & 80.6  & 87.9  & \textbf{19.7} / 21.6 \\
    & Ours    & 245.3 & \textbf{585.6}  & \textbf{678.7}  & 81.7  & 94.6  & 98.7  & 12.3 / 20.8 \\
    \midrule
    \multirow{4}{*}{HAWPv3}
    & LIMAP~\cite{limap} & 15.4  & 66.1   & 95.1   & 65.0  & 81.0  & 88.4  & 12.3 / 23.5 \\
    & LIMAP w/ depth~\cite{limap} & 82.2  & 182.3  & 189.9  & 80.8  & \textbf{99.7}  & \textbf{100} & \textbf{15.8} / \textbf{33.0} \\
    & CLMAP~\cite{Bai_2024_CLMAP} & 52.7  & 205.0  & 272.7  & 63.3  & 83.7  & 89.6  & 13.7 / 14.7 \\
    & Ours    & \textbf{184.3} & \textbf{512.9}  & \textbf{594.5}  & \textbf{82.8}  & 95.7  & 99.2  & 9.3 / 18.2 \\
    \midrule
    \multirow{4}{*}{DeepLSD}
    & LIMAP~\cite{limap} & 65.3  & 244.2  & 304.4  & 66.4  & 80.7  & 85.6  & 17.2 / 23.5 \\
    & LIMAP w/ depth~\cite{limap} & 261.8 & 492.8  & 503.1  & \textbf{84.2}  & \textbf{99.8}  & \textbf{100} & 17.5 / \textbf{24.9} \\
    & CLMAP~\cite{Bai_2024_CLMAP} & 138.0 & 455.8  & 551.1  & 63.8  & 80.7  & 87.5  & \textbf{20.5} / 21.9 \\
    & Ours    & \textbf{396.3} & \textbf{886.5}  & \textbf{1010.5} & 82.2  & 95.1  & 98.7  & 13.1 / 21.9 \\
    \midrule
    \multirow{4}{*}{ScaleLSD}
    & LIMAP~\cite{limap} & 24.5  & 100.6  & 142.4  & 64.2  & 80.9  & 87.0  & 14.0 / 29.3 \\
    & LIMAP w/ depth~\cite{limap} & 136.2 & 275.1  & 287.9  & 79.2  & \textbf{99.7}  & \textbf{100} & \textbf{15.6} / \textbf{35.6} \\
    & CLMAP~\cite{Bai_2024_CLMAP} & 107.9 & 391.1  & 533.9  & 62.9  & 82.2  & 89.1  & 14.7 / 15.9 \\
    & Ours    & \textbf{223.6} & \textbf{639.6}  & \textbf{754.1}  & \textbf{80.2}  & 95.4  & 98.7  & 11.0 / 32.0 \\
    \bottomrule
    \end{tabular}
    }
\end{table}

\paragraph{ScanNetV2 \& ScanNet++.}
We first evaluate our approach on the ScanNetV2 dataset~\cite{scannet-DaiCSHFN17} and the ScanNet++ dataset~\cite{scannetpp-YeshwanthLND23}. We use the corresponding plane mesh of each scene provided by these two datasets for the computation of metrics M1 and M2.
The average results of metric M1 are illustrated in Table~\ref{tab:scannet1}.
As can be seen that our method achieves a superior balance between accuracy and completeness of reconstructed 3D lines, with consistently lower completeness error and comparable accuracy error while maintaining higher precision, recall, and F1 scores.
Matching-based LIMAP and CLMAP are robust to reconstruct more accurate 3D line maps on the whole, and our method tends to reconstruct more complete line structures of the scene with a much larger number of 3D lines.
Due to the motion blur of images and the not-perfect depth maps estimated from Metric3dv2~\cite{Metric3Dv2}, LIMAP with depth would fit unreliable 3D lines from the detected 2D lines and get worse reconstruction results finally.
But this problem can be greatly alleviated in our method since the planar primitive can average the depth error of each region, so that our method achieves lower accuracy errors than LIMAP with depth.
The average results of metric M2 are reported in Table~\ref{tab:scannet2}.
The length recall of our method is significantly higher than that of other methods, albeit with competitive precision scores. LIMAP and CLMAP are better at building long line tracks and can track more image supports and line supports than our method.

The qualitative comparison of 3D line mapping is shown in Fig.~\ref{fig:scannet}. The scene name, number of used images, and the used line detector are tagged under each column. Our method excels not only in capturing fine structures but also in reconstructing a more comprehensive line map overall. LIMAP and CLMAP can basically reconstruct the structure of the whole scene on the ScanNetV2 dataset while using dense views with over 100 images, but they failed to reconstruct the complete structures of scenes on the ScanNet++ dataset with relatively sparse views.

\begin{figure}[!t]
    \centering
    \includegraphics[width=\linewidth]{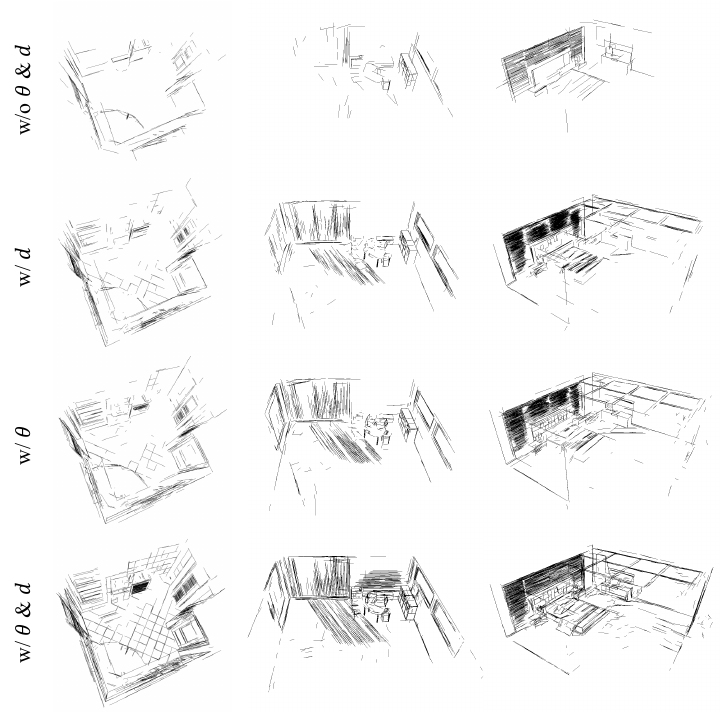}
    \caption{Illustration on the ablation of our proposed 2D line-3D edge assignment. First row: w/o $\theta$ \& $d$. Second row: only w/ $\theta$. Third row: only w/ $d$. Last row: w/ $\theta$ \& $d$.}
    \label{fig:abs_theta_dist}
\end{figure}

\paragraph{Hypersim.}
We further present comparisons of all methods on the first ten scenes of the Hypersim dataset~\cite{Hypersim}.
Following LIMAP~\cite{limap}, we use the point cloud recovered from the depth maps of each scene for the computation of the metrics M1 and M2.
The average results of metric M1 are reported in Table~\ref{tab:hypersim1}. As illustrated, our method and LIMAP with depth outperform LIMAP and CLMAP in terms of much lower accuracy error and higher precision, since Hypersim provides accurate depth maps.
Additionally, our method significantly outperforms other methods in terms of lower completeness error and higher recall and F1 scores. Our method can reconstruct more complete structures with more 3D lines.
The average results of metric M2 are reported in Table~\ref{tab:hypersim2}. Our method also achieves higher length recall than other methods, and a slightly lower precision than LIMAP with depth, while being less capable of building long line tracks than LIMAP and CLMAP.

The qualitative comparison of 3D line mapping is shown in Fig.~\ref{fig:hypersim}. We take reconstructed line maps of 3 scenes with 3 different line detectors for illustration. Our method reconstructs clearer and more complete line structures than other methods. LIMAP with depth also outperforms LIMAP and CLMAP.
But LIMAP with depth still has the risk of reconstructing unreasonable lines because it would fit unreliable 3D lines from detected 2D lines with incorrect depth maps, as indicated by the red square in Fig.~\ref{fig:hypersim}. This kind of case will increase the reconstruction error of LIMAP with depth.

\begin{table}[!t]
    \centering
    \caption{Ablation study of our proposed loss functions in \textbf{metric M1} on the Hypersim dataset~\cite{scannet-DaiCSHFN17} with lines detected by LSD~\cite{LSD}. All metrics are reported at 5 mm along with the average number of reconstructed lines.}
    \label{tab:abs_loss1}
    \resizebox{\linewidth}{!}{
        \begin{tabular}{cccccccccc}
        \toprule
        $\mathcal{L}_{euc}^{2d}$ & $\mathcal{L}_{ort}^{2d}$ & $\mathcal{L}_{group}$ & {ACC-L$\downarrow$} & {COMP-L$\downarrow$} & {PREC-L$\uparrow$} & {RECAL-L$\uparrow$} & {F-SCORE-L$\uparrow$} & {\#Lines$\uparrow$} \\
        \midrule
        & & & 0.0082 & 0.8447 & 0.9551 & 0.1269 & 0.2240 & 532   \\
        \checkmark & & & 0.0088 & 0.4648 & 0.9421 & 0.1512 & 0.2606 & 1358  \\
        & \checkmark & & 0.0083 & 0.4472 & 0.9441 & 0.1619 & 0.2764 & 1654  \\
        & & \checkmark & 0.0082 & 0.4689 & 0.9446 & 0.1330 & 0.2332 & 1301  \\
        \checkmark & \checkmark & & 0.0087 & 0.4516 & 0.9585 & 0.2180 & 0.3552 & 3109  \\
        \checkmark & & \checkmark & 0.0088 & 0.4232 & 0.9534 & 0.2011 & 0.3321 & 2929  \\
        & \checkmark & \checkmark & 0.0081 & 0.4566 & 0.9662 & 0.2504 & 0.3977 & \textbf{3439}  \\
        \checkmark  & \checkmark & \checkmark & \textbf{0.0079} & \textbf{0.3989} & \textbf{0.9921} & \textbf{0.2908} & \textbf{0.4424} & 2538  \\
        \bottomrule
        \end{tabular}
    }
\end{table}

\begin{table}[!t]
    \centering
    \caption{Ablation study of our proposed loss functions in \textbf{metric M2} on the Hypersim dataset~\cite{scannet-DaiCSHFN17} with lines detected by LSD~\cite{LSD}. $R_{\tau}$ and $P_{\tau}$ are reported at 1 mm, 5 mm, 10 mm along with the average number of supporting images/lines.}
    \label{tab:abs_loss2}
    \resizebox{\linewidth}{!}{
        \begin{tabular}{cccccccccc}
        \toprule
        $\mathcal{L}_{euc}^{2d}$ & $\mathcal{L}_{ort}^{2d}$ & $\mathcal{L}_{group}$ & R1 & R5 & R10 & P1 & P5 & P10 & \# supports \\
        \midrule
        & & & 11.7 & 32.8 & 54.1 & 81.0 & 92.6 & 98.6 & 5.3 / 6.2  \\
        \checkmark & & & 31.5  & 99.4  & 121.1 & 81.1 & 92.0 & 97.5 & 10.0 / 11.3 \\
        & \checkmark & & 39.5  & 117.6 & 141.8 & 81.4 & 92.5 & 97.9 & 10.4 / 11.8 \\
        & & \checkmark & 29.7  & 97.2  & 118.5 & 80.9 & 91.9 & 98.2 & 10.2 / 11.4 \\
        \checkmark & \checkmark & & 143.7 & 476.7 & 569.2 & 81.5 & 93.0 & 98.4 & 10.8 / 13.1 \\
        \checkmark & & \checkmark & 135.3 & 468.0 & 566.2 & 81.7 & 93.2 & 98.1 & 10.9 / 13.1 \\
        & \checkmark & \checkmark & 115.2 & 335.5 & 389.4 & 81.3 & 94.0 & 97.9 & 11.3 / 13.3 \\
        \checkmark  & \checkmark & \checkmark & \textbf{245.3} & \textbf{585.6} & \textbf{678.7} & \textbf{81.7} & \textbf{94.6} & \textbf{98.7} & \textbf{12.3} / \textbf{20.8} \\
        \bottomrule
        \end{tabular}
    }
\end{table}

\begin{table}[!t]
    \centering
    \caption{Ablation study of the proposed 2D line-3D edge assignment in \textbf{metric M1} on the Hypersim dataset~\cite{scannet-DaiCSHFN17} with lines detected by LSD~\cite{LSD}. All metrics are reported at 5 mm along with the average number of reconstructed lines.}
    \label{tab:abs_theta_dist1}
    \resizebox{\linewidth}{!}{
        \begin{tabular}{ccccccccc}
        \toprule
        \textit{theta}($\theta$) & \textit{dist}($d$) & {ACC-L$\downarrow$} & {COMP-L$\downarrow$} & {PREC-L$\uparrow$} & {RECAL-L$\uparrow$} & {F-SCORE-L$\uparrow$} & {\#Lines$\uparrow$} \\
        \midrule
        & & 0.0082 & 0.8447 & 0.9551 & 0.1269 & 0.2240 & 532   \\
        \checkmark & & 0.0087 & 0.4051 & 0.9879 & 0.2461 & 0.3866 & 2125   \\
        & \checkmark & 0.0082 & \textbf{0.3542} & 0.9900 & 0.2776 & 0.4260 & \textbf{2561}   \\
        \checkmark  & \checkmark & \textbf{0.0079} & 0.3989 & \textbf{0.9921} & \textbf{0.2908} & \textbf{0.4424} & 2538   \\
        \bottomrule
        \end{tabular}
    }
\end{table}

\begin{table}[!t]
    \centering
    \caption{Ablation study of the proposed 2D line-3D edge assignment in \textbf{metric M2} on the Hypersim dataset~\cite{scannet-DaiCSHFN17} with lines detected by LSD~\cite{LSD}. $R_{\tau}$ and $P_{\tau}$ are reported at 1 mm, 5 mm, 10 mm along with the average number of supporting images/lines.}
    \label{tab:abs_theta_dist2}
    \resizebox{\linewidth}{!}{
        \begin{tabular}{ccccccccc}
        \toprule
        \textit{theta}($\theta$) & \textit{dist}($d$) & R1 & R5 & R10 & P1 & P5 & P10 & \# supports \\
        \midrule
        & & 11.7 & 32.8 & 54.1 & 81.0 & 92.6 & 98.6 & 5.3 / 6.2  \\
        \checkmark & & 138.9 & 351.2 & 418.7 & 76.2 & 92.2 & 97.4 & 10.8 / 13.3  \\
        & \checkmark & 177.4 & 421.6 & 492.9 & 76.9 & 92.0 & 96.9 & 11.5 / 14.1  \\
        \checkmark  & \checkmark & \textbf{245.3} & \textbf{585.6} & \textbf{678.7} & \textbf{81.7} & \textbf{94.6} & \textbf{98.7} & \textbf{12.3} / \textbf{20.8}   \\
        \bottomrule
        \end{tabular}
    }
\end{table}

\subsection{Ablation Studies}

\subsubsection{The Design of Hybrid Losses}
We primarily conduct our ablation experiments on the first ten scenes of the Hypersim dataset~\cite{Hypersim} to validate the effectiveness of each design of our hybrid losses through controlled comparisons.
The choosing checkmark denotes the use of the designed loss, and the case in which no loss terms were employed indicates the use of the baseline method of PlanarSplatting~\cite{PlanarSplatting2024}.
Comparative results under evaluation metric M1 are reported in Table~\ref{tab:abs_loss1}.
As can be observed, our loss design facilitates the reconstruction of a greater number of 3D line segments, significantly reducing the overall completeness error. The improvement in completeness is directly reflected in higher recall and F1 scores. Since our method builds upon the baseline of PlanarSplating and employs depth maps to supervise the positions of planes and lines, it achieves comparable performance in terms of reconstruction error and precision. Nevertheless, our approach demonstrates superior accuracy in 3D line mapping compared to the baseline.

We also report comparative results under evaluation metric M2 in Table~\ref{tab:abs_loss2}.
Consistent with the conclusions in Table~\ref{tab:abs_loss1}, all ablation experiments achieve comparable reconstruction precision. However, compared to the original baseline of PlanarSplatting, each proposed loss component significantly improves the length recall of the built line tracks and enables tracking of longer image and line segment trajectories. Furthermore, combining different loss functions leads to further improvements in length recall and increases the number of supported line segments.
The above ablation results validate the effectiveness of the proposed design of hybrid losses.

\begin{table*}[!t]
\centering
\caption{\rev{Quantitative results of reconstruction with different plane splitting thresholds on Hypersim~\cite{Hypersim}.}}
\label{tab:split_gradient_metrics}
\resizebox{0.8\linewidth}{!}{
    \begin{tabular}{l|cccccc|ccccccc}
    \toprule
    Splitting & \multicolumn{6}{c|}{Line-Level Metric M1} & \multicolumn{7}{c}{Line-Track Metric M2} \\
    \cmidrule(lr){2-7} \cmidrule(lr){8-14}
    Threshold & {ACC-L$\downarrow$} & {COMP-L$\downarrow$} & {PREC-L$\uparrow$} & {RECAL-L$\uparrow$} & {F-SCORE-L$\uparrow$} & {\#Lines$\uparrow$} & R1 & R5 & R10 & P1 & P5 & P10 & \# supports \\
    \midrule
    0.02 & 0.0083 & 0.5031 & 0.9899 & 0.3372 & 0.4933 & 8232 & 549.0 & 1336.0 & 2033.3 & 82.0 & 94.1 & 97.4 & 10.5 / 11.9 \\
    0.1 & 0.0079 & 0.5223 & 0.9862 & 0.3214 & 0.4766 & 4053 & 413.4 & 987.9 & 1364.1 & 78.8 & 91.5 & 95.7 & 11.9 / 16.5 \\
    0.2 & 0.0076 & 0.5269 & 0.9924 & 0.3387 & 0.4963 & 3355 & 396.3 & 886.5 & 1010.5 & 82.2 & 95.1 & 98.7 & 13.1 / 21.9 \\
    1.0 & 0.0104 & 0.6757 & 0.9284 & 0.2189 & 0.3442 & 1145 & 69.9 & 190.9 & 223.2 & 68.7 & 82.7 & 88.6 & 11.7 / 13.5 \\
    2.0 & 0.0118 & 0.5753 & 0.9127 & 0.1919 & 0.3066 & 888 & 51.2 & 147.1 & 173.2 & 67.1 & 81.5 & 87.9 & 11.6 / 13.4 \\
    10.0 & 0.0164 & 0.6148 & 0.8907 & 0.1740 & 0.2775 & 744 & 40.4 & 114.4 & 137.8 & 66.3 & 81.0 & 86.9 & 11.4 / 13.1 \\
    \bottomrule
    \end{tabular}
}
\end{table*}

\begin{table*}[!t]
    \centering
    \caption{\rev{Quantitative results of reconstruction with different numbers of initial planes on Hypersim~\cite{Hypersim}.}}
    \label{tab:init_planes_metrics}
    \resizebox{0.8\linewidth}{!}{
        \begin{tabular}{l|cccccc|ccccccc}
        \toprule
        Initial & \multicolumn{6}{c|}{Line-Level Metric M1} & \multicolumn{7}{c}{Line-Track Metric M2} \\
        \cmidrule(lr){2-7} \cmidrule(lr){8-14}
        Number & {ACC-L$\downarrow$} & {COMP-L$\downarrow$} & {PREC-L$\uparrow$} & {RECAL-L$\uparrow$} & {F-SCORE-L$\uparrow$} & {\#Lines$\uparrow$} & R1 & R5 & R10 & P1 & P5 & P10 & \# supports \\
        \midrule
        200 & 0.0098 & 0.9547 & 0.9284 & 0.2323 & 0.3571 & 2265 & 223.3 & 622.8 & 876.0 & 80.6 & 93.2 & 95.5 & 12.4 / 18.7 \\
        2000 & 0.0076 & 0.5269 & 0.9924 & 0.3387 & 0.4963 & 3355 & 396.3 & 886.5 & 1010.5 & 82.2 & 95.1 & 98.7 & 13.1 / 21.9 \\
        5000 & 0.0074 & 0.5685 & 0.9738 & 0.3085 & 0.4604 & 3337 & 368.2 & 873.2 & 940.1 & 81.6 & 94.9 & 98.1 & 12.9 / 21.6 \\
        10000 & 0.0072 & 0.5598 & 0.9746 & 0.3169 & 0.4701 & 3312 & 343.5 & 791.8 & 959.2 & 82.6 & 94.2 & 97.5 & 12.3 / 20.5 \\
        20000 & 0.0073 & 0.5471 & 0.9743 & 0.3232 & 0.4776 & 3614 & 377.6 & 802.3 & 969.4 & 82.0 & 94.7 & 98.3 & 11.9 / 19.7 \\
        40000 & 0.0072 & 0.5331 & 0.9948 & 0.3401 & 0.4988 & 4016 & 420.8 & 954.1 & 1213.6 & 83.3 & 95.7 & 98.9 & 12.8 / 18.4 \\
        100000 & 0.0071 & 0.5321 & 0.9956 & 0.3471 & 0.5068 & 4660 & 417.7 & 978.5 & 1343.0 & 82.6 & 95.2 & 98.9 & 12.6 / 18.2 \\
        \bottomrule
        \end{tabular}
    }
\end{table*}

\begin{table*}[!t]
    \centering
    \caption{\rev{Quantitative results of reconstruction with different pairs of loss weights on Hypersim~\cite{Hypersim}.}}
    \label{tab:loss_weights}
    \resizebox{0.8\linewidth}{!}{
    \begin{tabular}{cc|cccccc|ccccccc}
    \toprule
    \multicolumn{2}{c|}{Loss weights} & \multicolumn{6}{c|}{Line-Level Metric M1} & \multicolumn{7}{c}{Line-Track Metric M2} \\
    \cmidrule(lr){1-2} \cmidrule(lr){3-8} \cmidrule(lr){9-15}
    $\alpha_{\mathbf{\Pi}}$ & $\alpha_{L}$ & {ACC-L$\downarrow$} & {COMP-L$\downarrow$} & {PREC-L$\uparrow$} & {RECAL-L$\uparrow$} & {F-SCORE-L$\uparrow$} & {\#Lines$\uparrow$} & R1 & R5 & R10 & P1 & P5 & P10 & \# supports \\
    \midrule
    0.1 & 0.1 & 0.0146 & 0.5882 & 0.9428 & 0.2453 & 0.3804 & 1348 & 74.3 & 155.0 & 177.3 & 86.7 & 96.5 & 98.5 & 10.4 / 12.2 \\
    1 & 0.1 & 0.0127 & 0.5188 & 0.9576 & 0.2098 & 0.3282 & 842 & 51.9 & 106.3 & 121.4 & 88.4 & 96.9 & 98.9 & 10.7 / 12.2 \\
    10 & 0.1 & 0.0076 & 0.5269 & 0.9924 & 0.3387 & 0.4963 & 3355 & 396.3 & 886.5 & 1010.5 & 82.2 & 95.1 & 98.7 & 13.1 / 21.9 \\
    10 & 1 & 0.0098 & 0.5421 & 0.9784 & 0.2791 & 0.4239 & 1876 & 109.7 & 218.8 & 246.6 & 87.6 & 96.9 & 99.1 & 11.2 / 13.0 \\
    10 & 10 & 0.0078 & 0.5749 & 0.9901 & 0.3415 & 0.5003 & 6141 & 678.6 & 1487.4 & 1842.8 & 84.2 & 96.2 & 98.6 & 10.2 / 15.8 \\
    \bottomrule
    \end{tabular}
    }
\end{table*}

\subsubsection{The Design of 2D Line to 3D Plane Edge Assignment}
In the Sec.~\ref{subsec:3d line proposals}, our assignment strategy adopts both the angle between the projected plane edge and the detected line segment and the perpendicular distance from the endpoints of the projected plane edge to the detected line as two complementary metrics to identify the optimal correspondence.
Here, we conduct ablation studies considering only the angle or only the perpendicular endpoint distance, respectively, to verify the necessity of combining both strategies. Not any checkmark chosen denotes the use of the original PlanarSplatting~\cite{PlanarSplatting2024}. We only apply these different strategies during the process of per-scene optimization, and the standard step of 3D line assignment (as described in Sec.~\ref{subsec:ltb}) is finally adopted.
The quantitative results of metric M1 are reported in Table~\ref{tab:abs_theta_dist1} and the quantitative results of metric M2 are reported in Table~\ref{tab:abs_theta_dist2}.
As reported in the results, each individual assignment strategy enhances both the accuracy and completeness of the final 3D line reconstruction, while promoting the formation of longer line tracks. The joint strategy, which integrates both angular consistency and geometric proximity, yields the optimal performance.
Qualitative comparisons are provided in Fig.~\ref{fig:abs_theta_dist}.
As can be seen, constructing correspondences using only angle or only distance each improves the quality of 3D line reconstruction. Our strategy searches for 2D line and 3D edge correspondences by considering both angular and spatial distance relationships, yielding the most complete and accurate 3D line maps.

\subsubsection{The Sensitivity to Heuristics and Hyperparameters}
\label{sec:sensitivity}
\rev{We conducted an ablation study on several key hyperparameters and heuristics used in our method on the standard Hypersim dataset~\cite{Hypersim}. Both metrics of M1 and M2 are reported in all results.
The quantitative results about the impact of different plane splitting thresholds on the final line mapping are reported in Table~\ref{tab:split_gradient_metrics}.
It can be seen that the threshold for plane splitting directly affects the final number of line segments, and that extra-large or over-small thresholds both degrade reconstruction quality.
The quantitative results about the impact of the number of initial planes on the final line mapping are reported in Table~\ref{tab:init_planes_metrics}.
It can be seen that too few initial planes have a significant impact on the final number of line segments and the reconstruction quality.
The quantitative results about the impact of different settings of loss weights on the final line mapping are reported in Table~\ref{tab:loss_weights}.
It can be seen that both loss weights affect the final number of line segments, but reconstruction quality is influenced more strongly by the plane loss weight.
}

\begin{table*}[!t]
\centering
\caption{Per-scene results of visual localization on 7Scenes~\cite{7scene}. We report the median translation and rotation error in cm and degrees, as well as the pose accuracy at a 5 cm / 5 deg threshold.}
\label{tab:7scenes_loc}
    \resizebox{\linewidth}{!}{
        \begin{tabular}{l|cccc|ccc}
        \toprule
         & \multicolumn{4}{c|}{Matching-based Methods} & \multicolumn{3}{c}{Learning-based Methods} \\
        \cmidrule(lr){2-5} \cmidrule(lr){6-8}
         Scene & Hloc\textsuperscript{point}~\cite{Hloc} & PtLine\textsuperscript{point \& line}~\cite{ptline} & Limap\textsuperscript{point \& line}~\cite{limap} & \textbf{Ours\textsuperscript{point \& line}} & Pl2Map\textsuperscript{point}~\cite{bui2024pl2map} & Pl2Map\textsuperscript{point \& line}~\cite{bui2024pl2map} & \textbf{Ours\textsuperscript{point \& line}}  \\
        \midrule
        Chess & 2.4 / 0.84 / 93.0 & 2.4 / 0.85 / 92.7 & 2.5 / 0.85 / 92.3 & \textbf{2.3} / \textbf{0.82} / \textbf{93.2} & 2.0 / 0.65 / 95.5 & \textbf{1.9} / \textbf{0.63} / 96.0 & \textbf{1.9} / \textbf{0.63} / \textbf{96.2}  \\
        Fire & 2.3 / 0.89 / 88.9 & 2.3 / 0.91 / 87.9 & 2.1 / 0.84 / \textbf{95.5} & \textbf{2.0} / \textbf{0.83} / 93.4 & 2.0 / 0.81 / 93.3 & \textbf{1.9} / \textbf{0.80} / \textbf{94.0} & \textbf{1.9} / \textbf{0.80} / 93.8 \\
        Heads & 1.1 / 0.75 / 95.9 & 1.2 / 0.81 / 95.2 & 1.1 / 0.76 / \textbf{95.9} & \textbf{1.0} / \textbf{0.74} / 95.4 & 1.2 / 0.74 / 97.8 & \textbf{1.1} / \textbf{0.71} / 98.2 & \textbf{1.1} / \textbf{0.71} / \textbf{98.5} \\
        Office & 3.1 / 0.91 / 77.0 & 3.2 / 0.96 / 74.5 & \textbf{3.0} / \textbf{0.89} / 78.4 & \textbf{3.0} / 0.90 / \textbf{78.6} & 2.8 / 0.78 / 82.3 & \textbf{2.7} / 0.74 / \textbf{84.3} & \textbf{2.7} / \textbf{0.73} / 83.9 \\
        Pumpkin & 5.0 / 1.32 / 50.4 & 5.1 / 1.35 / 49.0 & \textbf{4.7} / \textbf{1.23} / \textbf{52.9} & \textbf{4.7} / 1.24 / 52.7 & 3.5 / 0.96 / 63.1 & \textbf{3.4} / 0.93 / 64.1 & 3.5 / \textbf{0.92} / \textbf{64.3} \\
        RedKitchen & 4.2 / 1.39 / 58.9 & 4.3 / 1.42 / 58.0 & \textbf{4.1} / 1.39 / 60.2 & \textbf{4.1} / \textbf{1.36} / \textbf{60.7} & 3.8 / 1.13 / 66.7 & \textbf{3.7} / \textbf{1.10} / \textbf{68.9} & \textbf{3.7} / \textbf{1.10} / 68.5 \\
        Stairs & 5.2 / 1.46 / 46.8 & 4.8 / 1.33 / 51.9 & \textbf{3.7} / 1.02 / \textbf{71.1} & 3.9 / \textbf{0.97} / 65.9 & 8.5 / 2.4 / 27.8 & 7.6 / \textbf{2.0} / 33.3 & \textbf{7.4} / 2.1 / \textbf{34.1} \\
        \bottomrule
        \end{tabular}
    }
\end{table*}

\begin{figure}[!t]
    \centering
    \includegraphics[width=\linewidth]{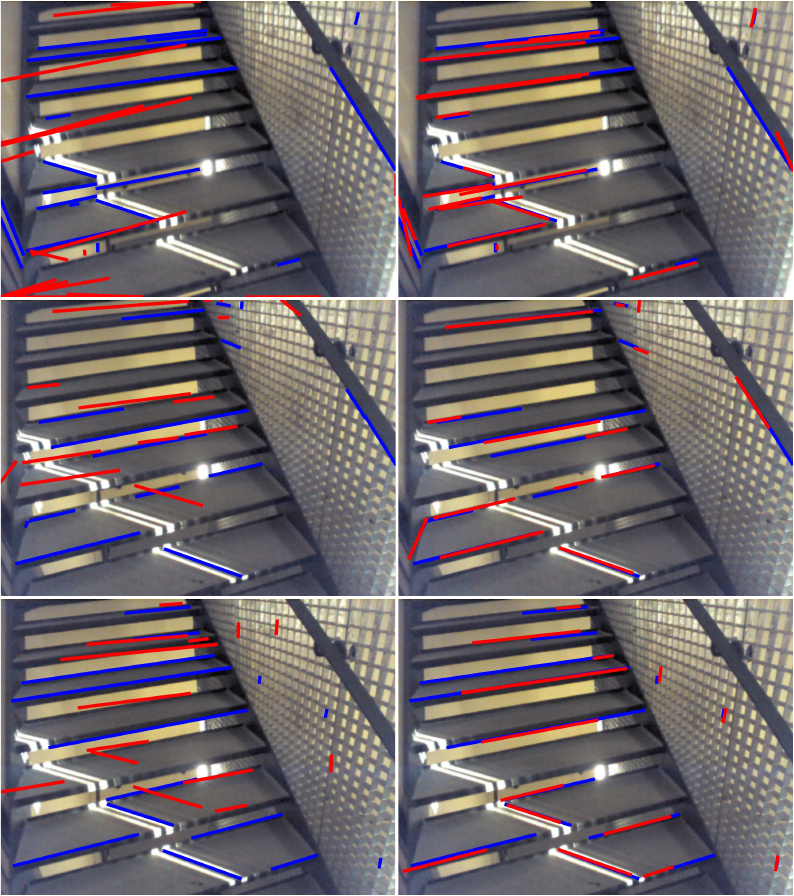}
    \caption{Line-assisted visual localization on Stairs from 7Scenes~\cite{7scene} based on \textbf{Hloc}~\cite{Hloc}. \textbf{Blue}: Detected 2D lines. \textbf{Red}: Projected 2D lines from ours 3D line map. \textbf{Left}: Projection by point-based estimated pose. \textbf{Right}: Projection by our line-assisted estimated pose.}
    \label{fig:stairs1}
\end{figure}

\begin{figure}[!t]
    \centering
    \includegraphics[width=\linewidth]{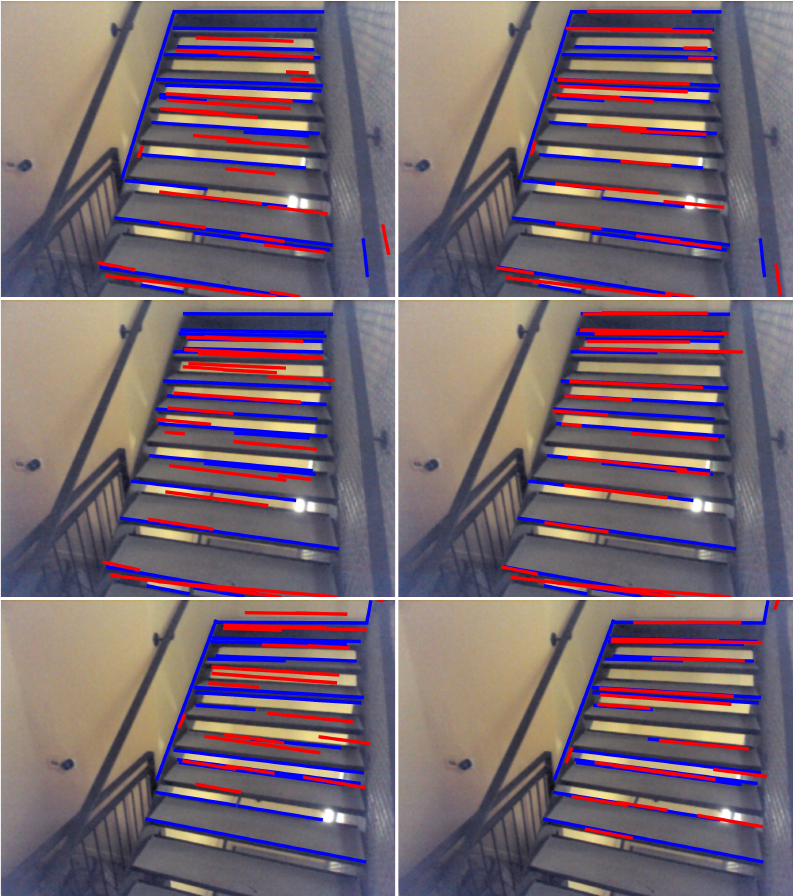}
    \caption{Line-assisted visual localization on Stairs from 7Scenes~\cite{7scene} based on \textbf{Pl2Map}~\cite{bui2024pl2map}. \textbf{Blue}: Detected 2D lines. \textbf{Red}: Projected 2D lines from ours 3D line map. \textbf{Left}: Projection by point-based estimated pose. \textbf{Right}: Projection by our line-assisted estimated pose.}
    \label{fig:stairs2}
\end{figure}

\subsection{Line-Assisted Visual Localization}
In this section, we present our experimental results for line-assisted visual localization using both point and line features on the 7Scenes dataset~\cite{7scene}.
The classic LSD~\cite{LSD} detector is adopted to detect 2D lines for 3D line mapping. We use the ground-truth depths and poses provided by the dataset and normal maps predicted by the pretrained Omnidata~\cite{omnidata-EftekharSMZ21} model for our 3D reconstruction.
The input to our visual localization pipeline is a set of 2D-3D point correspondences (from point-based methods, e.g., the matching-based SfM model or the learning-based model) and 2D-3D line correspondences (from our LiP-Map).
For the matching-based SfM model, we utilize the 2D-3D point correspondences processed by the point triangulator from COLMAP~\cite{COLMAP-SchonbergerF16} with SuperPoint~\cite{superpoint} from HLoc~\cite{Hloc}.
For the learning-based model, we employ the pretrained model from Pl2Map~\cite{bui2024pl2map} to obtain 2D–3D point correspondences.

For query images with unknown camera poses, we first use PoseLib~\cite{PoseLib} to estimate initial poses from 2D-3D point correspondences. Then, we query each image with the initial pose to generate 2D-3D line correspondences from our reconstructed 3D line map (as described in Sec.~\ref{subsec:3d line proposals} and Sec.~\ref{subsec:ltb}).
Finally, we use PoseLib~\cite{PoseLib} to estimate refined poses via both 2D-3D point correspondences and 2D-3D line correspondences.

The quantitative results for each scene are summarized in Table~\ref{tab:7scenes_loc}.
We report the median translation error (in cm), the median rotation error (in degrees), and the pose accuracy (defined as the percentage of poses with errors below 5 cm / 5 degrees) of the estimated poses over all test images of each scene.
As shown, the inclusion of our reconstructed 3D line map substantially reduces pose estimation error and improves the accuracy of point-based methods.
Compared with other line-assisted point-and-line joint pose estimation methods, our LiP-Map also achieves highly competitive performance.

To clearly demonstrate the improvement provided by our reconstructed 3D line map for visual localization, we present additional details for the most challenging "Stairs" scene from 7Scenes~\cite{7scene} in Fig.~\ref{fig:stairs1} and Fig.~\ref{fig:stairs2}. We first use the ground-truth (GT) pose to retrieve the corresponding set of 3D lines from our reconstructed model, and then respectively project these 3D lines onto the image plane using the pose estimated by the point-based method (left panel) and the line-assisted pose (right panel).
The pose estimated by the point-based method often projects 3D lines to displaced locations in the image, while the pose refined with our line-assisted pipeline achieves significantly more accurate projections, resulting in better visual consistency with the observed scene structures.

\begin{table}[!t]
\centering
\caption{\rev{The analysis of time and numbers of planes on the Hypersim dataset~[17]. All results of runtime are reported in seconds (s).}}
\label{tab:avg_time}
\resizebox{\linewidth}{!}{
\begin{tabular}{c ccc ccc cccc}
\toprule
Hypersim & \multicolumn{3}{c}{Plane Time(/iter)}
& \multicolumn{3}{c}{Line Time(/iter)}
& \multicolumn{4}{c}{Summary(60 epochs)} \\
\cmidrule(lr){2-4} \cmidrule(lr){5-7} \cmidrule(lr){8-11}
Dataset
& Rasterizer & Loss & Total
& Assignment & Loss & Total
& Average & Total & Init. \#Planes & Final \#Planes \\
\midrule
ai\_001\_001 & 0.0036 & 0.0029 & 0.0067 & 0.0221 & 0.0009 & 0.0241 & 5.3799 & 322.7921 & 2000 & 12900 \\
ai\_001\_002 & 0.0036 & 0.0029 & 0.0066 & 0.0206 & 0.0010 & 0.0226 & 5.1115 & 306.6906 & 1955 & 7448 \\
ai\_001\_003 & 0.0032 & 0.0028 & 0.0061 & 0.0192 & 0.0009 & 0.0211 & 4.9280 & 295.6779 & 1640 & 6337 \\
ai\_001\_004 & 0.0041 & 0.0036 & 0.0078 & 0.0263 & 0.0010 & 0.0285 & 6.2292 & 373.7514 & 1897 & 19704 \\
ai\_001\_005 & 0.0034 & 0.0031 & 0.0067 & 0.0223 & 0.0009 & 0.0242 & 5.7705 & 346.2297 & 1834 & 13549 \\
ai\_001\_006 & 0.0040 & 0.0035 & 0.0076 & 0.0230 & 0.0010 & 0.0252 & 5.9670 & 358.0211 & 1767 & 6384 \\
ai\_001\_007 & 0.0038 & 0.0032 & 0.0072 & 0.0224 & 0.0010 & 0.0245 & 5.6613 & 339.6774 & 1954 & 6888 \\
ai\_001\_008 & 0.0036 & 0.0033 & 0.0070 & 0.0211 & 0.0010 & 0.0231 & 5.3834 & 323.0037 & 1794 & 6608 \\
ai\_001\_009 & 0.0037 & 0.0029 & 0.0067 & 0.0214 & 0.0010 & 0.0235 & 5.6190 & 337.1397 & 1946 & 7105 \\
ai\_001\_010 & 0.0033 & 0.0034 & 0.0068 & 0.0222 & 0.0009 & 0.0240 & 5.4591 & 327.5479 & 1943 & 13718 \\ \hline
Average     & 0.0036 & 0.0032 & 0.0069 & 0.0221 & 0.0010 & 0.0241 & 5.5509 & 333.0532 & 1873 & 10064 \\
\bottomrule
\end{tabular}
}
\end{table}

\begin{figure}[!t]
\centering
\includegraphics[width=\linewidth]{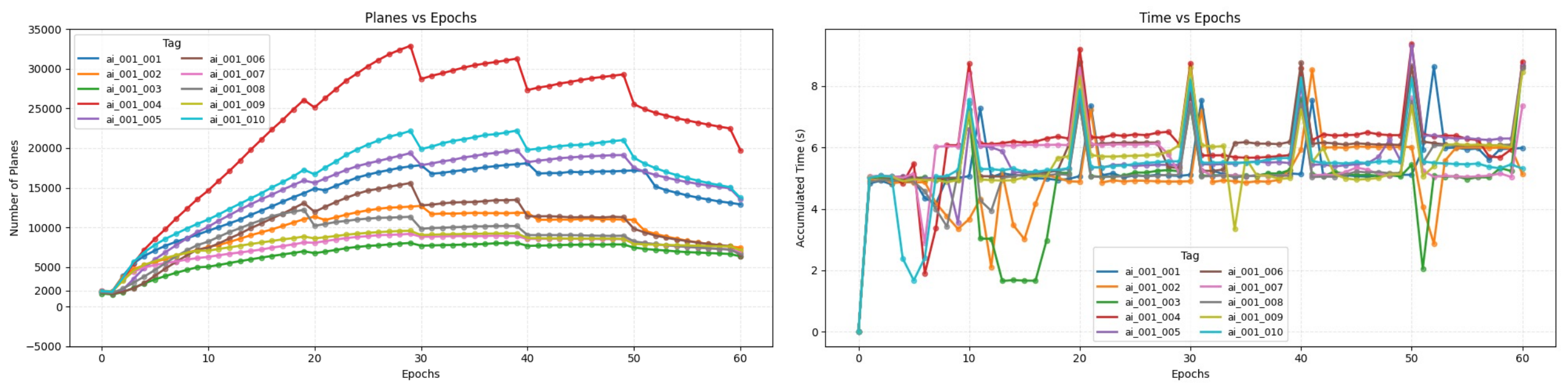}
\vspace{-3mm}
\caption{\rev{The analysis of time and numbers of planes on the Hypersim dataset~[17]. We report how the number of planes and the optimization time evolve over optimization epochs.}}
\label{fig:opt_details}
\end{figure}

\begin{figure*}[!t]
\centering
\includegraphics[width=\linewidth]{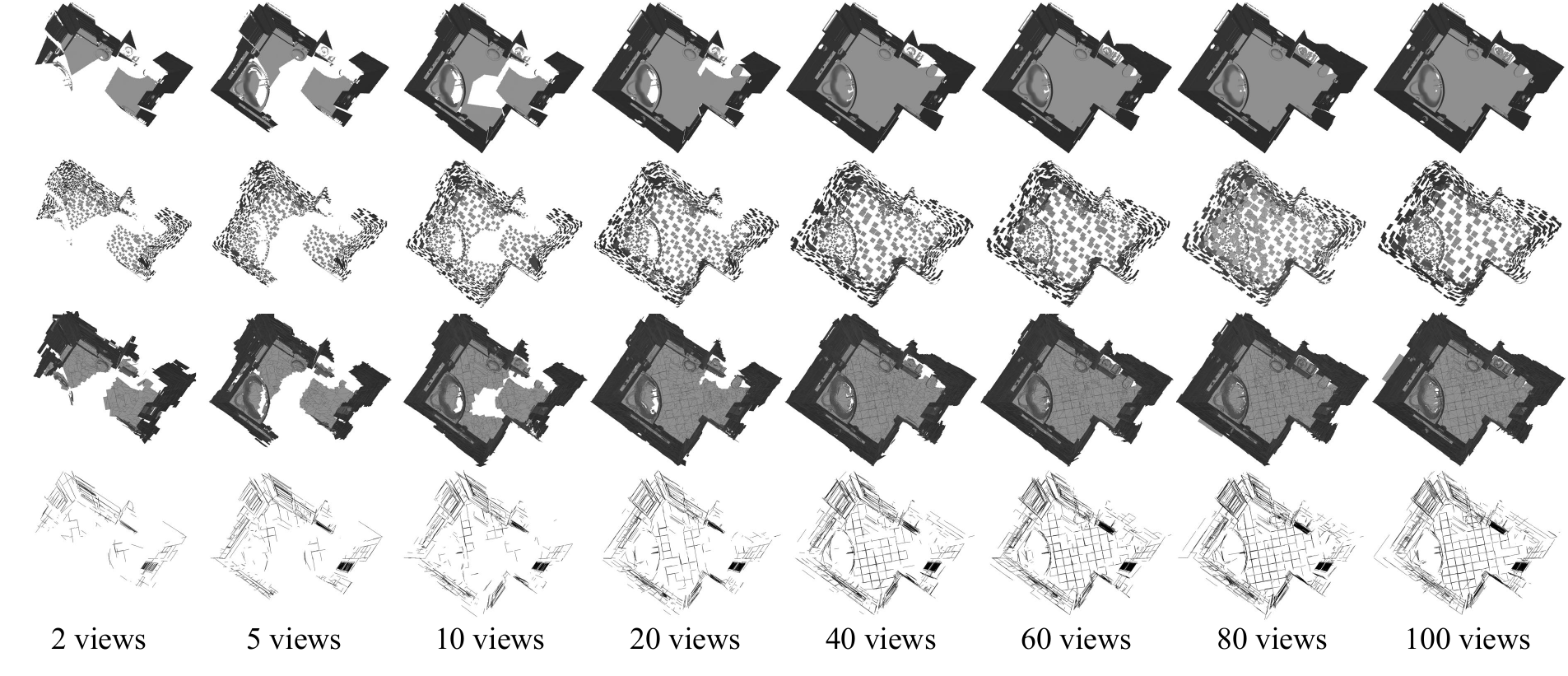}
\vspace{-3mm}
\caption{\rev{Qualitative comparison of results with different numbers of input views. First row: initial mesh. Second row: initial planar mesh. Third row: final planar mesh. Last row: final 3D lines.}}
\label{fig:input_views}
\end{figure*}

\begin{figure*}[!]
\centering
\includegraphics[width=\linewidth]{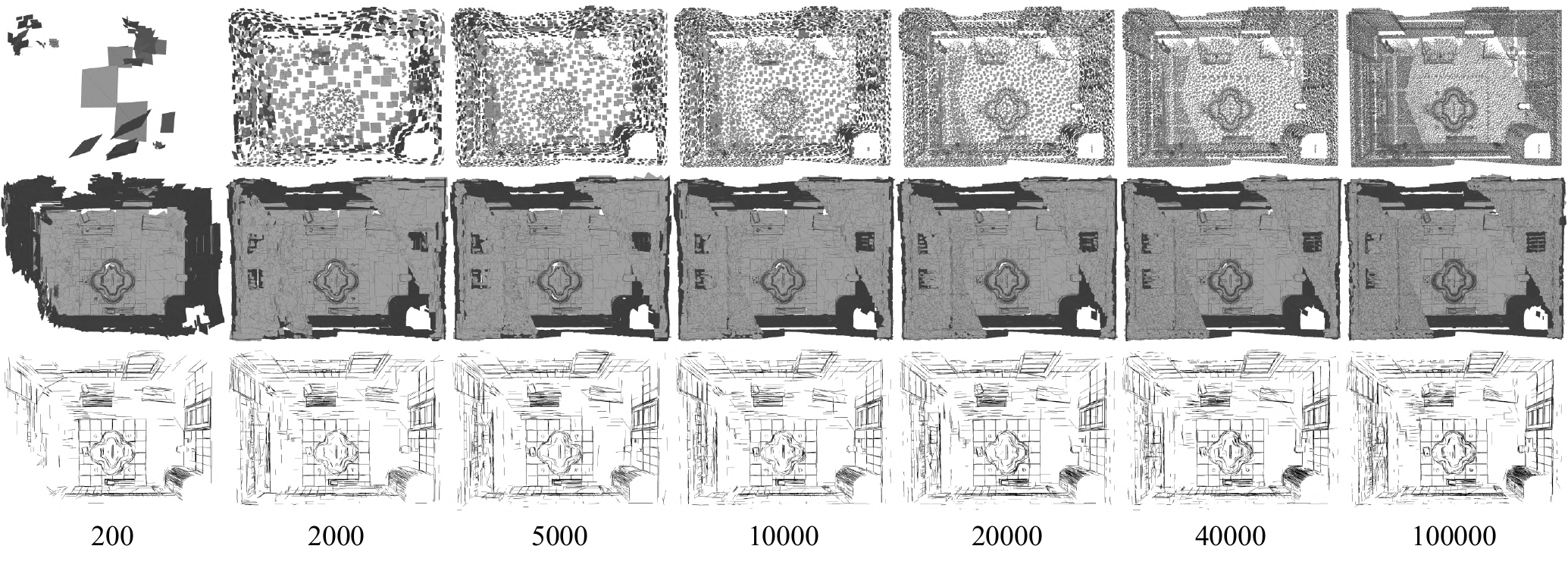}
\vspace{-3mm}
\caption{\rev{Qualitative comparison of results with different numbers of initial planar primitives. First row: initial planar mesh. Second row: final planar mesh. Last row: final 3D lines.}}
\label{fig:plane_num}
\end{figure*}

\subsection{Analysis of Efficiency and Robustness}
\rev{
In this section, we analyze the efficiency and robustness of our optimization framework on the first ten scenes of Hypersim~\cite{Hypersim}, using the DeepLSD~\cite{DeepLSD} detector, on a single NVIDIA RTX A6000 with 48GB of memory.}

\paragraph{Details of per-scene optimization.}
\rev{We first fix the number of initial planar primitives to 2000 for each scene, use 100 input views, and optimize for 60 epochs.
We provide a detailed breakdown of the average runtime cost of the two core modules during each iteration and report the total runtime of optimization and the per-epoch runtime for each scene in Table~\ref{tab:avg_time}.
The reported ``Init. \#Planes'' denotes the number of planar primitives remaining after the initial pruning step, which removes planes that are not visible under any viewpoints.
The reported ``Final. \#Planes'' denotes the number of planar primitives after the optimization converges.
The number of planes and the runtime at every epoch for each scene are reported in Fig.~\ref{fig:opt_details}.}

\begin{table}[!t]
\centering
\caption{\rev{Optimization time and number of planes versus the number of input views.}}
\label{tab:input_views}
\resizebox{\linewidth}{!}{
\begin{tabular}{l| cccccccc}
\toprule
Input Views & 2 & 5 & 10 & 20 & 40 & 60 & 80 & 100 \\
\midrule
Iterations & 1000 & 1000 & 2000 & 2000 & 3000 & 3000 & 4000 & 5000  \\
Time(s) & 32.6058 & 35.3922 & 101.3842 & 104.9645 & 165.1081 & 162.7495 & 221.6956 & 275.3702  \\
Init. \#Planes & 1982 & 1947 & 1929 & 1925 & 1865 & 1890 & 1874 & 1875 \\
Final. \#Planes & 2011 & 3065 & 4571 & 6102 & 7622 & 7747 & 8018 & 8676  \\
\bottomrule
\end{tabular}
}
\end{table}

\begin{table}[!t]
\centering
\caption{\rev{Optimization time and number of planes versus the initial number of planar primitives.}}
\label{tab:init_planes}
\resizebox{\linewidth}{!}{
\begin{tabular}{l| ccccccc}
\toprule
Numbers & 200 & 2000 & 5000 & 10000 & 20000 & 40000 & 100000 \\
\midrule
Iterations & 6000 & 6000 & 6000 & 6000 & 6000 & 6000 & 6000 \\
Time(s) & 331.9121 & 333.0532  & 345.8118 & 346.6410 & 356.7424 & 391.7404 & 443.6078  \\
Init. \#Planes & 184 & 1873 & 4811 & 9676 & 19315 & 38399 & 95517  \\
Final. \#Planes & 7692 & 10064 & 11677 & 12579 & 14810 & 18940 & 27961  \\
\bottomrule
\end{tabular}
}
\end{table}

\paragraph{Scalability with the number of input views.}
\rev{We first investigate how the optimization time and the number of planes scale with the number of input views in Table~\ref{tab:input_views}.
It should be noted that when the input views are too sparse (e.g., only 2 views), 60 epochs (120 iterations) are not sufficient for reconstruction, so we run more iterations (e.g., 1000) to achieve high-quality results.
In practice, the total runtime is dominated by the number of optimization iterations and is partly influenced by the number of planar primitives.
The qualitative comparison of results is illustrated in Fig.~\ref{fig:input_views}, taking the `ai\_001\_001' scene as an example.
It can be seen that our method is robust to the reconstruction across different numbers of input views.}

\paragraph{Scalability with the number of planes.}
\rev{ We also investigate how the optimization time and the number of planes scale with the initial number of planar primitives.
As reported in Table~\ref{tab:init_planes}, the optimization time does not increase sharply even when the number of initial planes rises substantially, and the final number of planes after convergence remains within a relatively stable range.
The qualitative comparison of results are visualized in Fig.~\ref{fig:plane_num}, taking the `ai\_001\_007' scene as an example.
These results indicate that our method is robust to the number of planes used for initialization.
Notably, although initializing with 200 planar primitives yields a poor initial planar mesh at the beginning, the optimization can still converge to a relatively complete reconstruction in the end.}

\subsection{Generalization to Non-Planar Structures}
\begin{figure*}[htb!]
\centering
\includegraphics[width=\linewidth]{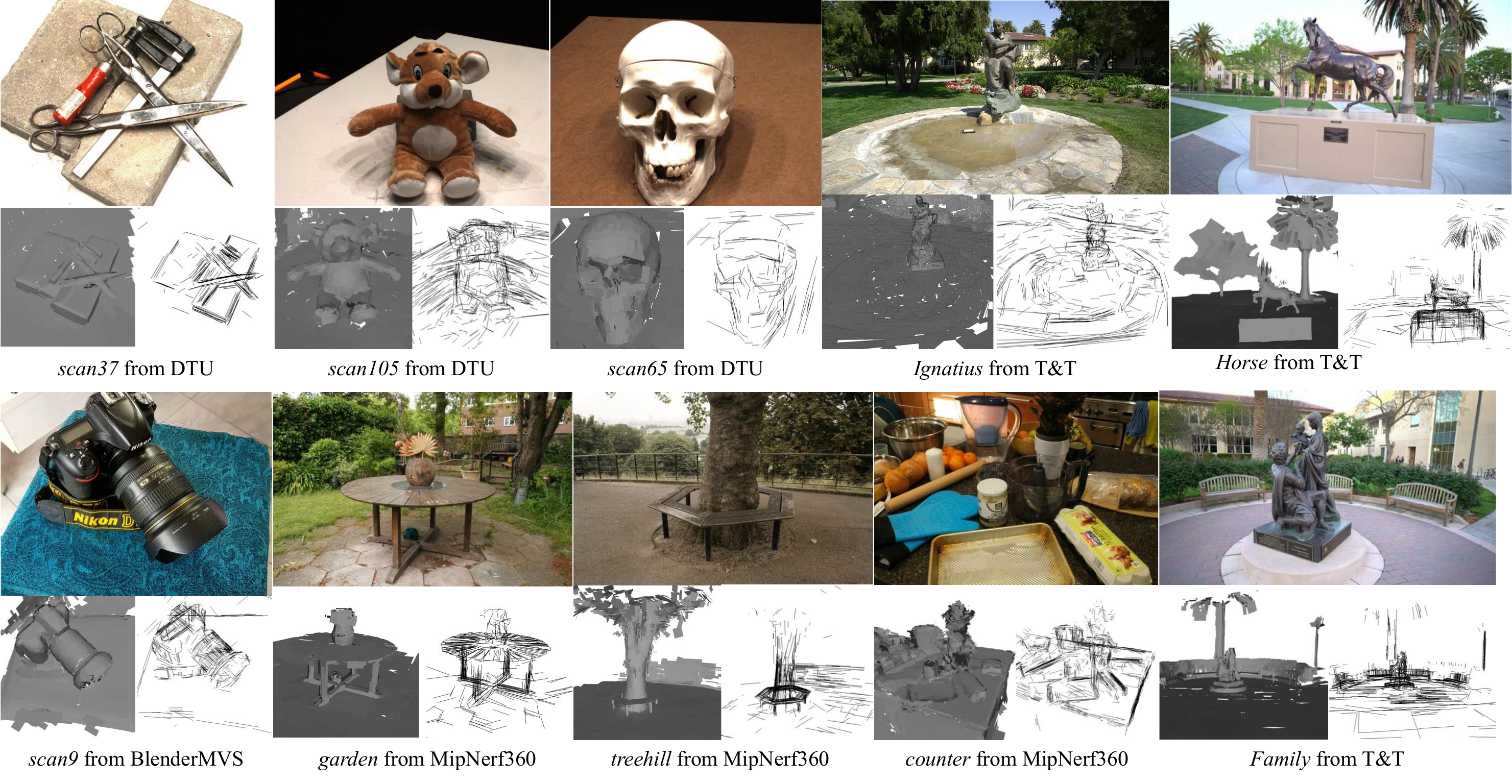}
\vspace{-3mm}
\caption{\rev{Qualitative results of reconstructed planar mesh and 3D line maps on non-planar structural scenes.}}
\label{fig:non-planar}
\end{figure*}

\rev{To further investigate the generalization ability of our method on generic non-planar structures, we conducted experiments using only multi-view images from scenes in some common datasets, including DTU~\cite{DTU}, Tanks and Temples~\cite{tanks_temples}, BlenderMVS~\cite{BlenderMVS}, and MipNerf360~\cite{mipnerf360}.
To explore this capability as thoroughly as possible, we use noisy poses, depth maps, and normal maps predicted by VGGT for optimization.
As shown in Fig.~\ref{fig:non-planar}, these scenes include non-planar and irregular structures such as curved chairs, circular tables, trees, slender fence railings, street lamps, sculptures, and other objects with complex geometry.
But our method can largely meet the reconstruction needs of these scenes by organizing and optimizing the topological relationships among a large number of planes.
Although some planes in the reconstructed planar surface may be inaccurate, the strict thresholds in our assignment strategy (as described in Sec.~\ref{subsec:ltb}) enforce strong geometric consistency between 2D and 3D lines, allowing these inaccurate plane edges to be filtered out.
}

\subsection{More Results with VGGT}
\begin{figure}[!t]
    \centering
    \includegraphics[width=\linewidth]{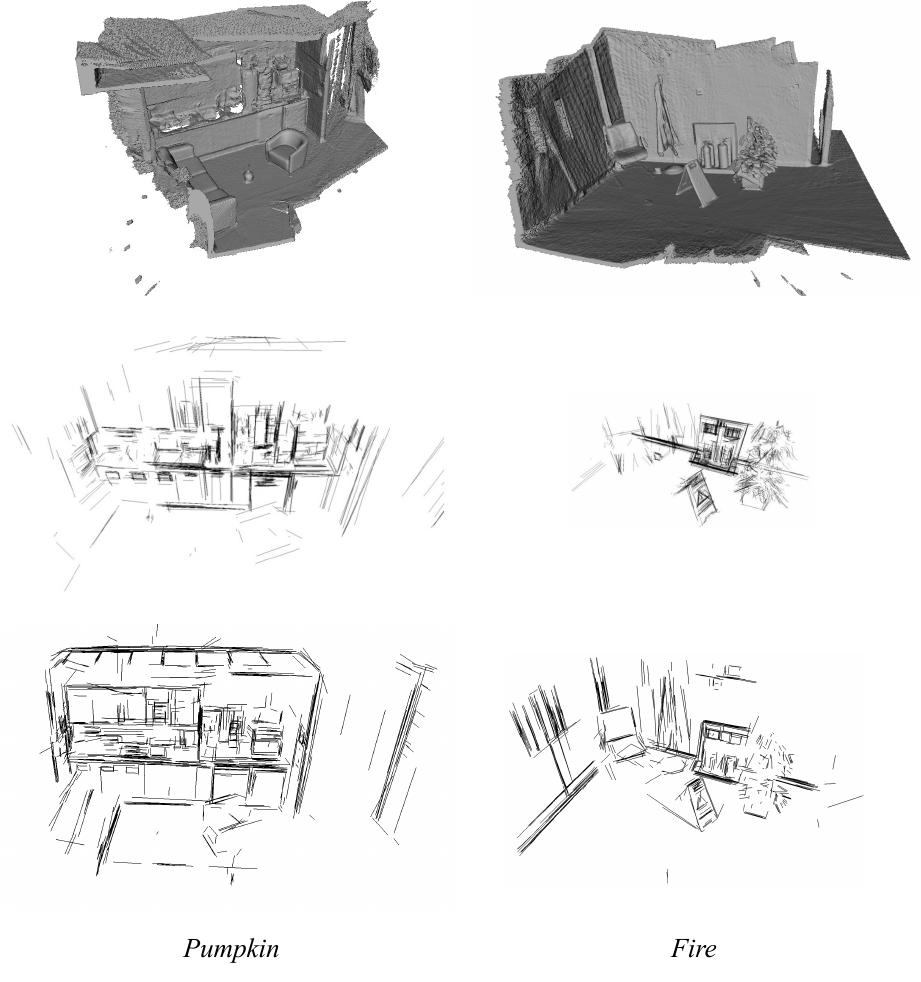}
    \caption{Qualitative comparison with Pl2Map~\cite{bui2024pl2map} on 7scenes~\cite{7scene}. Top: the ground-truth mesh. Middle: Pl2Map with ground-truth. Bottom: Our results with VGGT.}
    \label{fig:7scene_compar}
\end{figure}

\begin{figure}[!t]
    \centering
    \includegraphics[width=\linewidth]{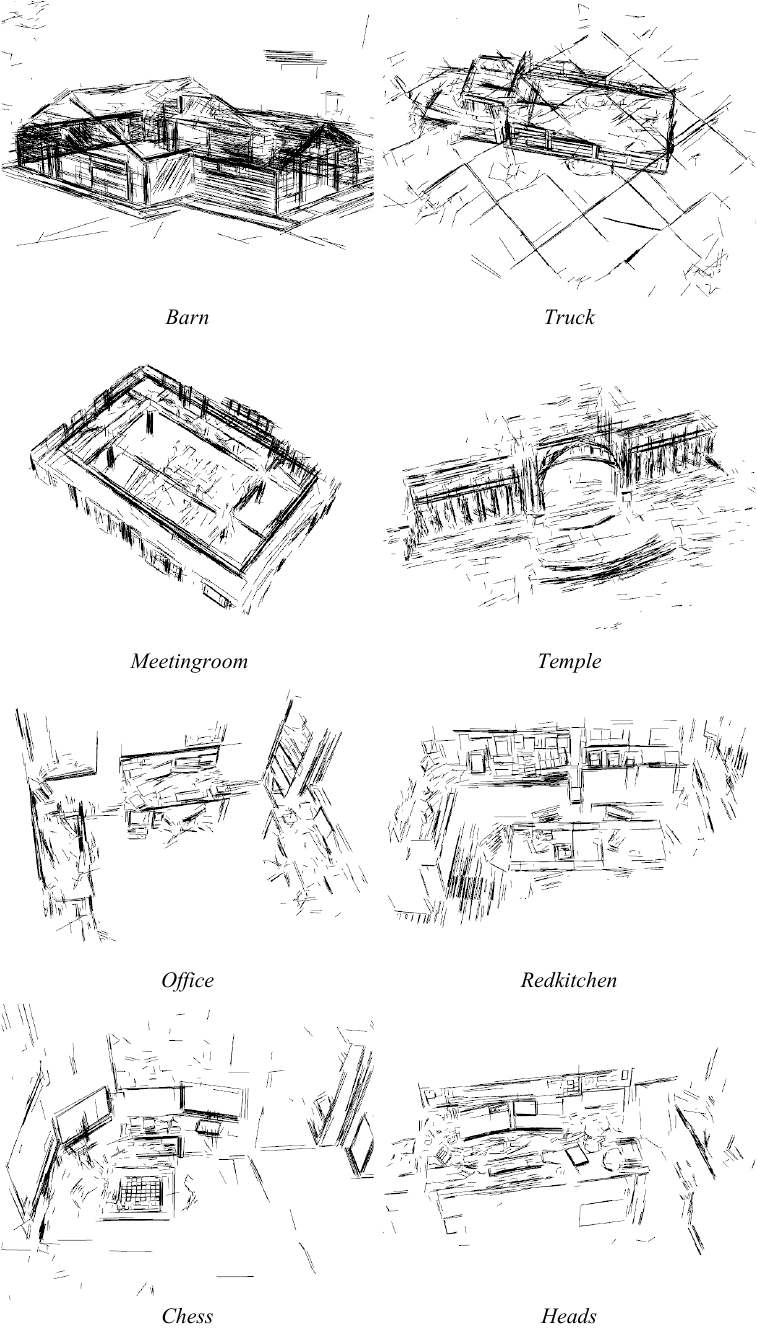}
    \caption{Qualitative results of our 3D line maps on Tanks and Temples~\cite{tanks_temples} and 7scene~\cite{7scene}. ``Barn'', ``Truck'', ``Meetingroom'' and ``Temple'' are from the Tanks and Temples dataset. ``Office'', ``Redkitchen'', ``Chess'' and ``Heads'' are from the 7scenes dataset.}
    \label{fig:7scene_compar2}
\end{figure}

We additionally conduct visual experiments on the 7Scenes dataset~\cite{7scene} and the Tanks and Temples dataset~\cite{tanks_temples}. Different from using the ground-truth poses and depth maps provided by the dataset, we apply the recent VGGT~\cite{wang2025vggt} to regress poses, depth maps, and normal maps for optimization. This is a more challenging setting to reconstruct 3D lines with imperfect poses and depth maps.
Since all hyperparameters in our optimization process are scale-invariant, no modifications are required when integrating VGGT, as the overall framework remains compatible under scale variations.
We present a visual comparison of reconstruction results with Pl2Map~\cite{bui2024pl2map} in Fig.~\ref{fig:7scene_compar}.
Our method can also achieve more complete and accurate 3D line maps, even with outputs from VGGT.
We also show some additional results of our 3D line maps on the 7Scenes dataset~\cite{7scene} and the Tanks and Temples dataset~\cite{tanks_temples} in Fig.~\ref{fig:7scene_compar2}.

\section{Conclusion}\label{sec:conclusion}
In this paper, we present \method, a line-plane joint optimization framework for 3D line mapping. Our method leverages 2D primal sketches of line segments and 2.5D sketches of depth and normal representations from posed multi-view images to optimize the geometric synergy between 3D planes and 3D line segments. Through an online optimization process, we simultaneously achieve high-performance 3D line mapping alongside 3D plane reconstruction with prominent topological structure.
Our proposed \method produces accurate, complete, and detailed 3D line mapping, seamlessly integrating with any 2D line segment detector, including both classical (e.g., LSD) and learning-based approaches. By bypassing the challenging 2D line segment matching step, our method significantly outperforms state-of-the-art approaches on both real-world and synthetic indoor scene benchmarks.
We hope and believe that our structured 3D reconstruction approach will contribute to compact spatial representations, shaping the next era of spatial intelligence in the near future.

\ifCLASSOPTIONcaptionsoff
  \newpage
\fi


\bibliographystyle{IEEEtran}
\bibliography{IEEEabrv,main}

\clearpage
\appendices
\section{More Implementation Details} \label{appdA}
\paragraph{Ray-to-Plane Intersection.}
We first calculate the intersections between planar primitives and the rays emitted from pixels of the image. Specifically, given a ray $\mathbf{r} = \{\mathbf{o}, \mathbf{d}\}$ starting from the camera center $\mathbf{o} \in \mathbb{R}^3$ with direction $\mathbf{d} \in \mathbb{R}^3$, its intersection $\mathbf{x}_{\pi}^{\mathbf{r}} \in \mathbb{R}^3$ to one planar primitive $\pi$ can be calculated as:
\begin{equation}\label{eq:ray-plane-intersection}
    \mathbf{x}_{\pi}^{\mathbf{r}} = \mathbf{o} + \frac{(\mathbf{p}_{\pi} - \mathbf{o} \cdot \mathbf{n}_{\pi})}{\mathbf{d} \cdot \mathbf{n}_{\pi}} \mathbf{d},
\end{equation}
where $\mathbf{p}_{\pi}$ and $\mathbf{n}_{\pi}$ are the center and the normal of the planar primitive $\pi$.

\paragraph{Plane Splatting Function.}
After achieving the ray-to-plane intersection $\mathbf{x}_{\pi}^{\mathbf{r}}$, we then calculate its splatting weight with our plane splatting function, which will be used for rendering.

For a given intersection $\mathbf{x}_{\pi}^{\mathbf{r}}$ between the ray $\mathbf{r}$ and the planar primitive $\pi$, we first calculate its projection distance $\mathcal{P}_{X}, \mathcal{P}_{Y} \in \mathbb{R}$ to the X-axis and Y-axis of the planar primitive as:
\begin{equation}
    \mathcal{P}_{X} = (\mathbf{x}_{\pi}^{\mathbf{r}}-\mathbf{p}_{\pi})\cdot \mathbf{v}^x_{\pi}, \quad \mathcal{P}_{Y} = (\mathbf{x}_{\pi}^{\mathbf{r}}-\mathbf{p}_{\pi})\cdot \mathbf{v}^y_{\pi}.
\end{equation}
Then, we calculate the splatting weight along the X-axis of the plane $\pi$ as:
\begin{equation}
    w_X(\mathbf{x}_{\pi}^{\mathbf{r}}) =
    \begin{cases}
        2\sigma(5\lambda (r_{\pi}^{x+} -|\mathcal{P}_X| )), \quad \text{if}~\mathcal{P}_{X} > 0\\
        2\sigma(5\lambda (r_{\pi}^{x-} -|\mathcal{P}_X| )), \quad \text{otherwise}
    \end{cases},
\end{equation}
where $\sigma(\cdot)$ is the Sigmoid function and $\lambda$ is the hyperparameter to control the splatting weight. Similarly, we continue to calculate the splatting weight along the Y-axis of the plane $\pi$ as:
\begin{equation}
    w_Y(\mathbf{x}_{\pi}^{\mathbf{r}}) =
    \begin{cases}
        2\sigma(5\lambda (r_{\pi}^{y+} -|\mathcal{P}_Y| )), \quad \text{if}~\mathcal{P}_{Y} > 0\\
        2\sigma(5\lambda (r_{\pi}^{y-} -|\mathcal{P}_Y| )), \quad \text{otherwise}
    \end{cases},
\end{equation}
where $r_{\pi}^{x+}, r_{\pi}^{x-}, r_{\pi}^{y+}, r_{\pi}^{y-}$ are the radii parameters of the plane $\pi$.
The value of $\lambda$ is increased with an exponential function during optimization up to the maximum value of 300 as:
\begin{equation}
    \lambda = min(20e^{(-( 1 - 0.001 * ite))}, 300),
\end{equation}
where $ite$ means the iteration number during optimization.
At last, the final splatting weight can be calculated as:
\begin{equation}\label{eq:plane-rbf}
    w(\mathbf{x}_{\pi}^{\mathbf{r}}) = \begin{cases}
        w_X , \quad  \text{if}~w_X < w_Y \\
        w_Y, \quad \text{otherwise}
    \end{cases}.
\end{equation}

\paragraph{Planar Blending Composition.}
For all ray-to-plane intersections, we filter them with a splatting weight lower than 0.0001 and then sort the remaining intersections according to their depth from near to far. Then, $M$ nearest intersections of each ray are selected to render ($M=5$ in this paper). Denote the selected intersections of a ray $\mathbf{r}$ as $P^{\mathbf{r}}=\{ \mathbf{x}_{\pi_{\tau(j)}}^{\mathbf{r}} \}_{j=1}^{M}$. Here, $\tau(j)$ indicates the index of the plane among all planar primitives. Finally, we render the depth and normal map of a certain image $\mathbf{I}$ as:
\begin{equation}
    \mathbf{D}_{\text{render}}^{\Pi}(\mathbf{r}) = \sum_{j=1}^{M} T_j w(\mathbf{x}_{\pi_{\tau(j)}}^{\mathbf{r}}) t_j,
\end{equation}

\begin{equation}
    \mathbf{N}_{\text{render}}^{\Pi}(\mathbf{r}) = \sum_{j=1}^{M} T_j w(\mathbf{x}_{\pi_{\tau(j)}}^{\mathbf{r}}) \mathbf{n}_{\pi_{\tau(j)}},
\end{equation}
where
\begin{equation}
    T_j = \prod_{i=1}^{j-1}(1-w(\mathbf{x}_{\pi_{\tau(i)}}^{\mathbf{r}})).
\end{equation}
Here, $t_j$ is the depth of the intersection and $\mathbf{n}_{\pi_{\tau(j)}}$ is the normal of the planar primitive $\pi_{\tau(j)}$.
To supervise the rendered depth and normal map, we use the ground-truth depth/normal maps of the given dataset or the depth/normal maps
predicted by the pretrained model of Metric3Dv2~\cite{Metric3Dv2} and Omnidata~\cite{omnidata-EftekharSMZ21} to serve as pseudo labels.
The depth and normal maps used for supervision are denoted as $\mathbf{D}_{\text{pre}}(\mathbf{r})$ and $\mathbf{N}_{\text{pre}}(\mathbf{r})$.
Finally, the render loss can be calculated as:
\begin{equation}
\begin{split}
    \mathcal{L}_{\text{render}}^{\Pi} = & \alpha_1 \sum_{\mathbf{r} \in \mathbf{I}} \| 1-\mathbf{N}_{\text{render}}^{\Pi}(\mathbf{r})^{\top}\mathbf{N}_{\text{pre}}(\mathbf{r})) \|_1 + \\
    & \alpha_1 \sum_{\mathbf{r} \in \mathbf{I}} \| \mathbf{N}_{\text{render}}^{\Pi}(\mathbf{r}) - \mathbf{N}_{\text{pre}}(\mathbf{r})) \|_1 + \\
    & \alpha_2 \sum_{\mathbf{r} \in \mathbf{I}} \| \mathbf{D}_{\text{render}}^{\Pi}(\mathbf{r}) - \mathbf{D}_{\text{pre}}(\mathbf{r})\|_1,
\end{split}
\end{equation}
where $\alpha_1=5.0$, $\alpha_2=1.0$, $\alpha_3=2.0$, $\textbf{r}$ is the ray emitted from pixel of image $\mathbf{I}$.

\begin{figure}[!t]
    \centering
    \includegraphics[width=\linewidth]{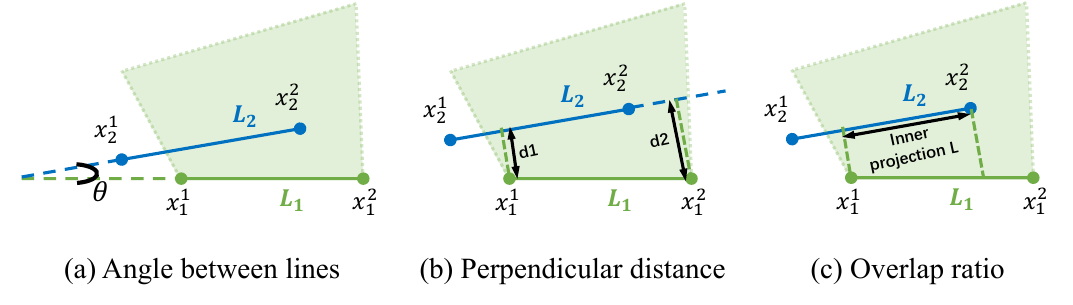}
    \caption{Geometric consistency measures. We propose three scale-invariant measures for ensuring the geometric consistency between 3D planar edges and 2D detected lines.}
    \label{fig:geometric_cons}
\end{figure}

\section{Geometric Consistency Measures.} \label{appd:geometric_cons}
We have defined some geometric constraints in our paper and applied these constraints to extract the final 3D line maps and to build line tracks.
Here, we introduce the computation of these geometric constraints and give a toy example in Fig.~\ref{fig:geometric_cons} for illustration. Assume the green region is projected from the 3D planar primitive, $(x_1^1,x_1^2)$ are the endpoints of the projected 2D line $L_1$ from the 3D planar edge, and $(x_2^1,x_2^2)$ are the endpoints of the detected 2D line $L_2$.
We compute three geometric measures as follows:
\begin{itemize}
    \item The angle distance $d_{ang}$ (denoted as $\theta$) between the projected 2D lines and each detected 2D line. As shown in Fig.~\ref{fig:geometric_cons}(a), the $\theta$ is computed as:
    \begin{equation}
        \theta = \arccos \left( \operatorname{clamp}(\frac{\left \| \left ( x_{1}^{2}-x_{1}^{1} \right ) \cdot \left ( x_{2}^{2}-x_{2}^{1} \right ) \right \|}{\left \| \left ( x_{1}^{2}-x_{1}^{1} \right ) \right \| } ,-1,1) \right).
    \end{equation}
    The angle distance used in Sec.~\ref{subsec:3d line proposals} and Sec.~\ref{subsec:ltb} is also computed in this way.

    \item The maximum orthogonal distance $d_{dist}$ from the endpoints of the projected 2D line to the detected 2D line. As shown in Fig.~\ref{fig:geometric_cons}(b), the $d_{\rm dist}$ is computed as:
    \begin{equation}
        d_{\rm dist} = \max \left(d_1, d_2 \right), d_i=\frac{\left \| \left ( x_{1}^{i}-x_{2}^{1} \right ) \times \left ( x_{1}^{i}-x_{2}^{2} \right ) \right \|}{\left \| \left ( x_{2}^{2}-x_{2}^{1} \right ) \right \| }.
    \end{equation}
    The orthogonal distance $d_{dist}$ defined in Sec.~\ref{subsec:ltb} is also computed in this way.

    \item The overlap ratio $d_{overlap}$ between the projected 2D line and the detected 2D line. As shown in Fig.~\ref{fig:geometric_cons}(C), the $d_{overlap}$ is computed as:
    \begin{equation}
        d_{overlap} = \frac{\left \| L \right \|}{\left \| L_2 \right \|},
    \end{equation}
    where $L$ is the inner projection portion of $L_1$ on $L_2$. The overlap ratio used in Sec.~\ref{subsec:ltb} is computed in this way.
\end{itemize}

\section{Computation of Evaluation Metrics} \label{appd:metric}

Given the ground-truth model (mesh or point cloud) and the reconstructed 3D lines, we introduce how to compute the widely-used standard metric M1 and the line tracks-based metric M2.
The basic step is to calculate the closest distance between the 3D line segments and the model.
To this end, we first uniformly sample points along each 3D line segment and then build a KD-Tree over the ground-truth model or the sampled points for efficient computation of the distance between the sampled points and the ground-truth model.

\paragraph{Metric M1.}
To construct the input points $P_{pre}$, we uniformly sample 100 points along each 3D line segment for line-level metrics and collect the two endpoints of each 3D line segment for junction-level metrics. Given the ground-truth point cloud $P_{gt}$, we build the KD-tree as:
\begin{equation}
\begin{split}
    & {\rm KDT}_{pre} = {\rm KDTree}(P_{pre}), \\
    & {\rm KDT}_{gt} = {\rm KDTree}(P_{gt}),
\end{split}
\end{equation}
where algorithm $\rm KDTree$ is adopted from the library Scikit-learn~\cite{scikit-learn}.
Then we query the ${\rm KDT}_{gt}$ with points $P_{pre}$ and query the ${\rm KDT}_{pre}$ with point cloud $P_{gt}$ to get the distance:
\begin{equation}
    \begin{split}
        & D_{pre} = {\rm KDT}_{gt}(P_{pre}), \\
        & D_{gt} = {\rm KDT}_{pre}(P_{gt}).
    \end{split}
\end{equation}
The accuracy and completeness are computed by the following:
\begin{equation}
    \begin{split}
        & {\rm ACC} = \frac{1}{n_1}\sum_{i=1}^{n_1}D_{gt}^{i}, \\
        & {\rm COMP} = \frac{1}{n_2}\sum_{i=1}^{n_2}D_{pre}^{i},
    \end{split}
\end{equation}
where $n_1$ is the number of points in the group of points $P_{pre}$ and $n_2$ is the number of points in the point cloud $P_{gt}$.
Given the distance threshold $\tau_d$ (0.05m in our paper), the precision, recall, and F1 score are computed by:
\begin{equation}
    \begin{split}
        & {\rm PREC} = \frac{1}{n_1}\sum_{i=1}^{n_1}\mathbf{1}(D_{pre}^{i}<=\tau_d), \\
        & {\rm RECALL} = \frac{1}{n_2}\sum_{i=1}^{n_2}\mathbf{1}(D_{gt}^{i}<=\tau_d), \\
        & {\rm F1} = 2 \cdot  \frac{{\rm PREC} \cdot {\rm RECALL}}{{\rm PREC} + {\rm RECALL}}
    \end{split}
\end{equation}
where $\mathbf{1}(\cdot)$ is the indicator function.

\paragraph{Metric M2.}
Given the reconstructed 3D lines, following LIMAP~\cite{limap}, we uniformly sample 1000 points along each 3D line segment and build the KD-tree to obtain the closest distances $D_{pre}$.
Given the threshold $\tau_d$ of distance (1/5/10 mm or 5/10/50 mm in our paper), the proportion of each 3D line segment that has points with distance less than $\tau_d$ is calculated as:
\begin{equation}
    {\rm ratios_i} = \frac{1}{n}\sum_{j=1}^{n} \mathbf{1}(D_{pre}^{(i,j)}<=\tau_d)
\end{equation}
where $n=1000$ is the number of sampled points of each line.
The sum of the lengths of the line portions within $\tau_d$ mm from the GT model is computed as:
\begin{equation}
    R_{\tau_d} = \sum_{i=1}^{m} {\rm Len}_i \cdot {\rm ratios_i},
\end{equation}
where ${\rm Len}_i$ is the length of the i-th 3D line.
The percentage of tracks that are within $\tau_d$ mm from the GT model is computed as:
\begin{equation}
    P_{\tau_d} = 100 \cdot \frac{1}{m} \sum_{i=1}^{m} \mathbf{1}({\rm ratios}_i>0).
\end{equation}

\section{Global Line Merging}
Our pipeline may inherit errors from the 2D line detector, which often manifest as fragmented detections, duplicated line hypotheses, or slight endpoint misalignments across views. As a lightweight remedy, we provided a global merging strategy to consolidate redundant 3D line segments while preserving geometric consistency.

\noindent\textbf{Baseline and local merging:} Taking our original method without merging as baseline, our local merging strategy groups 3D lines corresponding to the same detected line and merges them into a single line segment.

\noindent\textbf{Global merging:} Our global merging strategy uses DBSCAN to cluster 3D lines associated with each 2D detected line across all views, propagates unique identifiers through these clusters to establish global correspondences, and merges lines sharing the same identifier using PCA-based alignment to determine the final merged line segments. The algorithm is shown in Algorithm~\ref{alg:global_merge}. We provide the corresponding implementation in our code repository and the threshold is set to $\tau_{dbscan}=0.01$ in our implementation.

\begin{figure*}[!t]
\centering
\includegraphics[width=\linewidth]{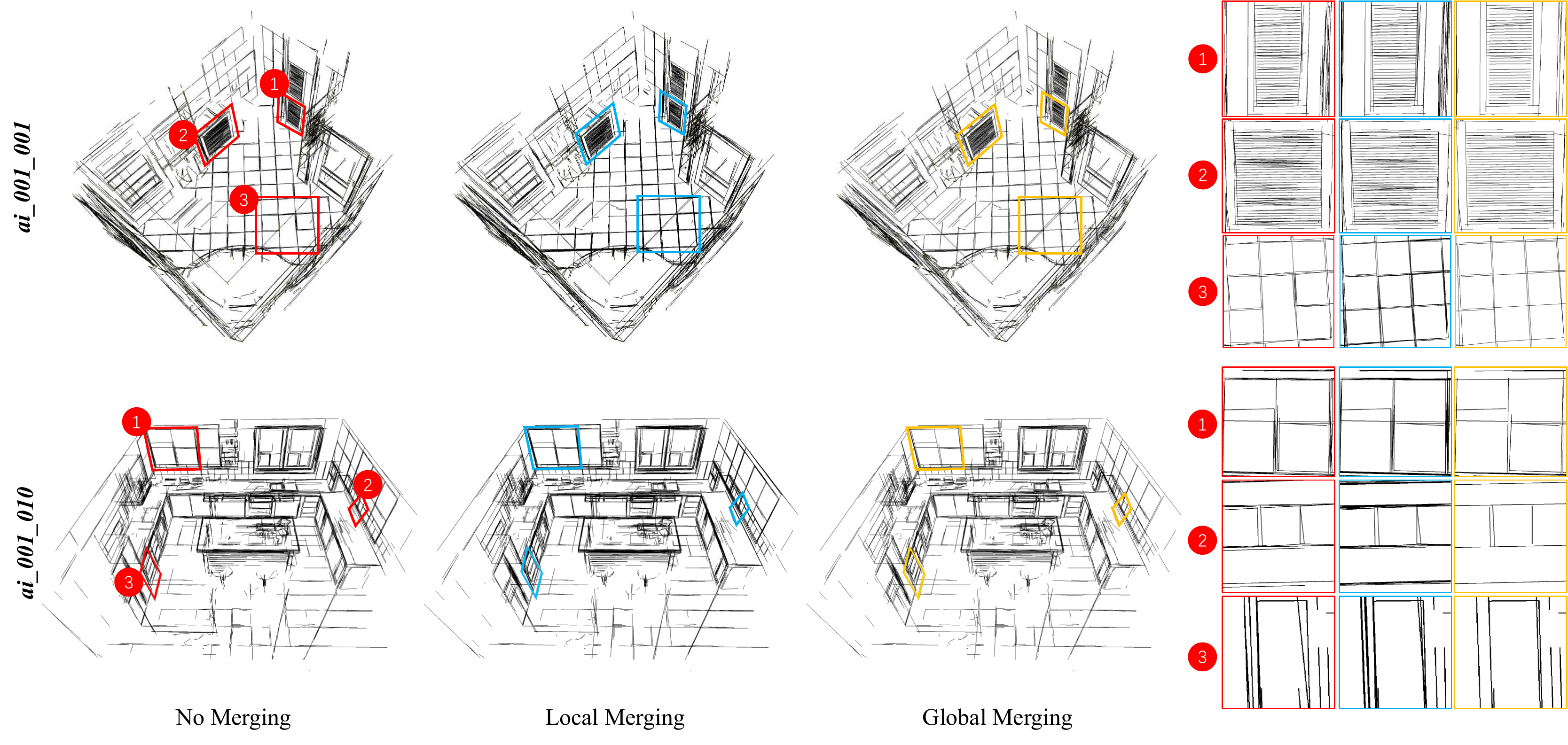}
\vspace{-3mm}
\caption{Qualitative results of our line merging strategies on two scenes from the Hypersim dataset~\cite{Hypersim}.
We crop some patches for detailed visualization. \textbf{Red}: original LiP-Map w/o merging. \textbf{Blue}: LiP-Map w/ local merging. \textbf{Orange}: LiP-Map w/ global merging.
}
\label{fig:merge}
\end{figure*}

The qualitative comparison of results on \textrm{ai\_001\_001} and \textrm{ai\_001\_010} scenes from the Hypersim dataset~\cite{Hypersim} is shown in Fig.~\ref{fig:merge}.
Compared to the original results without merging, the local merging strategy can clean up incorrect lines and merge corresponding lines, which would also incorporate many more redundant lines.
The global merging strategy further consolidates corresponding lines, removes redundant lines, and ultimately improves the quality and completeness of the line mapping.
Due to the complexity, the global merging strategy is more computationally expensive than the local merging strategy, which is acceptable based on the final results.

\begin{algorithm}[!h]
    \scriptsize
    \caption{Global Line Merging Strategy}
    \label{alg:global_merge}
    \KwIn{Optimized 3D planes $\mathcal{P}$, detected 2D lines $\{\mathbf{L}_{i}\}_{i=1}^N$ from all views, perpendicular distance threshold $\tau_{dbscan}$}
    \KwOut{Merged 3D line map $\mathcal{L}_{merged}$}

    \tcp{Step 1: Multi-view line association}
    \ForEach{view $i \in \{1, \ldots, N\}$}{
        \ForEach{detected 2D line $\mathbf{l}_{i,j} \in \mathbf{L}_{i}$}{
            Find all 3D lines $\mathcal{L}_{i,j} \subset \mathcal{L}$ by casting rays from the 1-pixel region of $\mathbf{l}_{i,j}$ and collecting edges from intersected 3D planes\;

            \tcp{Step 2: Local DBSCAN clustering}
            Compute pairwise perpendicular distances for all line pairs in $\mathcal{L}_{i,j}$ by sampling 10 points per line\;
            Apply DBSCAN with threshold $\tau_{dbscan}$ to obtain clusters $\{\mathcal{C}_{i,j,k}\}$\;

            \tcp{Step 3: Global identifier propagation}
            \ForEach{cluster $\mathcal{C}_{i,j,k}$}{
                \If{any line in $\mathcal{C}_{i,j,k}$ has a global identifier $g_{id}$}{
                    Assign $g_{id}$ to all lines in $\mathcal{C}_{i,j,k}$\;
                }
                \Else{
                    Create new global identifier $g_{id}^{new}$ and assign to all lines in $\mathcal{C}_{i,j,k}$\;
                }
            }
        }
    }

    \tcp{Step 4: PCA-based line merging}
    Group all 3D lines by their global identifiers $\{\mathcal{G}_g\}$\;
    \ForEach{global group $\mathcal{G}_g$}{
        Apply PCA to obtain main direction $\mathbf{d}_g$ and mean point $\mathbf{p}_g$\;
        Project all endpoints in $\mathcal{G}_g$ onto the main line to find $t_{min}$ and $t_{max}$\;
        Construct merged line: $\mathbf{l}_g = (\mathbf{p}_g + t_{min} \cdot \mathbf{d}_g, \mathbf{p}_g + t_{max} \cdot \mathbf{d}_g)$\;
        Add $\mathbf{l}_g$ to $\mathcal{L}_{merged}$\;
    }

    \Return{$\mathcal{L}_{merged}$}
\end{algorithm}

\section{Enhancement for 3D Planar Reconstruction}

\begin{figure}[!t]
    \centering
    \includegraphics[width=\linewidth]{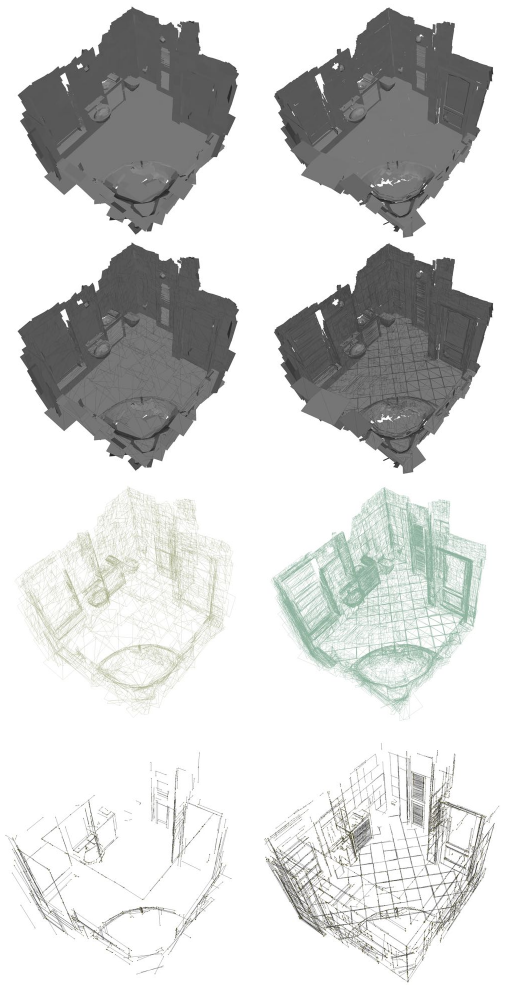}
    \caption{Illustration of the comparison of planar mesh/edges reconstruction with baseline PlanarSplatting~\cite{PlanarSplatting2024} on ``ai\_001\_001" from Hypersim~\cite{Hypersim}. Left column: PlanarSplatting. Right column: ours. First row: planar surface. Second row: planar mesh. Third row: planar edges. Last row: 3D line maps.}
    \label{fig:abs_plane}
\end{figure}

Since our \method enhances the reconstruction of plane edges, leading to a more accurate perception of physical boundaries, it can further improve performance in 3D planar reconstruction.
Following PlanarSplatting~\cite{PlanarSplatting2024}, we use the metrics including Variation of Information (VOI), Rand Index (RI), and Segmentation Covering (SC).
As shown in Table~\ref{tab:abs_plane_seg}, our \method achieves superior performance compared to PlanarSplatting~\cite{PlanarSplatting2024}.
We also illustrate a comparison of reconstruction results in Fig.~\ref{fig:abs_plane}.
As evident from the comparison, our method effectively aligns the edges of planar primitives with the line structures in the scene, resulting in reconstructed meshes with more prominent geometric topology.
This leads to more accurate and plausible reconstructions, enabling the extraction of a complete and precise 3D line map from the scene.

\begin{table}[!t]
    \centering
    \caption{Quantitative comparison of planar reconstruction results on the ScanNetV2 dataset~\cite{scannet-DaiCSHFN17}.}
    \label{tab:abs_plane_seg}
    \small
    \setlength{\tabcolsep}{10pt}
        \begin{tabular}{l|ccc}
        \toprule
         & {VOI$\downarrow$} & {RI$\uparrow$} & {SC$\uparrow$}\\
        \midrule
        PlanarSplatting~\cite{PlanarSplatting2024} & 2.489 & 0.951 & 0.530 \\
        Ours  & \textbf{2.469} & \textbf{0.952} & \textbf{0.532} \\
        \bottomrule
        \end{tabular}
\end{table}

\section{Limitations}

Despite its advantages, our method has certain limitations that warrant further study. Due to errors in 2D line detection, duplicate 3D line segments may appear, potentially violating the parsimony principle in 3D line mapping.
Additionally, the plane-based representation of 3D line segments is particularly well-suited for reconstructing architectural structures in man-made environments, though it has limited capability in reconstructing non-planar, unstructured objects. Nevertheless, we argue that 3D line segment reconstruction primarily aims to perceive geometric features in the environment and efficiently capture the outlines and topological structures of buildings, generating lightweight, semantically rich vectorized models—areas in which our method demonstrates clear advantages.

\section{More Qualitative Results of Our Line Maps}

\begin{figure*}[!t]
    \centering
    \begin{subfigure}[t]{0.3\textwidth}
    \centering
    \includegraphics[width=\linewidth]{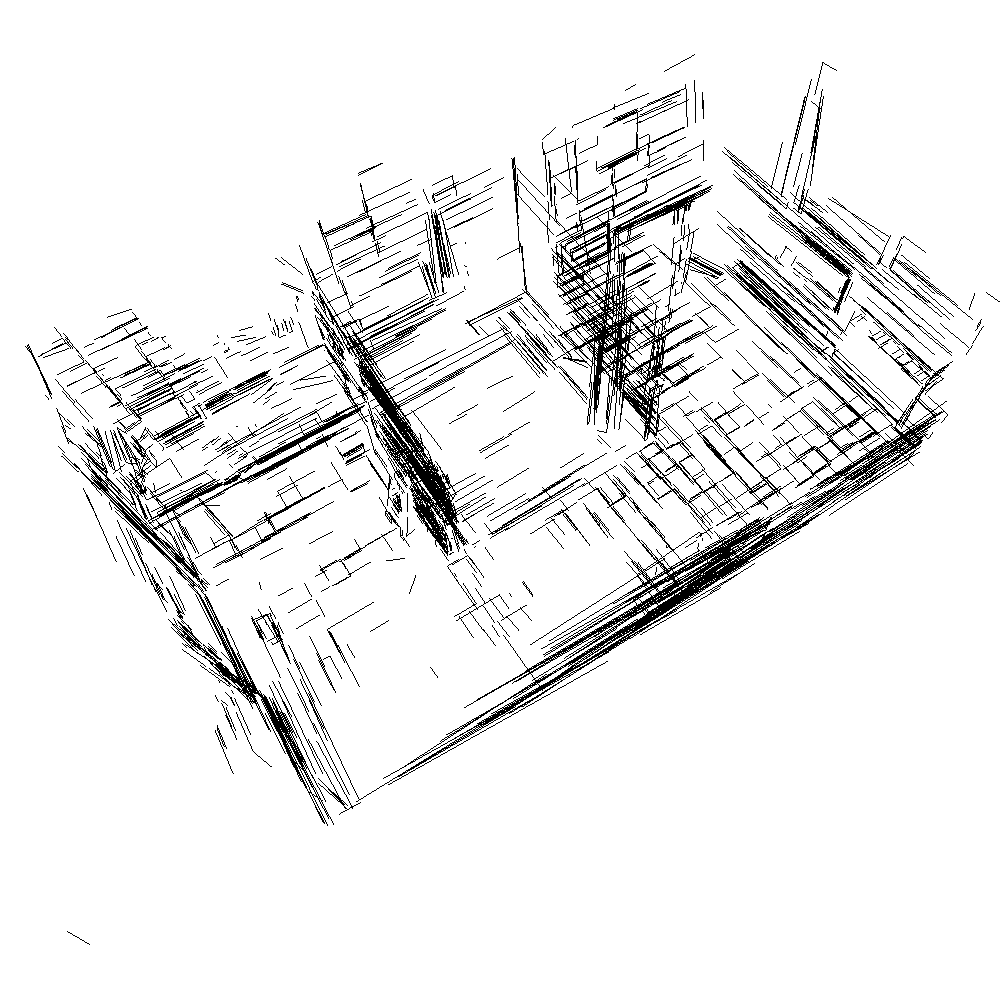}
    \caption{\textit{scene0084\_00}}
    \end{subfigure}
    \begin{subfigure}[t]{0.3\textwidth}
    \centering
    \includegraphics[width=\linewidth]{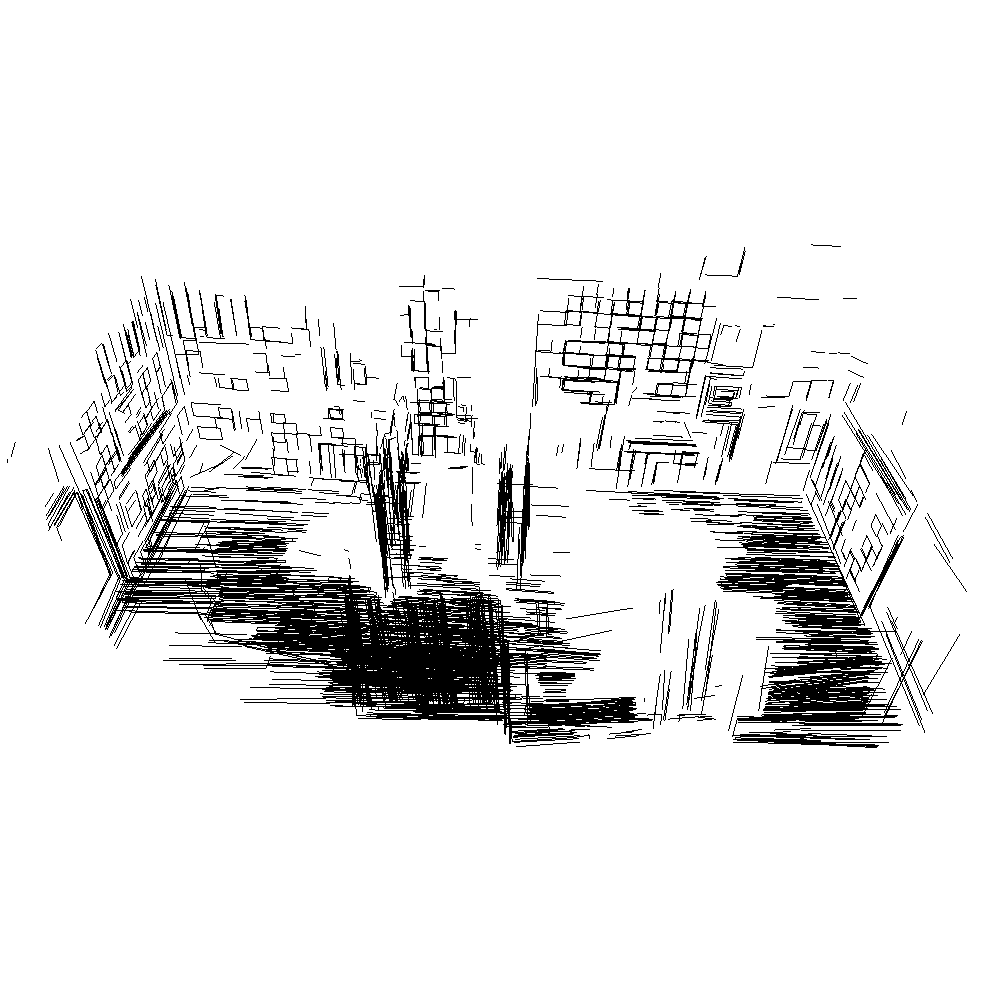}
    \caption{\textit{scene0086\_00}}
    \end{subfigure}
    \begin{subfigure}[t]{0.3\textwidth}
    \centering
    \includegraphics[width=\linewidth]{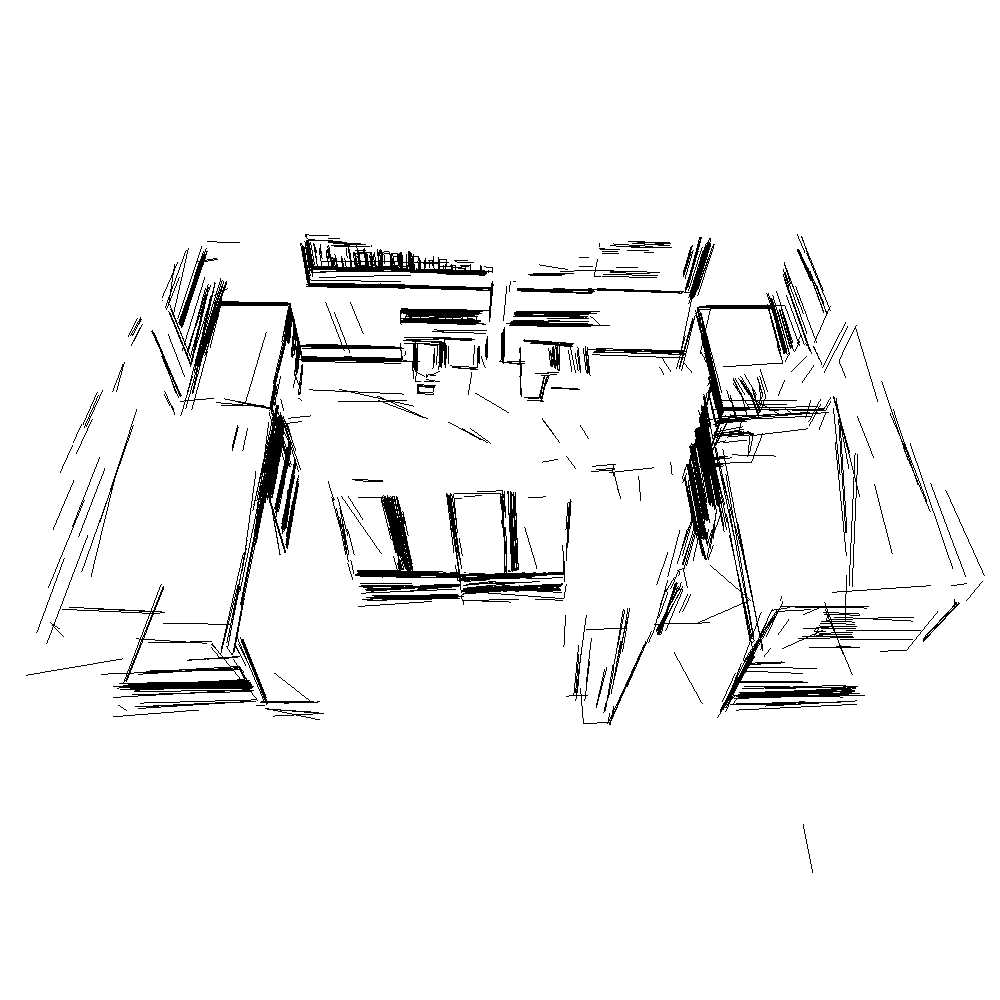}
    \caption{\textit{scene0217\_00}}
    \end{subfigure}
    \hfill

    \begin{subfigure}[t]{0.3\textwidth}
    \centering
    \includegraphics[width=\linewidth]{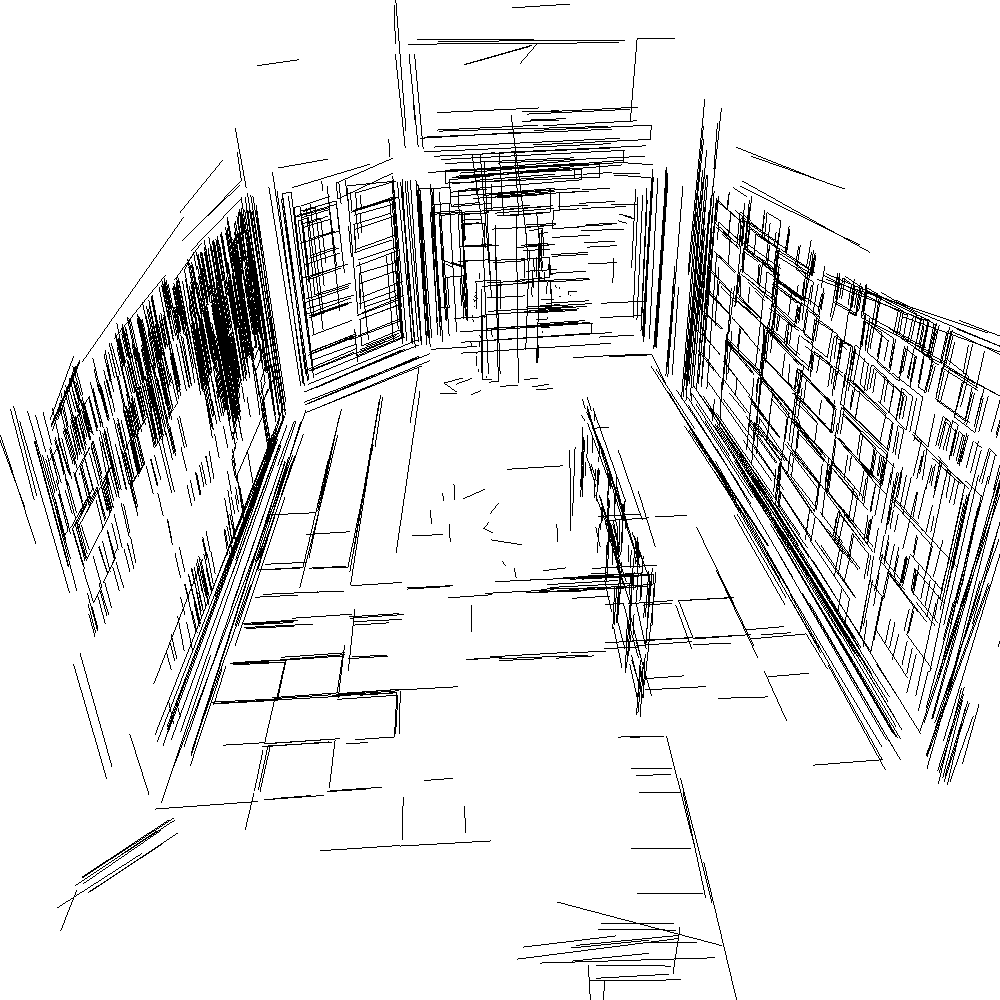}
    \caption{\textit{scene0304\_00}}
    \end{subfigure}
    \begin{subfigure}[t]{0.3\textwidth}
    \centering
    \includegraphics[width=\linewidth]{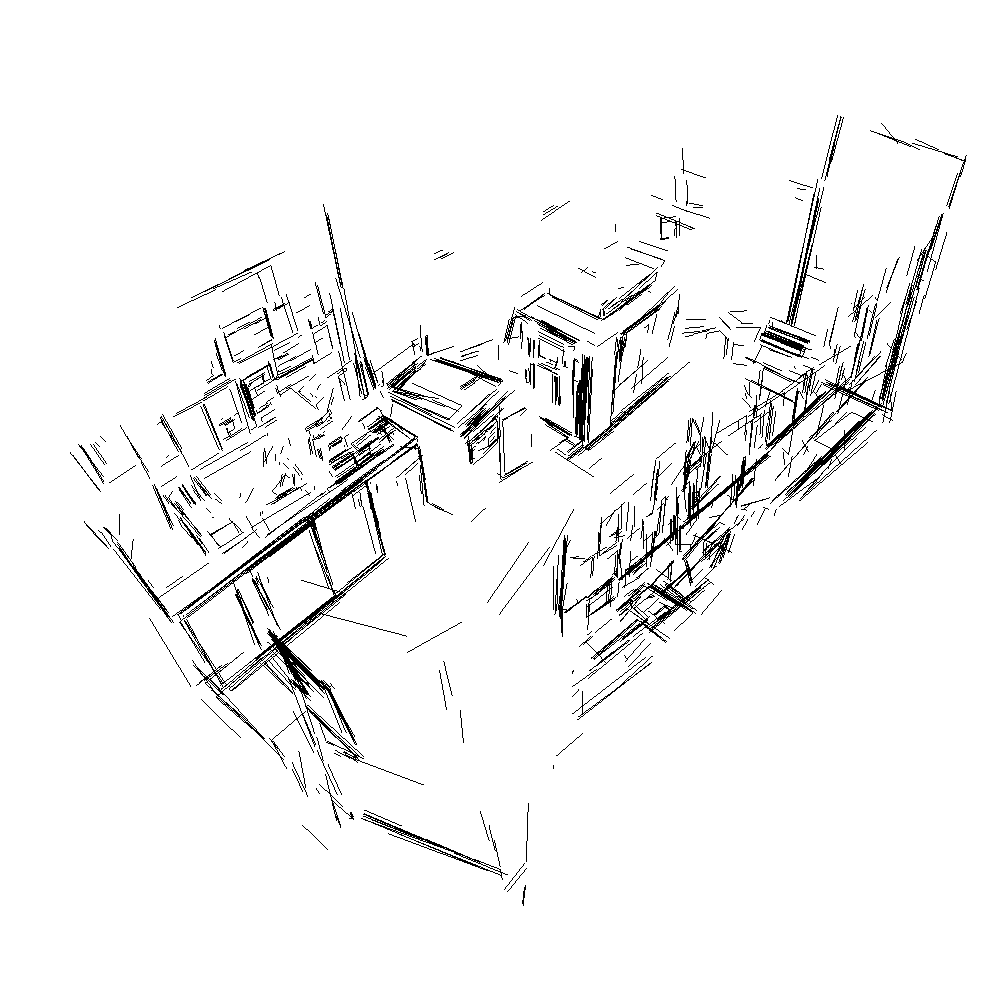}
    \caption{\textit{scene0462\_00}}
    \end{subfigure}
    \begin{subfigure}[t]{0.3\textwidth}
    \centering
    \includegraphics[width=\linewidth]{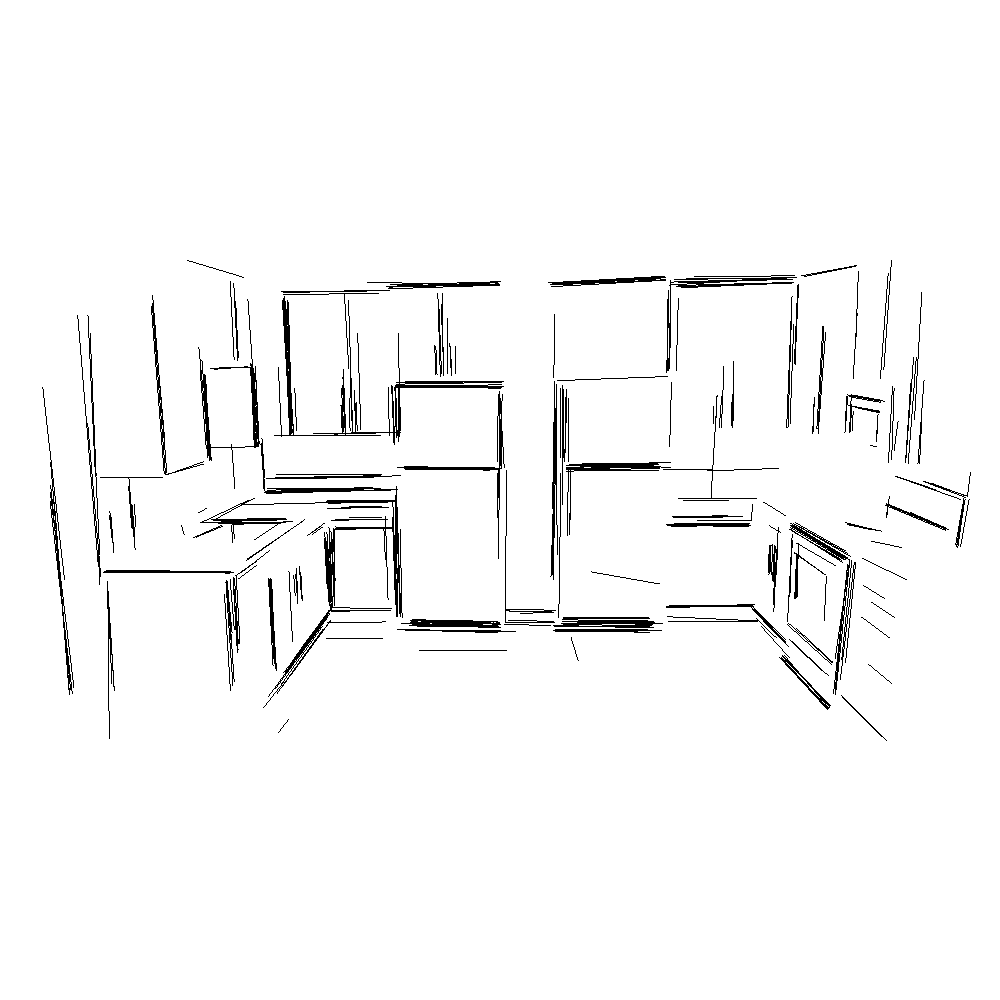}
    \caption{\textit{scene0488\_00}}
    \end{subfigure}
    \hfill

    \begin{subfigure}[t]{0.3\textwidth}
    \centering
    \includegraphics[width=\linewidth]{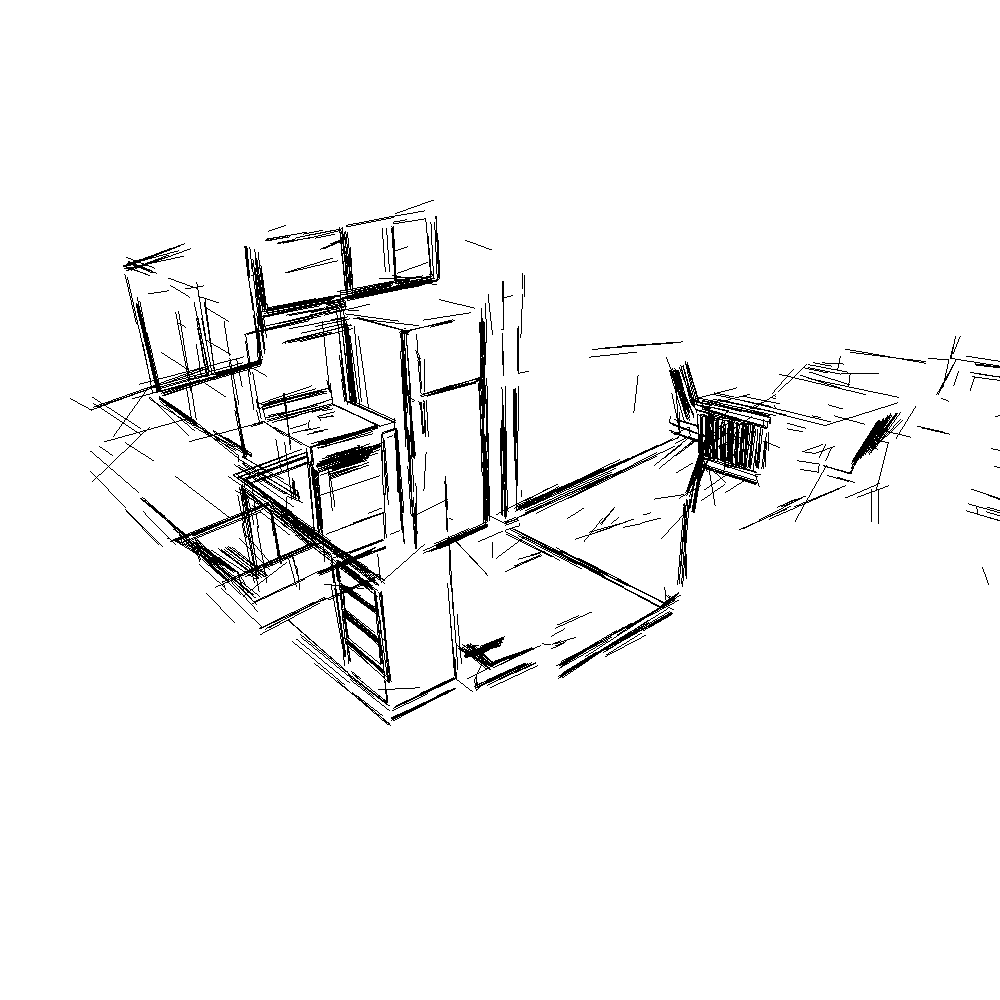}
    \caption{\textit{scene0651\_00}}
    \end{subfigure}
    \begin{subfigure}[t]{0.3\textwidth}
    \centering
    \includegraphics[width=\linewidth]{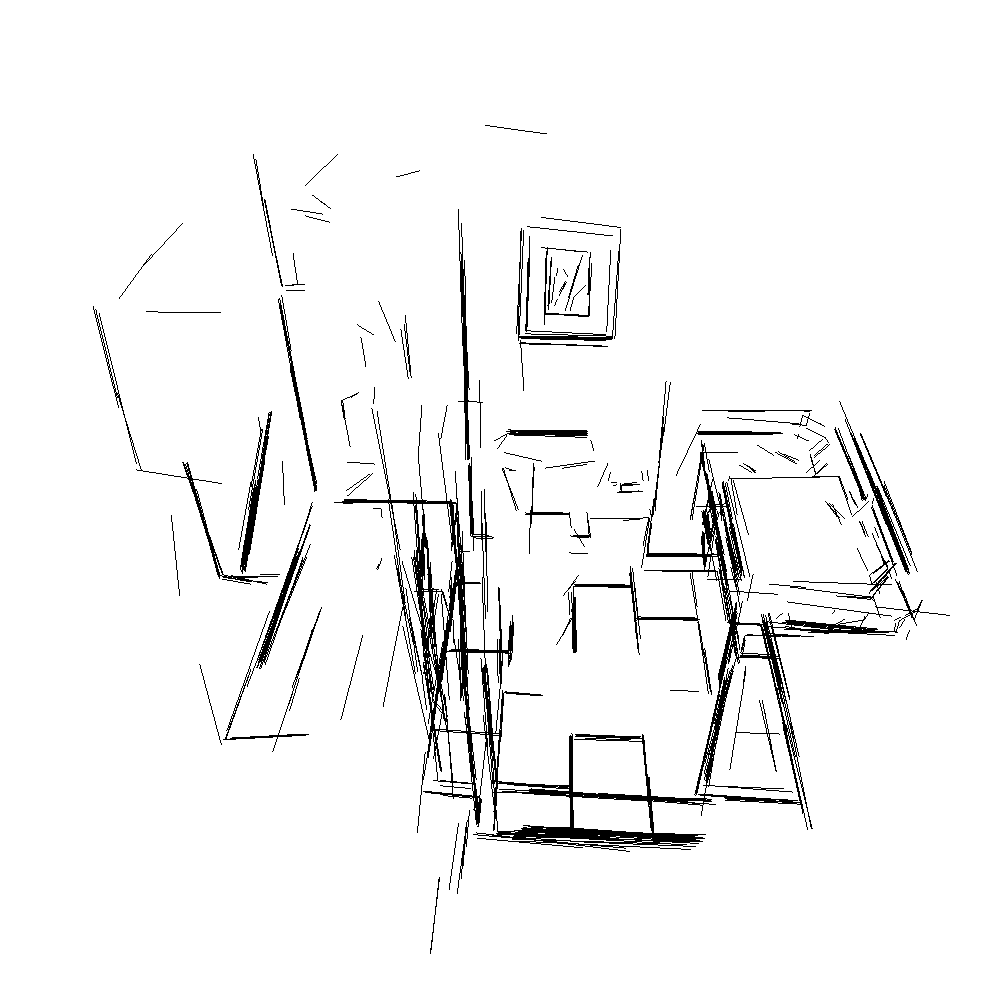}
    \caption{\textit{scene0664\_00}}
    \end{subfigure}
    \begin{subfigure}[t]{0.3\textwidth}
    \centering
    \includegraphics[width=\linewidth]{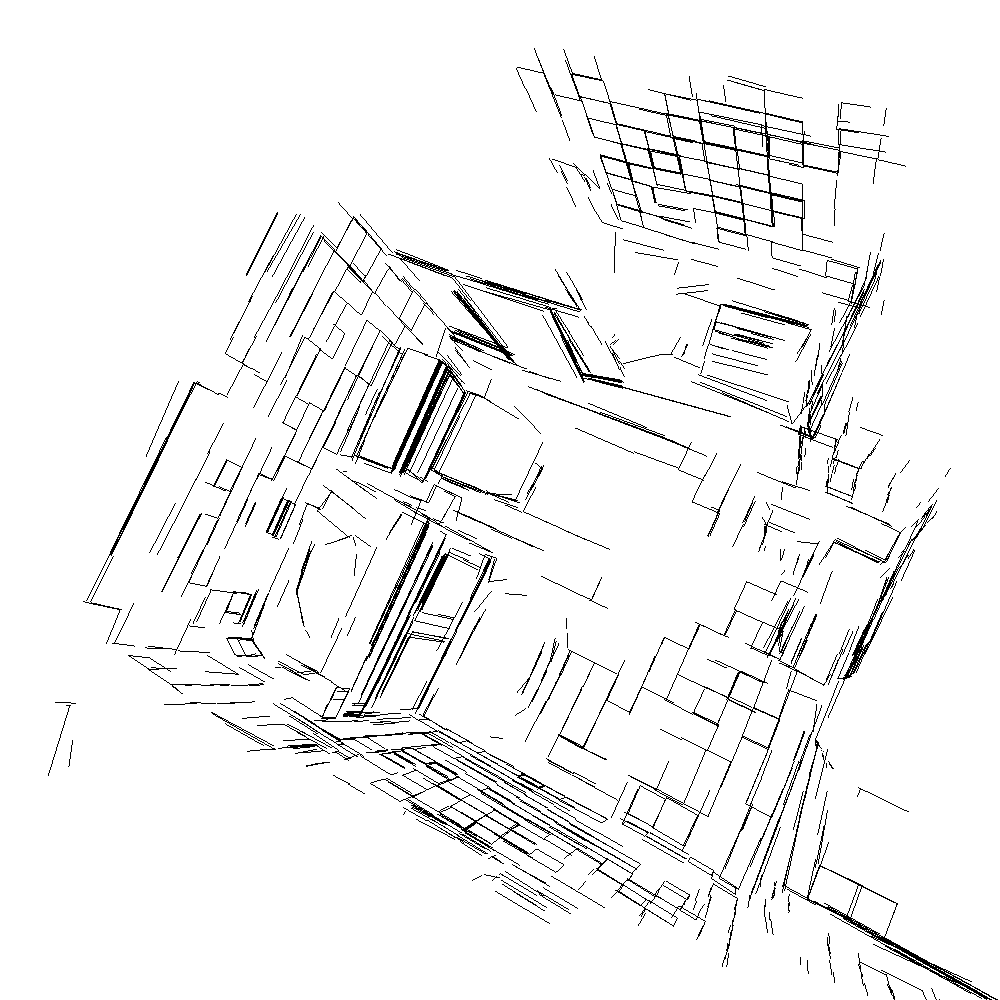}
    \caption{\textit{scene0693\_00}}
    \end{subfigure}
    \caption{More qualitative results of the 3D line maps recovered by our method on the ScanNetV2 dataset~\cite{scannet-DaiCSHFN17}.}
    \label{fig:v2s}
\end{figure*}

\begin{figure*}[!t]
    \centering
    \begin{subfigure}[t]{0.3\textwidth}
    \centering
    \includegraphics[width=\linewidth]{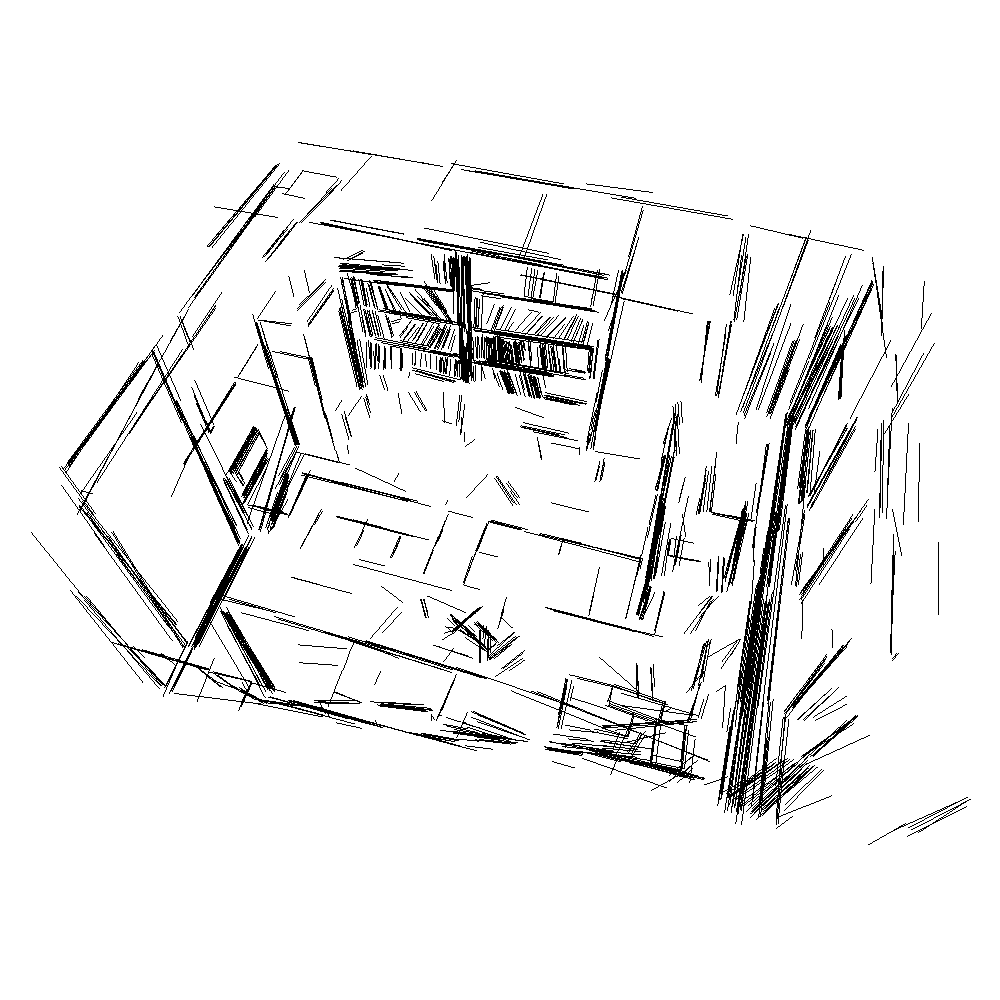}
    \caption{\textit{8a20d62ac0}}
    \end{subfigure}
    \begin{subfigure}[t]{0.3\textwidth}
    \centering
    \includegraphics[width=\linewidth]{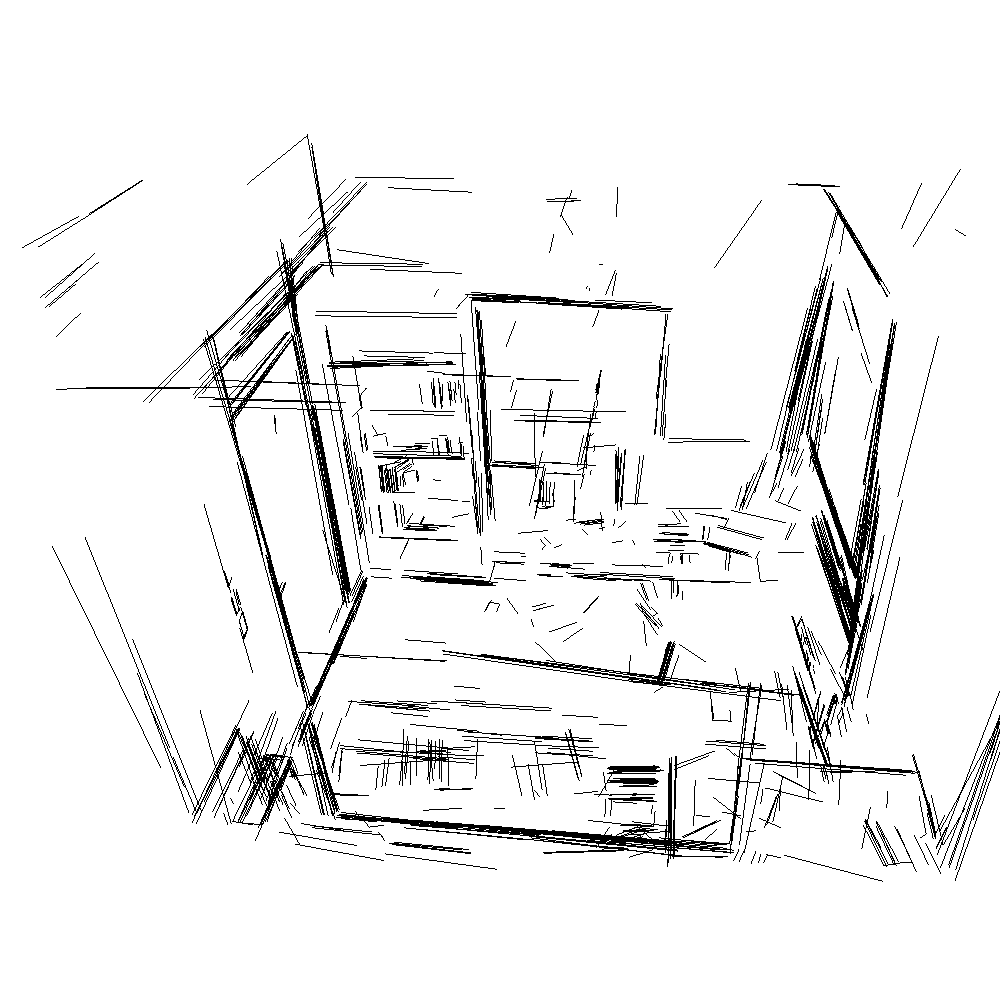}
    \caption{\textit{9b74afd2d2}}
    \end{subfigure}
    \begin{subfigure}[t]{0.3\textwidth}
    \centering
    \includegraphics[width=\linewidth]{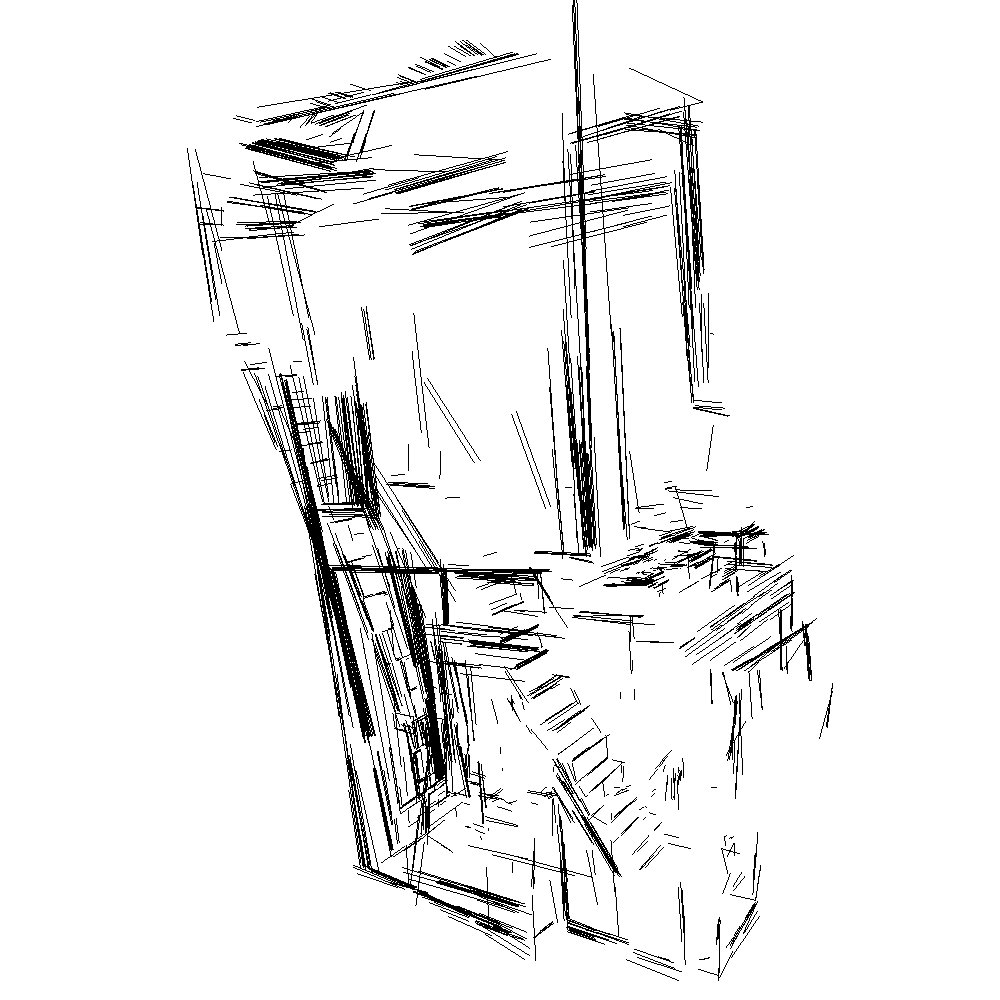}
    \caption{\textit{9f21bdec45}}
    \end{subfigure}
    \hfill

    \begin{subfigure}[t]{0.3\textwidth}
    \centering
    \includegraphics[width=\linewidth]{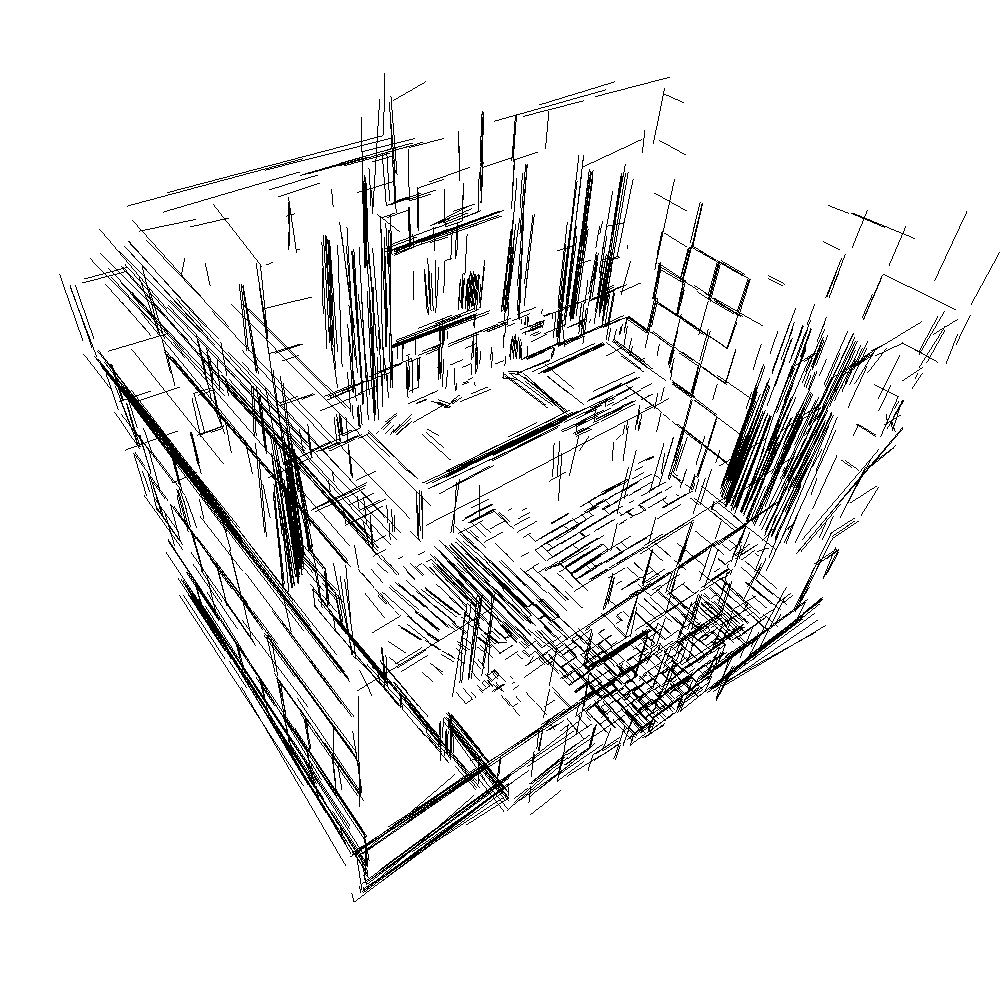}
    \caption{\textit{45b0dac5e3}}
    \end{subfigure}
    \begin{subfigure}[t]{0.3\textwidth}
    \centering
    \includegraphics[width=\linewidth]{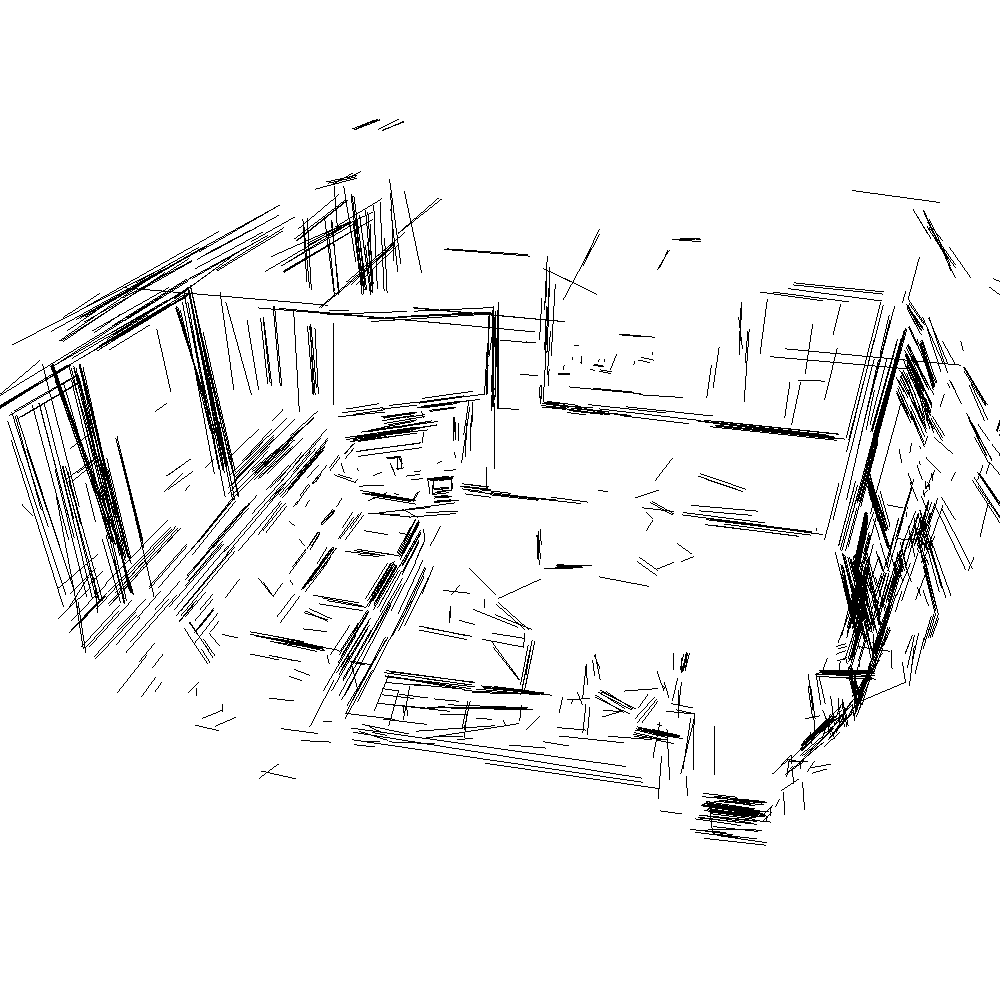}
    \caption{\textit{bc2fce1d81}}
    \end{subfigure}
    \begin{subfigure}[t]{0.3\textwidth}
    \centering
    \includegraphics[width=\linewidth]{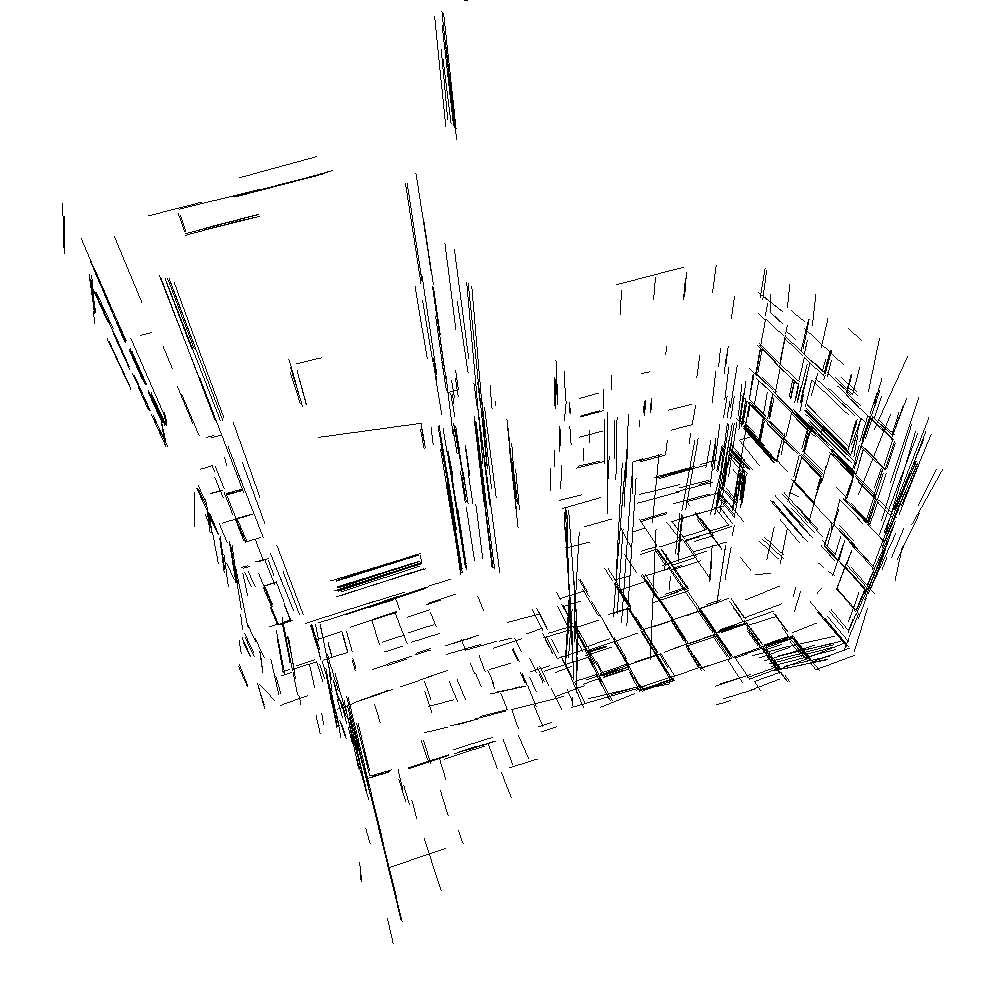}
    \caption{\textit{f3685d06a9}}
    \end{subfigure}
    \caption{More qualitative results of the 3D line maps recovered by our method on the ScanNet++ dataset~\cite{scannetpp-YeshwanthLND23}.}
    \label{fig:pps}
\end{figure*}

\begin{figure*}[!t]
    \centering
    \begin{subfigure}[t]{0.3\textwidth}
    \centering
    \includegraphics[width=\linewidth]{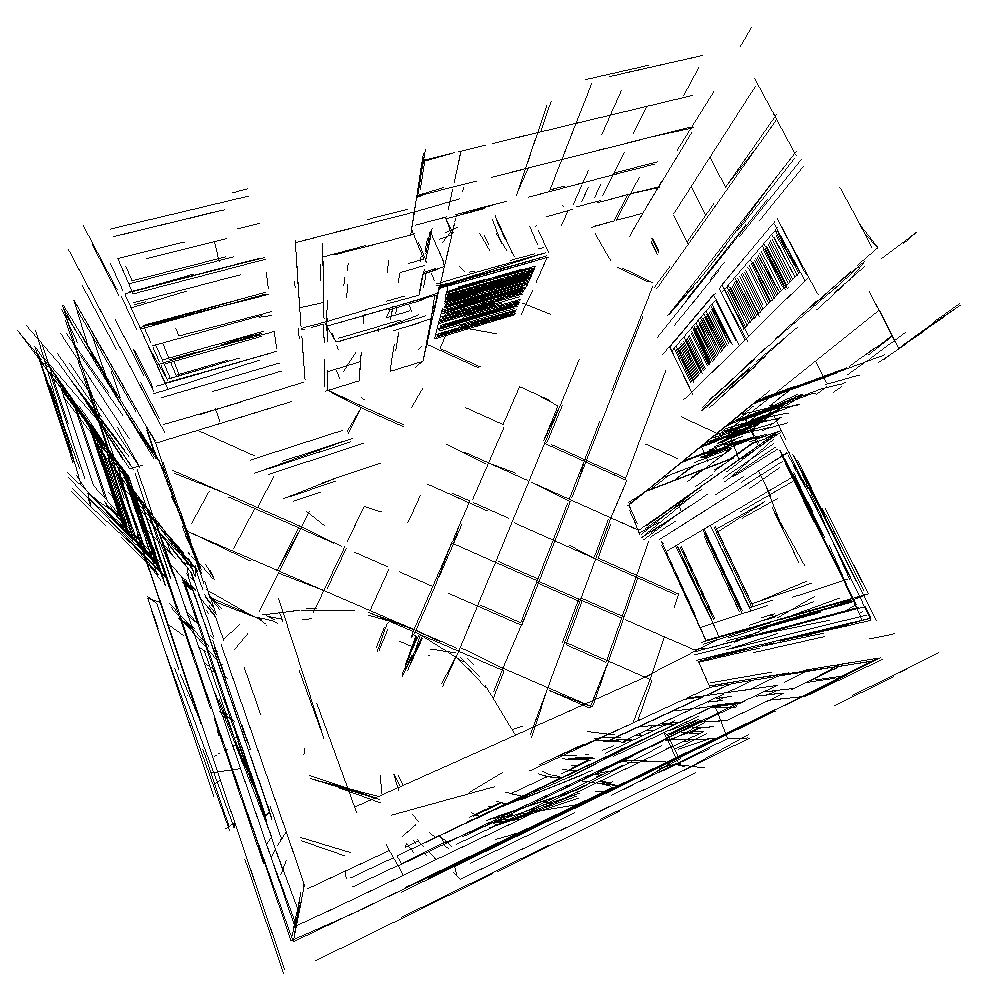}
    \caption{\textit{ai\_001\_001}}
    \end{subfigure}
    \begin{subfigure}[t]{0.3\textwidth}
    \centering
    \includegraphics[width=\linewidth]{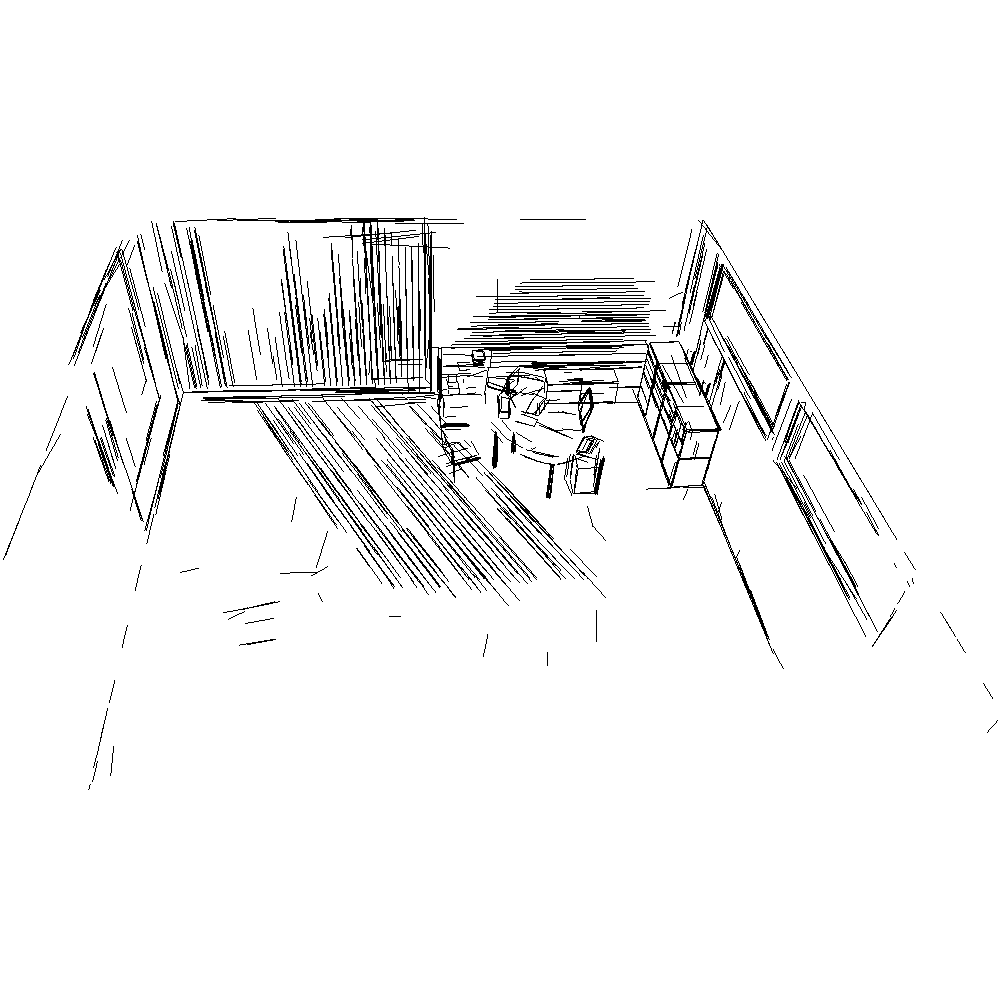}
    \caption{\textit{ai\_001\_003}}
    \end{subfigure}
    \begin{subfigure}[t]{0.3\textwidth}
    \centering
    \includegraphics[width=\linewidth]{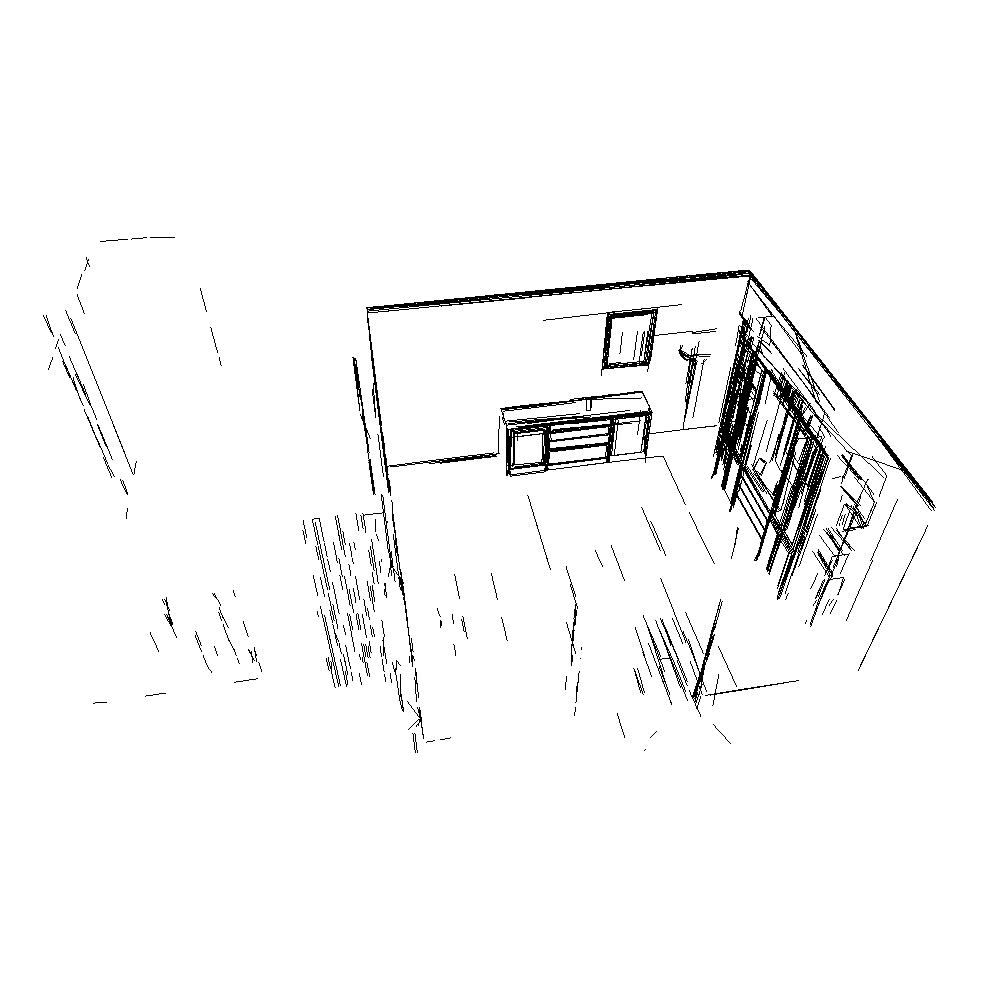}
    \caption{\textit{ai\_001\_004}}
    \end{subfigure}
    \hfill

    \begin{subfigure}[t]{0.3\textwidth}
    \centering
    \includegraphics[width=\linewidth]{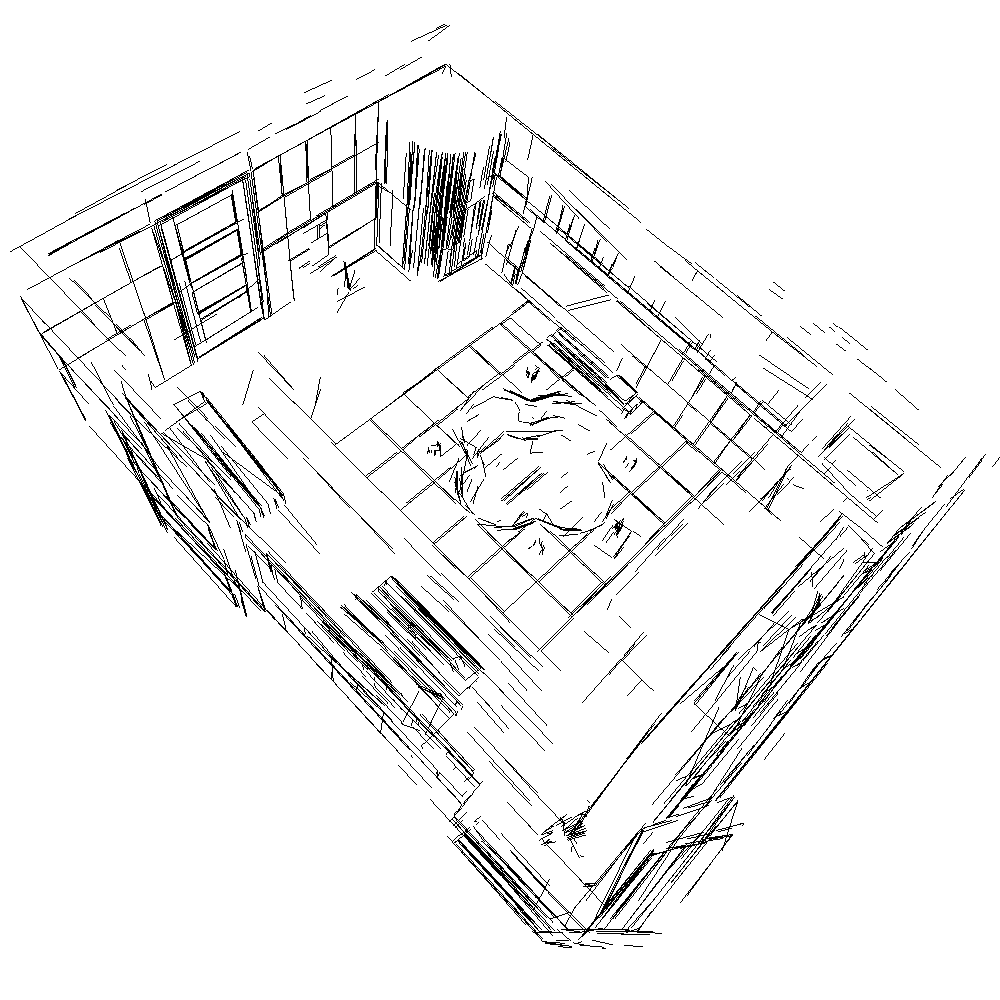}
    \caption{\textit{ai\_001\_007}}
    \end{subfigure}
    \begin{subfigure}[t]{0.3\textwidth}
    \centering
    \includegraphics[width=\linewidth]{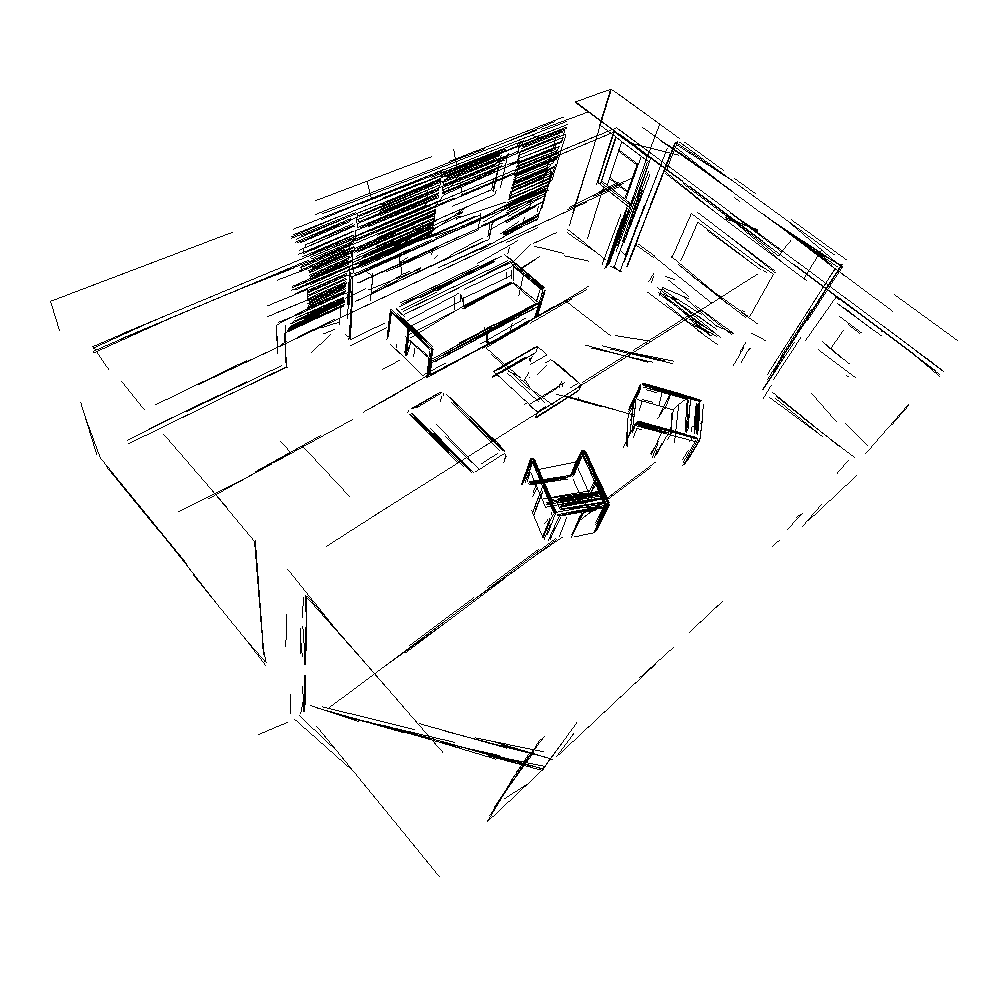}
    \caption{\textit{ai\_001\_008}}
    \end{subfigure}
    \begin{subfigure}[t]{0.3\textwidth}
    \centering
    \includegraphics[width=\linewidth]{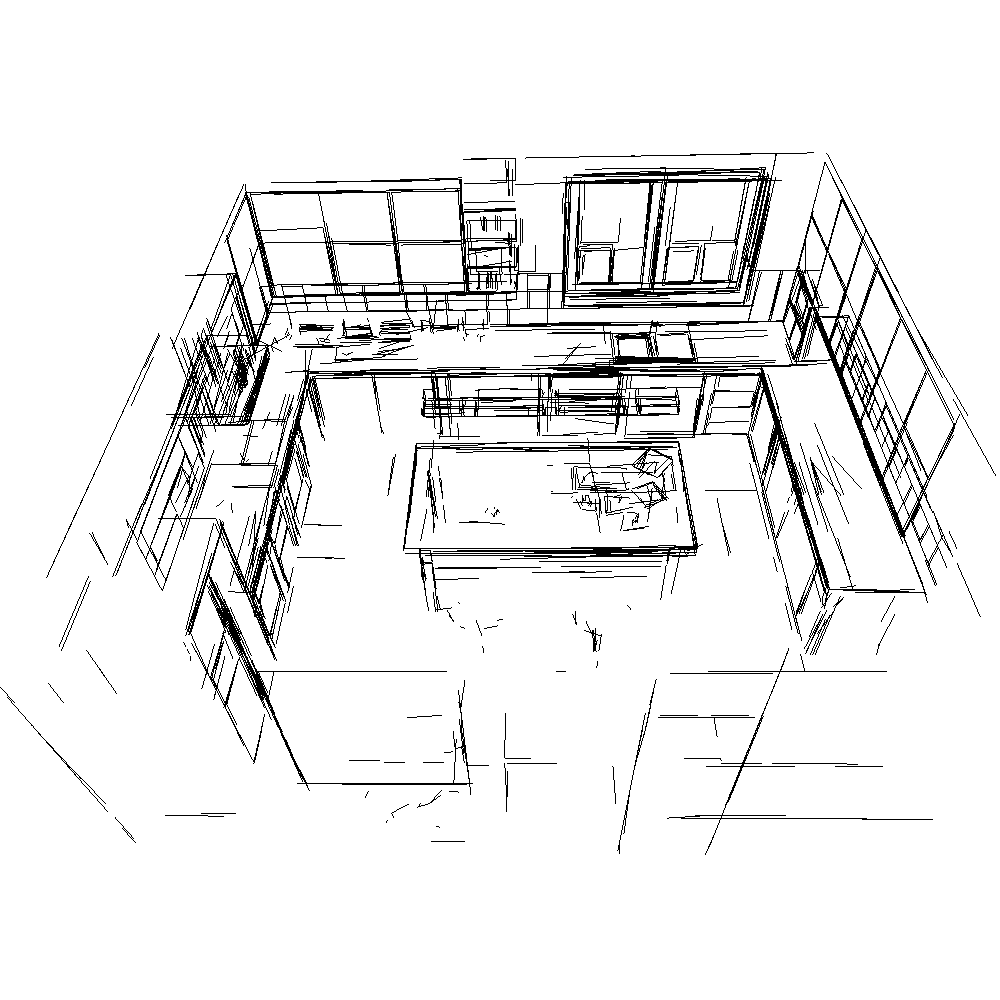}
    \caption{\textit{ai\_001\_010}}
    \end{subfigure}
    \caption{More qualitative results of the 3D line maps recovered by our method on the Hypersim dataset~\cite{Hypersim}.}
    \label{fig:hypersims}
\end{figure*}

We visualize more qualitative results of our 3D line maps on scenes from the ScanNetV2 dataset~\cite{scannet-DaiCSHFN17}(Fig~\ref{fig:v2s}), the ScanNet++ dataset~\cite{scannetpp-YeshwanthLND23} (Fig.~\ref{fig:pps}), and the Hypersim dataset~\cite{Hypersim} (Fig.~\ref{fig:hypersims}).

\section{Qualitative Results on The Sensitivity to Heuristics and Hyperparameters}

We have reported the quantitative results on the sensitivity to heuristics and hyperparameters in Sec.\ref{sec:sensitivity}.
In this section, we would like to provide more detailed discussions and the corresponding qualitative results.

\begin{table}[!t]
\centering
\caption{Optimization time and number of planes versus the splitting threshold.}
\label{tab:split_gradient}
\resizebox{\linewidth}{!}{
\begin{tabular}{l| cccccc}
\toprule
Threshold & 0.02 & 0.1 & 0.2 & 1.0 & 2.0 & 10 \\
\midrule
Time(s) & 948.5209 & 388.9839 & 333.0532 & 312.5148 & 305.0371 & 301.3305 \\
Init. \#Planes & 1873 & 1873 & 1873 & 1873 & 1873 & 1873  \\
Final. \#Planes & 115615 & 19876 & 10064 & 2071 & 1369 & 1072 \\
\bottomrule
\end{tabular}
}
\end{table}

\begin{figure*}
\centering
\includegraphics[width=\linewidth]{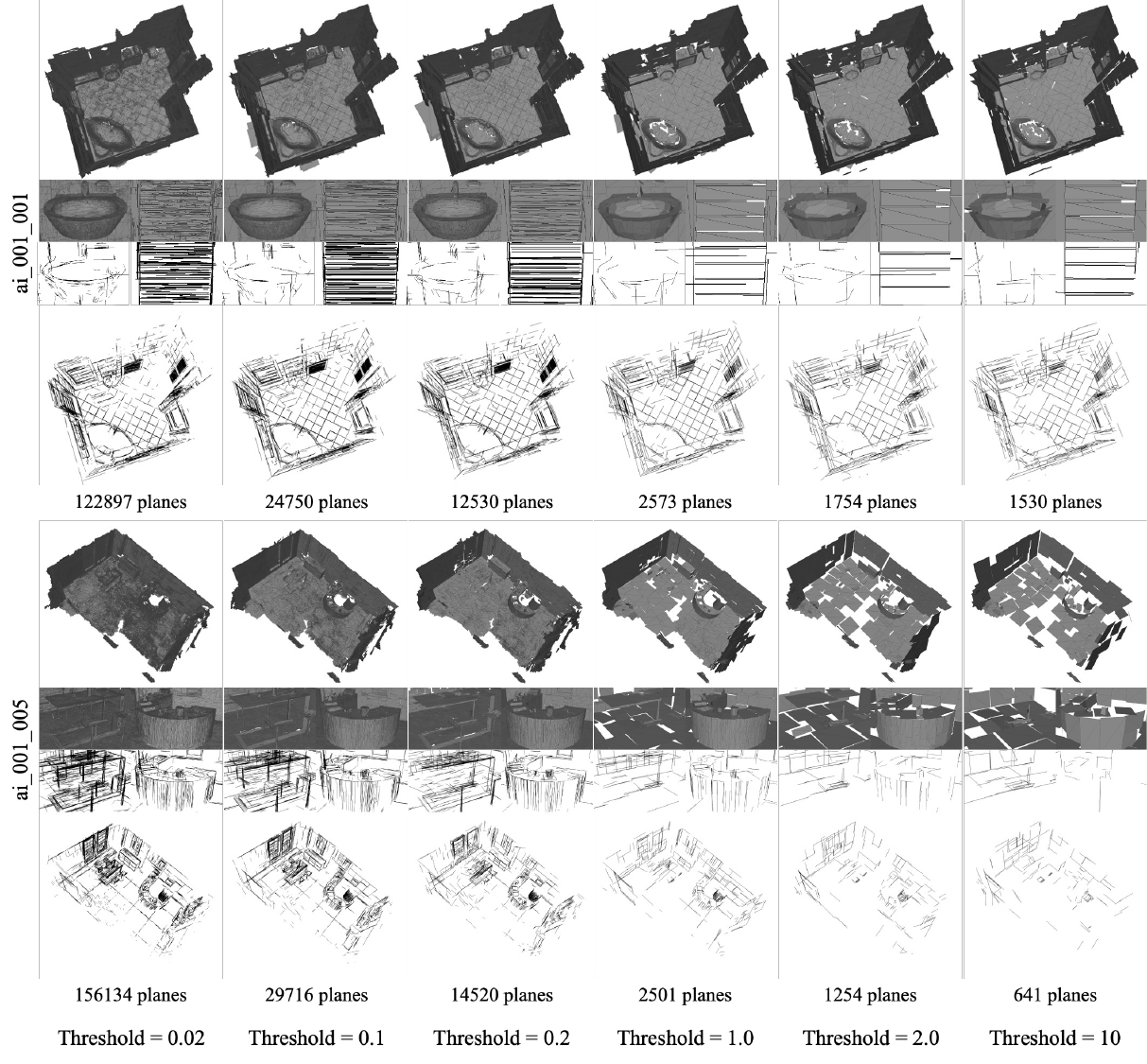}
\vspace{-3mm}
\caption{Qualitative results of reconstruction with different plane splitting thresholds. Some patches are cropped for detailed comparison, and the final numbers of planes are attached below.}
\label{fig:split_gradient}   
\end{figure*}

\paragraph{The plane splitting threshold (based on radius gradients).}
This parameter is fixed to 0.2 in all experiments.
On the one hand, we keep it consistent with the setting in PlanarSplatting~\cite{PlanarSplatting2024}. 
On the other hand, a small threshold makes planes split too aggressively, causing the number of planes to explode, whereas a large threshold slows down splitting, leaving too few planes to support fine-grained reconstruction. 
We validate this claim by reporting statistics on the optimization time and the number of planes in Table~\ref{tab:split_gradient} and illustrating some details of reconstruction in Fig.~\ref{fig:split_gradient}. 

\paragraph{The initial number of random planar primitives.}
We have reported detailed results about how the optimization time varies with the number of initial planes in Table~\ref{tab:init_planes}, and provided the corresponding qualitative comparisons in Fig.~\ref{fig:plane_num}.
Notably, a larger number of planes for initialization is indeed necessary to obtain better reconstruction results when meeting large-scale scenes (as shown in Fig.~\ref{fig:failure}(b)). Therefore, the most straightforward choice is to use a large initialization value for all scenes (e.g., 50,000), and the only trade-off is a slight increase in optimization time (as reported in Table~\ref{tab:init_planes}). The 2000 initial planes are sufficient for general indoor scenes.

\begin{figure*}
\centering
\includegraphics[width=\linewidth]{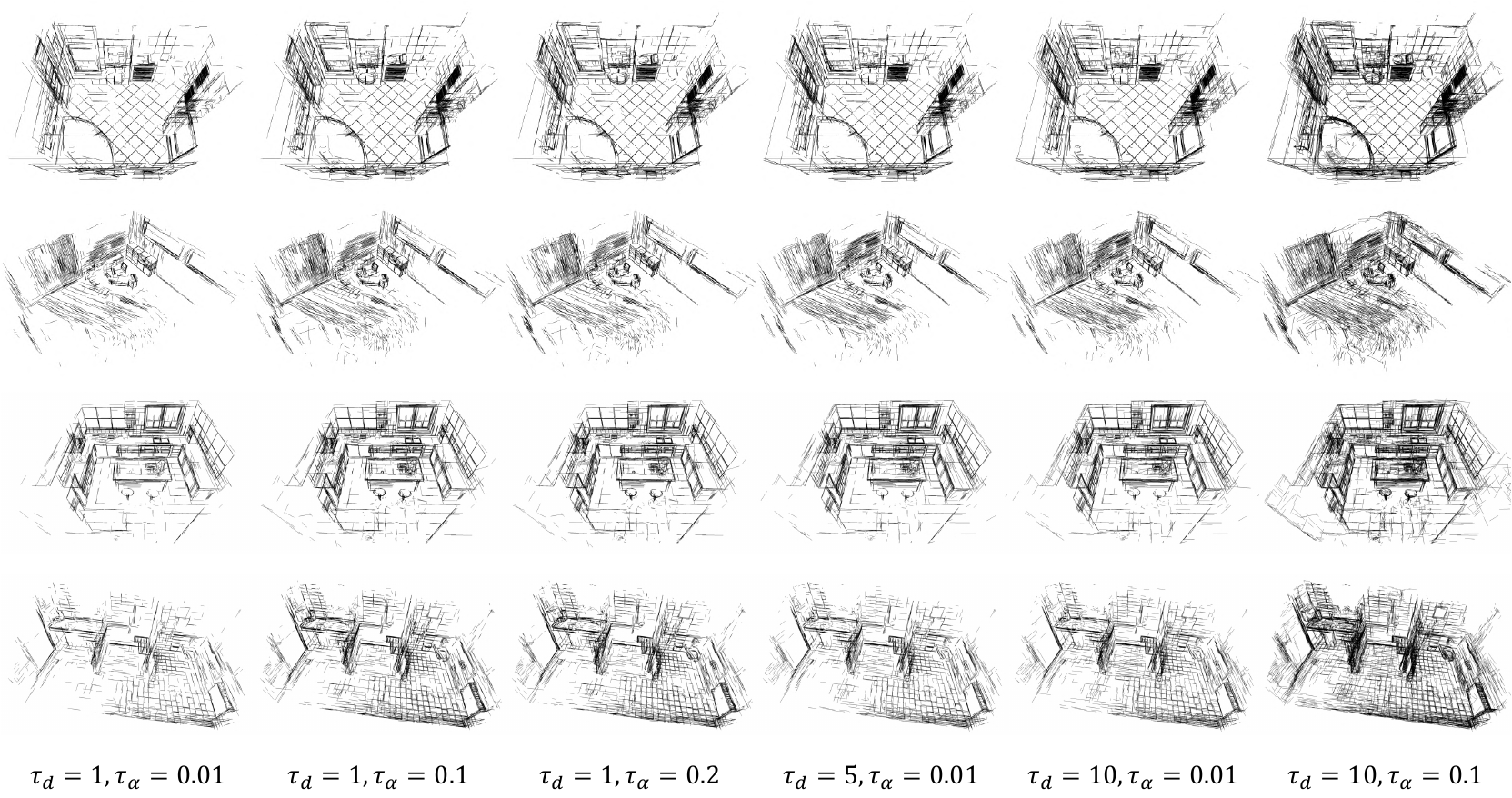}
\vspace{-3mm}
\caption{Qualitative results of reconstruction with different pairs of assignment thresholds for 3D line mapping.}
\label{fig:ass_thresholds}   
\end{figure*}

\paragraph{The assignment thresholds for 3D line mapping.}
The assignment thresholds are only used in the inference stage to achieve the final line mapping after optimization, as described in Sec.~\ref{subsec:ltb}.
We use a pair of very strict thresholds, $\tau_d = 1$ pixel and $\tau_{\alpha} = 0.01$ rad (about 0.57 degrees), in our paper to ensure accurate 3D line mapping. Relaxing the thresholds will yield many more 3D lines and output more complete 3D line maps, as shown in Fig.~\ref{fig:ass_thresholds}, which also introduces a lot of noisy lines.

\begin{figure*}
\centering
\includegraphics[width=\linewidth]{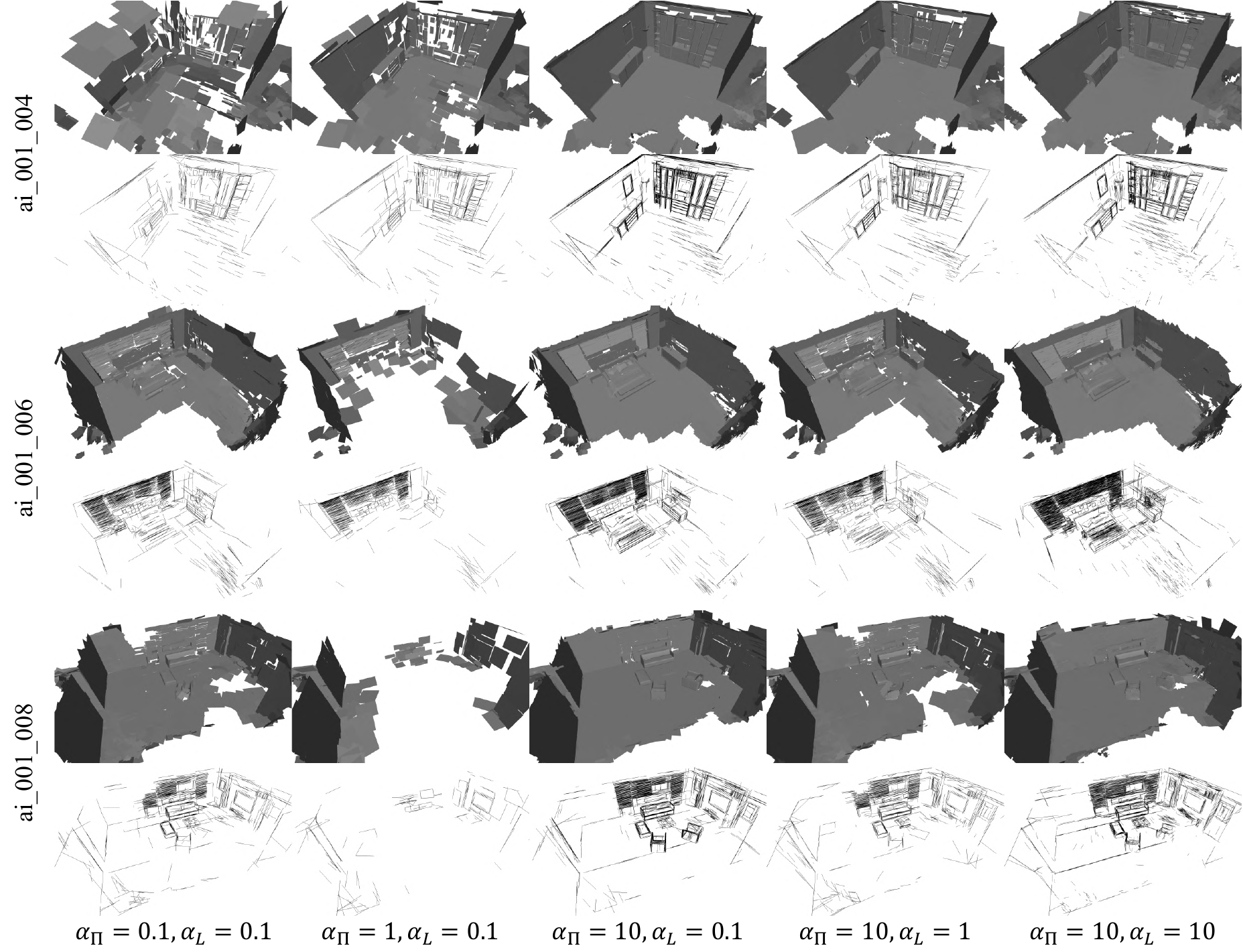}
\vspace{-3mm}
\caption{Qualitative results of reconstruction with different pairs of loss weights for 3D line mapping.}
\label{fig:loss_weights}   
\end{figure*}

\paragraph{The balancing weights $\alpha_{\mathbf{\Pi}}$,$\alpha_{L}$ in the joint optimization.}
Our choice of balancing weights, with $\alpha_{\mathbf{\Pi}}=10$,$\alpha_{L}=0.1$, is intended to bring the planar surface optimization loss and the 3D line optimization loss to a comparable numerical scale. This balance also makes the optimization process highly stable.
We are also pleased to provide qualitative visualizations of some results in Fig.~\ref{fig:loss_weights}. 
It can be analyzed that the large weight $\alpha_{L}$ for 3D line optimization can cause the number of planes to grow rapidly, thereby increasing the optimization time. In contrast, the small weight $\alpha_{\mathbf{\Pi}}$ for the planar-surface optimization will reduce the number of planes, which compromises complete and fine-grained scene reconstruction.

\section{Dependence on Initialization Quality:}

We agree that it is crucial to understand whether the performance gains are primarily due to the proposed joint optimization framework or simply inherited from high-quality priors.

\begin{figure*}
\centering
\includegraphics[width=\linewidth]{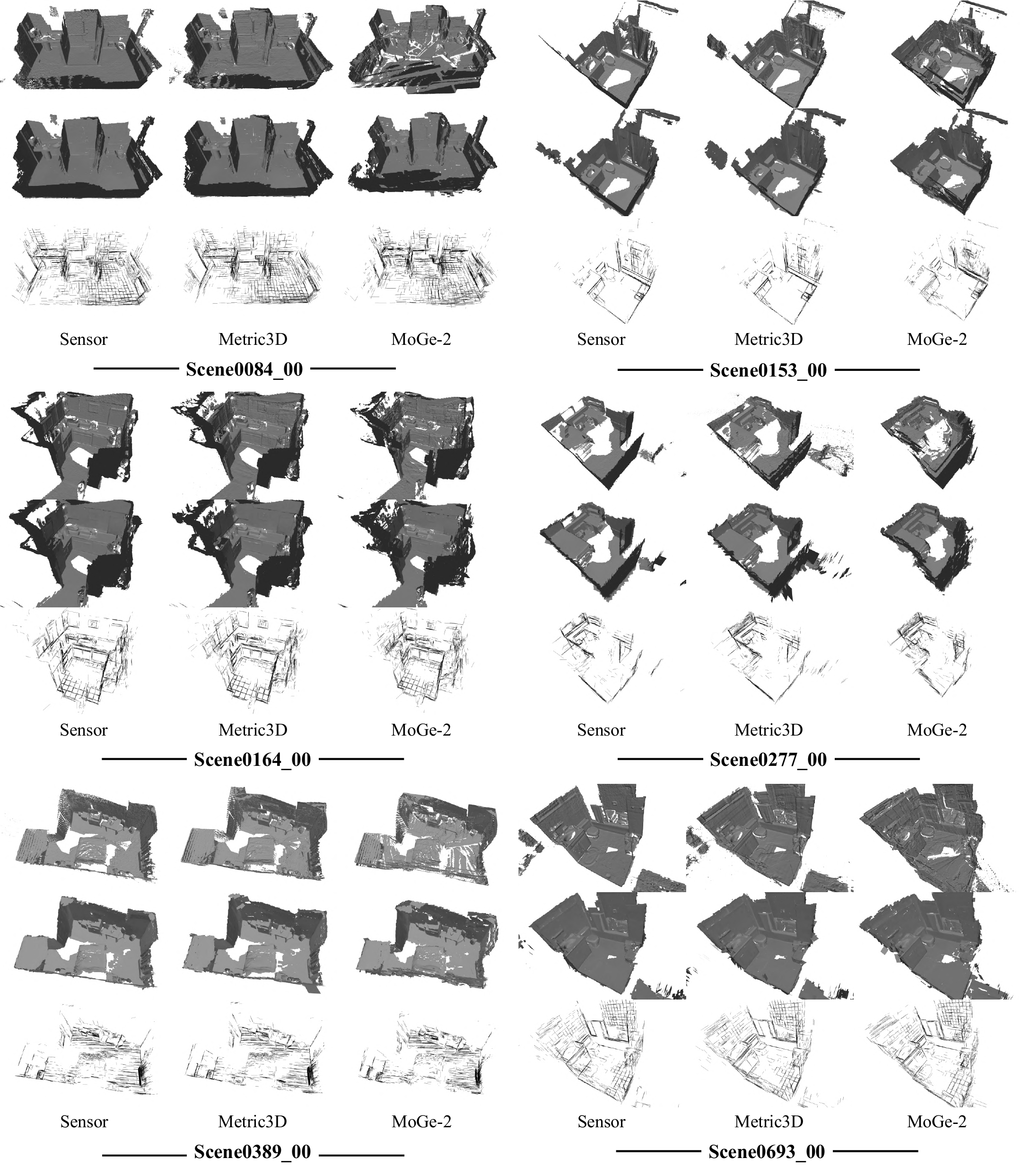}
\vspace{-3mm}
\caption{Qualitative comparison of initial meshes (first row), our planar meshes (second row), and our 3D line maps (last row) on ScanNetV2~\cite{scannet-DaiCSHFN17}. Normal maps predicted by Omnidata~\cite{omnidata-EftekharSMZ21} are adopted in this experiment.}
\label{fig:ds_mesh_line}   
\end{figure*}

\paragraph{ScanNetV2.}
To address this, we have conducted additional experiments on ScanNetV2~\cite{scannet-DaiCSHFN17} using different quality depth maps, such as raw sensor depth maps, monocular depth maps predicted by Metric3D~\cite{Metric3Dv2}, and depth maps from the recent SOTA method MoGe-2~\cite{moge2}.
Some qualitative results are illustrated in Fig.~\ref{fig:ds_mesh_line} for detailed comparison.
The results show that while the quality of initial depth maps does influence the final outcome, our joint optimization framework is still able to significantly improve the geometry and can generate more complete and smoother meshes compared to the initial meshes. 
In particular, the line map accuracy and surface consistency are notably enhanced through the refinement process, even when starting from noisy or sparse depth estimates.

\begin{figure*}
\centering
\includegraphics[width=\linewidth]{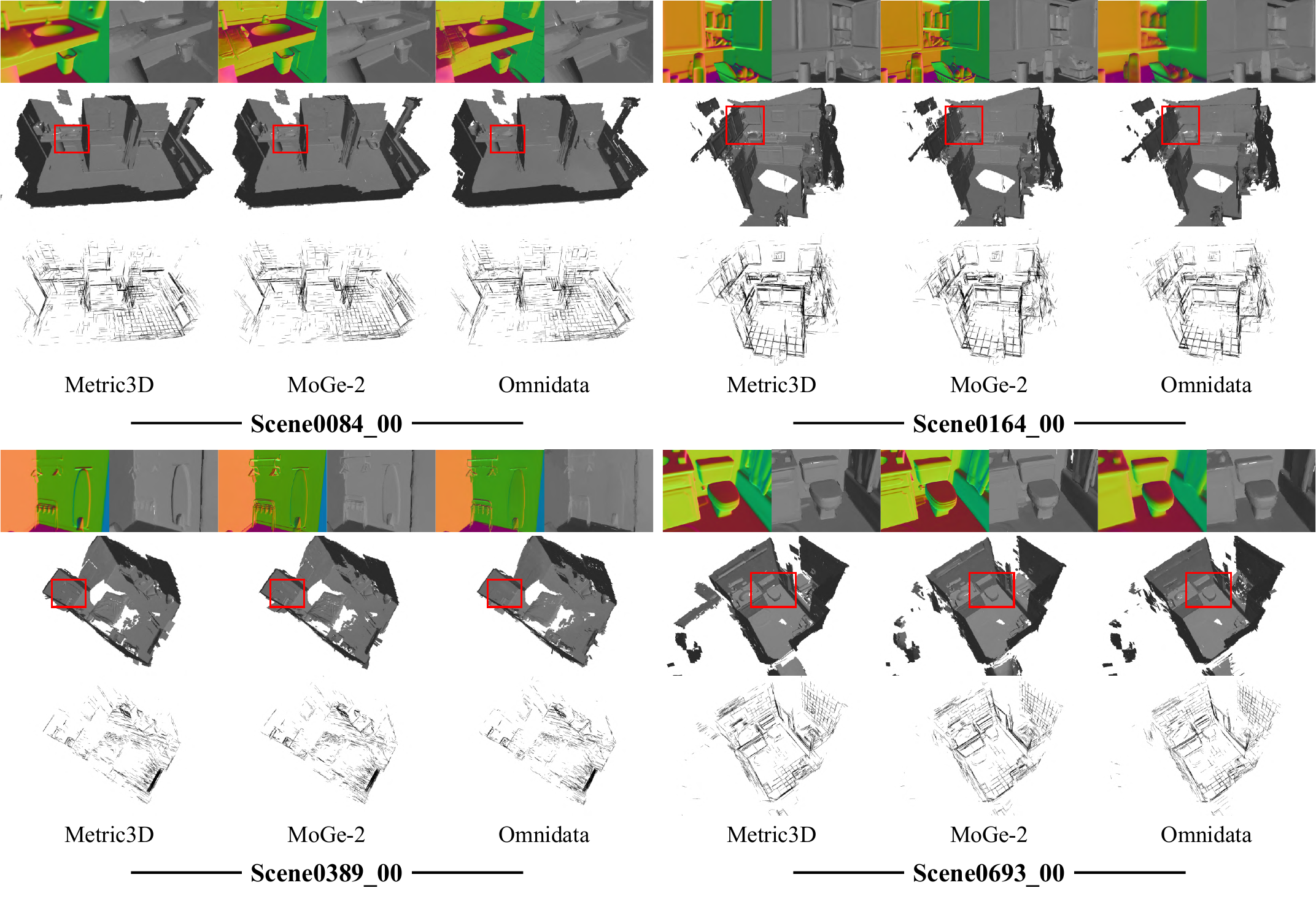}
\vspace{-3mm}
\caption{Qualitative comparison of our planar meshes and 3D line maps reconstructed with normal maps obtained from different models on ScanNetV2~\cite{scannet-DaiCSHFN17}. The raw sensor depth maps are adopted in this experiment.}
\label{fig:nm_mesh_line}   
\end{figure*}

Furthermore, we conducted experiments on ScanNetV2~\cite{scannet-DaiCSHFN17} to investigate the impact of normal maps on our method. We use sensor depth maps and perform optimization with normal maps from Metric3D~\cite{Metric3Dv2}, MoGe-2~\cite{moge2}, and Omnidata~\cite{omnidata-EftekharSMZ21}, respectively. The results are shown in Fig.~\ref{fig:nm_mesh_line}.
We per-scene illustrate the normal map of one view and the corresponding part of the reconstructed mesh as a concise comparison to show in detail the impacts of normal maps on the optimization results.
It can be seen that normal maps may have some influence on the final optimized mesh, but all settings still produce similarly high-quality 3D line maps in the end.

\begin{figure*}[!h]
\centering
\includegraphics[width=0.8\linewidth]{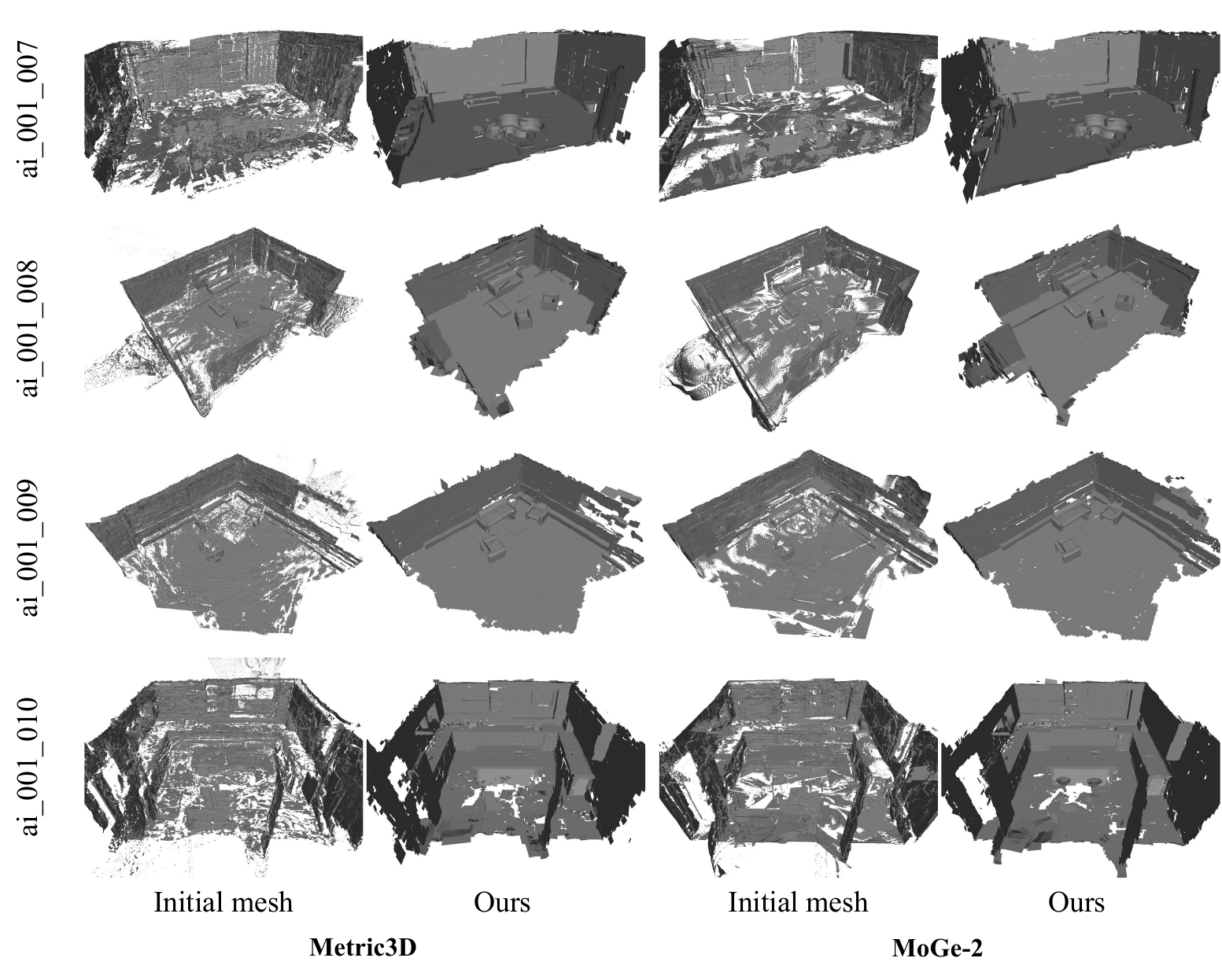}
\vspace{-3mm}
\caption{Qualitative comparison of initial meshes and our planar meshes on Hypersim~\cite{Hypersim}.}
\label{fig:hd_mesh1}   
\end{figure*}

\begin{figure*}[!h]
\centering
\includegraphics[width=0.9\linewidth]{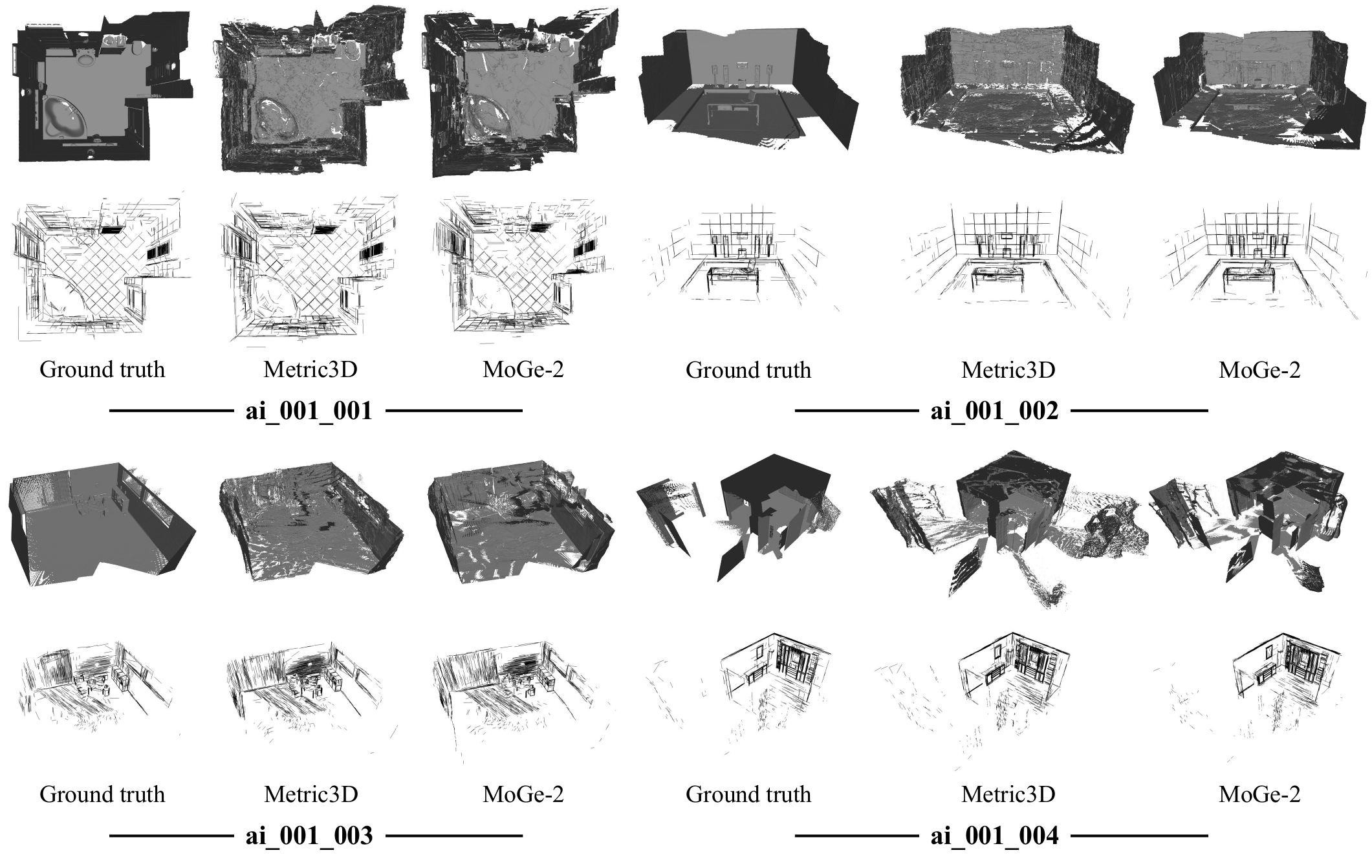}
\vspace{-3mm}
\caption{Qualitative comparison of 3D line mapping with different quality of initial meshes on Hypersim~\cite{Hypersim}.}
\label{fig:hd_line1}   
\end{figure*}

\begin{figure*}[!h]
    \centering
    \includegraphics[width=\linewidth]{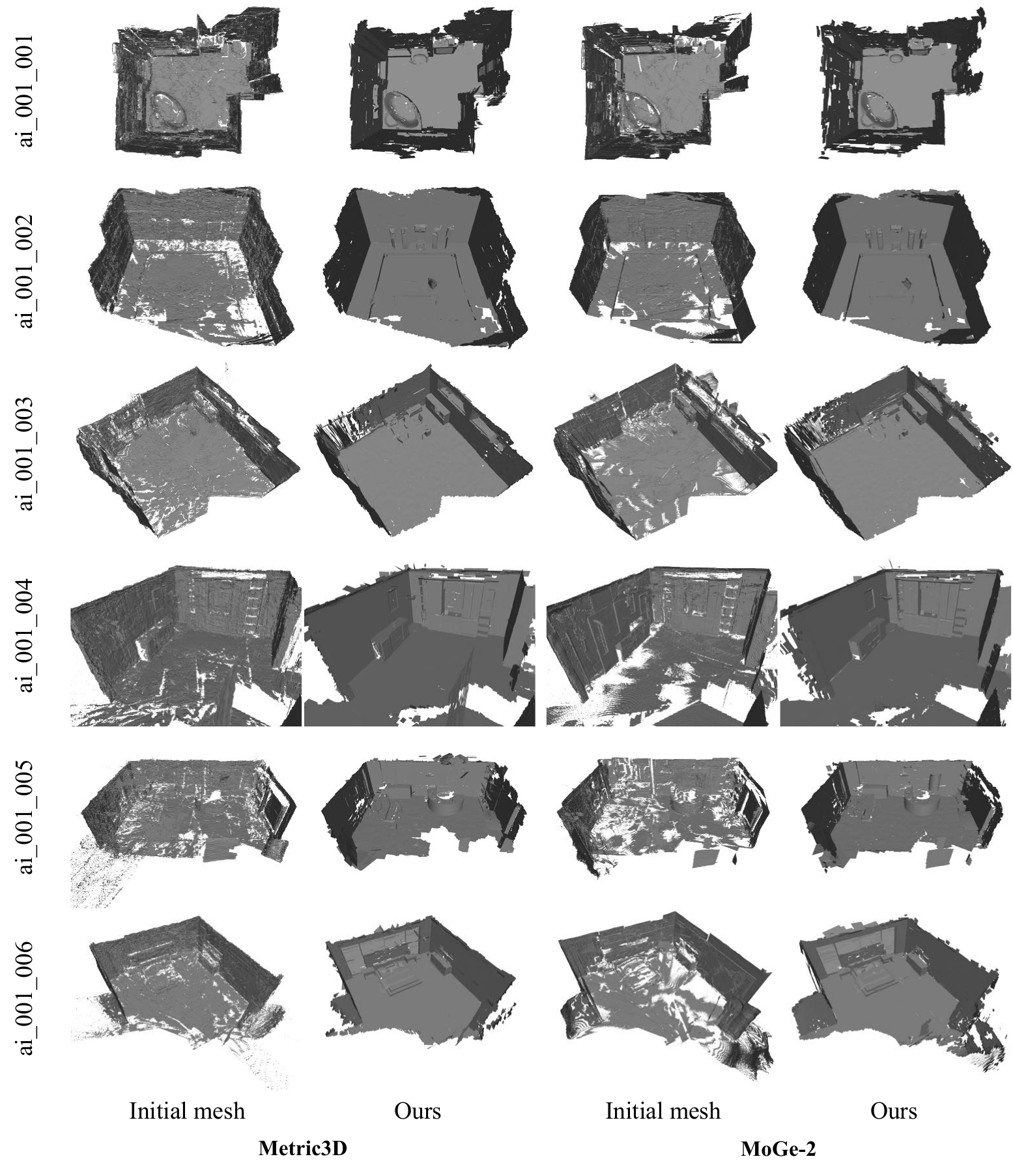}
    \vspace{-3mm}
    \caption{Additional qualitative comparison of initial meshes and our planar meshes on Hypersim~\cite{Hypersim}.}
    \label{fig:hd_mesh2}   
    \end{figure*}

\begin{figure*}[!h]
\centering
\includegraphics[width=\linewidth]{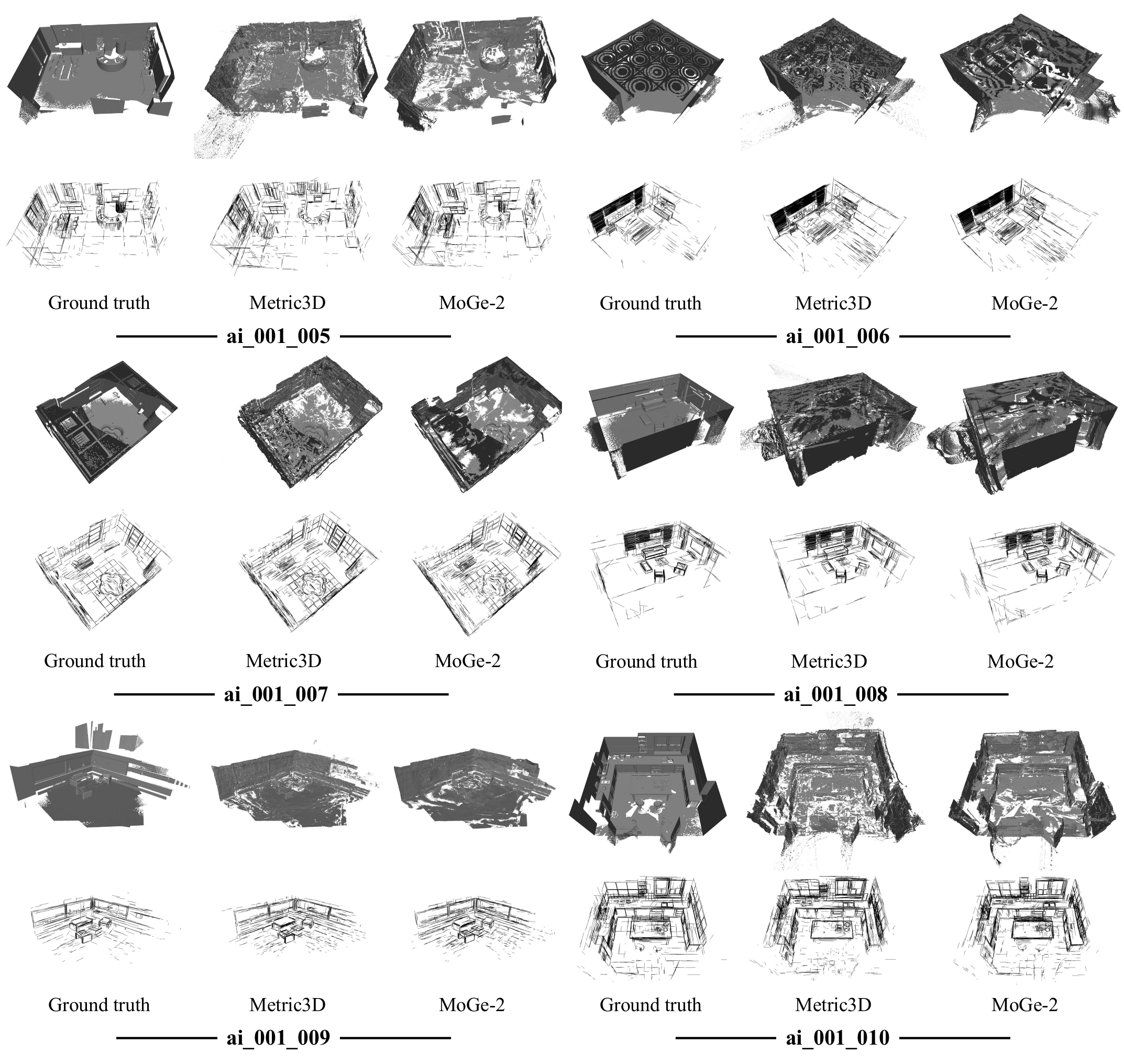}
\vspace{-3mm}
\caption{Additional qualitative comparison of 3D line mapping with different quality of initial mesh/depth on Hypersim~\cite{Hypersim}.}
\label{fig:hd_line2}   
\end{figure*}

\paragraph{Hypersim.}
Similarly, regarding the impact of depth maps on the optimization results, we conducted extensive experiments on the Hypersim dataset~\cite{Hypersim} using ground-truth depth maps, depth maps from Metric3D, and depth maps from the recent SOTA method MoGe-2, while keeping all other settings identical.
Fig.~\ref{fig:hd_mesh1} compares initial meshes from Metric3D~\cite{Metric3Dv2} and MoGe-2~\cite{moge2} against our optimized planar meshes, demonstrating that our method recovers high-quality planar meshes even from low-quality initialization.
Fig.~\ref{fig:hd_line1} shows that our optimization framework produces high-quality line maps comparable to those initialized with ground-truth meshes, even when starting from poor-quality initial meshes.
Additional results on other Hypersim scenes are shown in Fig.~\ref{fig:hd_mesh2} and Fig.~\ref{fig:hd_line2}.
    
The extensive results above provide evidence that our optimization framework is designed to be robust to certain levels of noise and sparsity in the input depth maps and normal maps, and demonstrate that the proposed optimization contributes meaningfully to the final output, beyond merely fitting the input priors.

\section{More Results about Generalization to Non-Planar Structures}
\begin{figure*}
\centering
\includegraphics[width=\linewidth]{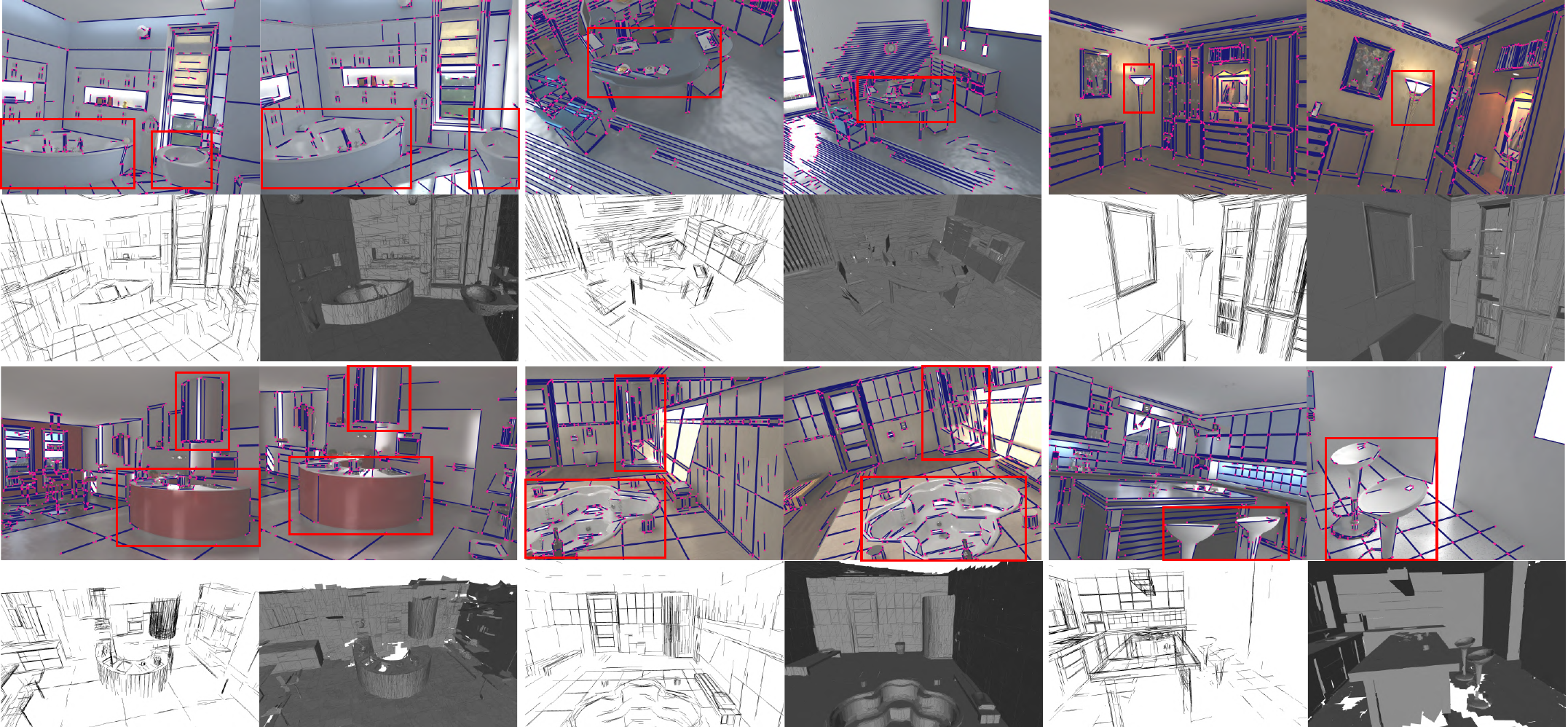}
\vspace{-3mm}
\caption{Qualitative results of non-planar structure reconstruction on Hypersim~\cite{Hypersim} scenes. We show 2D line detection, reconstructed meshes, and 3D line maps for curved structures such as chairs and circular tables.}
\label{fig:hypersim_curves}   
\end{figure*}

We have investigated the generalization ability of our method on generic non-planar structures and conducted experiments using only multi-view images from scenes in some common datasets.
In this section, we select several views from Hypersim~\cite{Hypersim} scenes that contain such non-planar geometry, and visualize 2D line detection, the meshes, and line maps reconstructed by our method in Fig.~\ref{fig:hypersim_curves}.
It can be seen that our method can approximate curved surfaces by fitting these non-planar structures with a dense set of planar primitives, which demonstrates the satisfaction of the needs for reconstructing many common non-planar structures in man-made indoor scenes.

\begin{figure*}
\centering
\includegraphics[width=\linewidth]{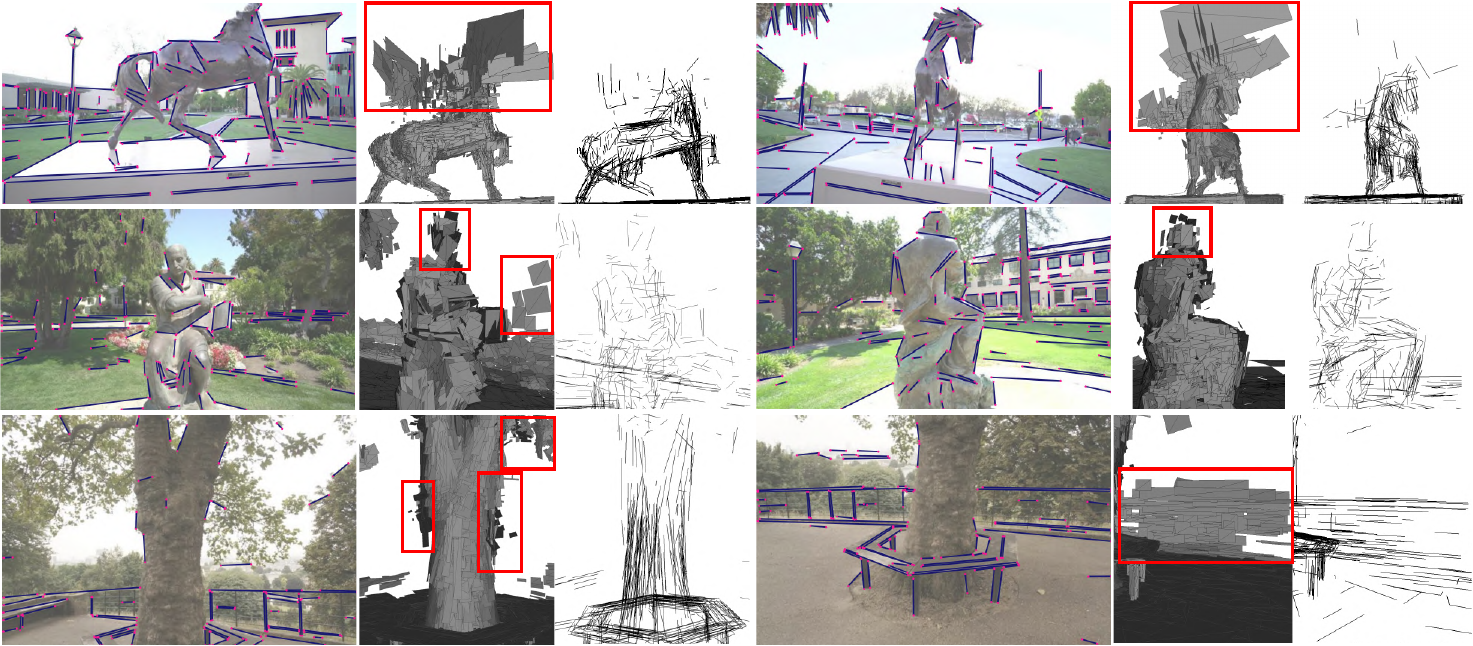}
\vspace{-3mm}
\caption{Qualitative results of locally non-planar structure. The 2D line detection results, along with the reconstructed 3D planar surfaces and 3D line segments, are illustrated for reference.}
\label{fig:non-planar-details}   
\end{figure*}

Additionally, we illustrate some local failure cases of non-planar scenes in Figure~\ref{fig:non-planar-details}.
Red boxes highlight inaccurate or hallucinated planes that often arose to satisfy the supervision from the noisy depth maps or normal maps.
However, the strict thresholds used in our assignment strategy (as described in Sec.~\ref{subsec:ltb}) enforce strong geometric consistency between 2D and 3D lines, allowing these inaccurate plane edges to be filtered out.

\section{Failure Cases}

\begin{figure*}
    \centering
    \includegraphics[width=\linewidth]{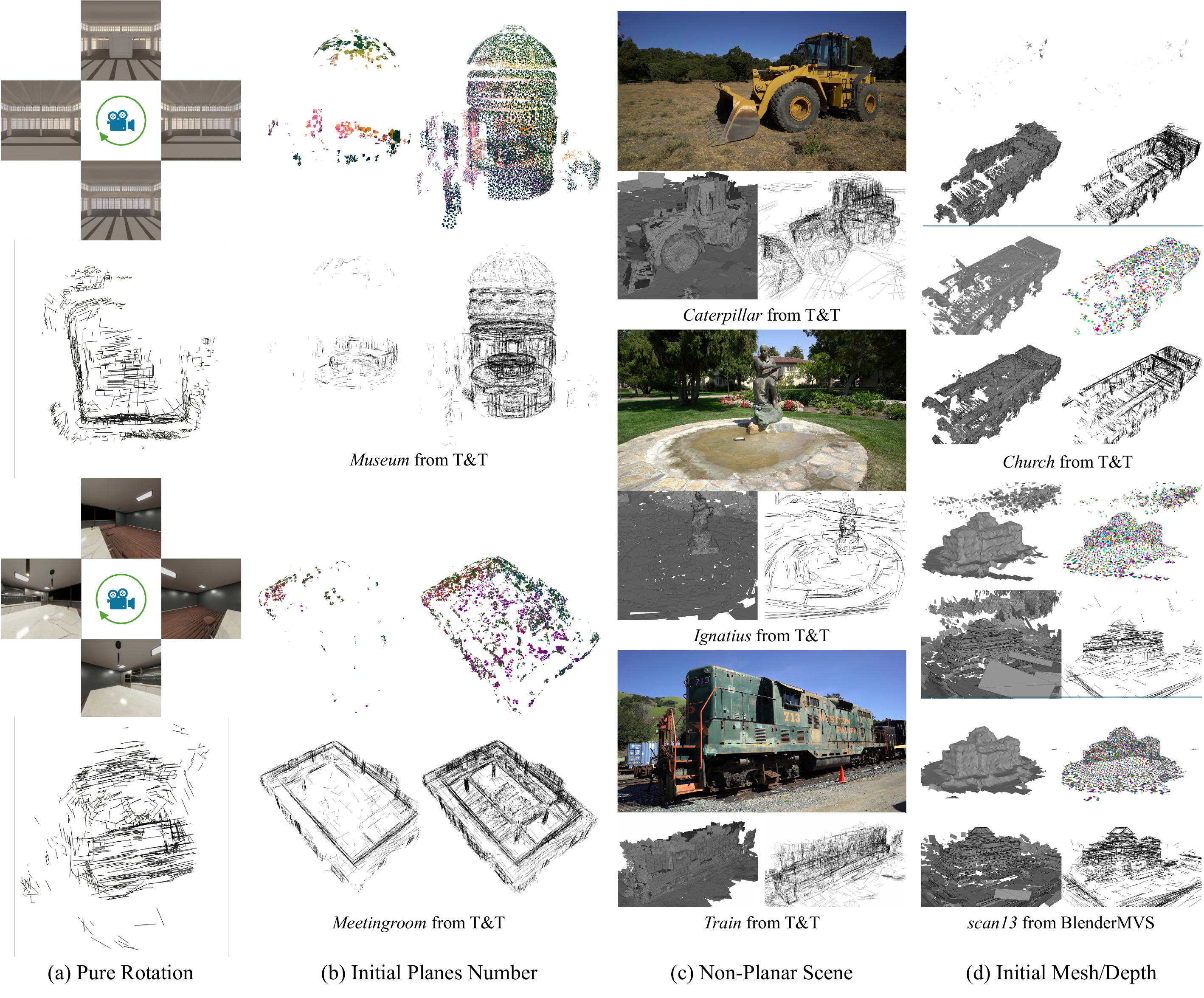}
    \vspace{-3mm}
    \caption{Visualization of our fully/partially failure cases.}
    \label{fig:failure}
\end{figure*}

 We have summarized some failure cases under four different conditions and illustrated qualitative results in Fig.~\ref{fig:failure}, including completely failed cases and partially unsatisfactory cases. 
    To explore the boundaries of our method as thoroughly as possible, all results are reconstructed with VGGT’s~\cite{wang2025vggt} imperfect outputs.
    (a) Pure rotation: This is a classic ill-posed problem. We render 100 views for each of two scenes in Blender by rotating the camera without any translation. Purely rotational sequences make outputs from VGGT unreliable, causing the initial mesh to exhibit severe geometric drift and ultimately leading to reconstruction failure.
    (b) Initial number of planes: Given the same initial mesh, initializing the planar surface with insufficient plane primitives can also lead to the reconstruction failing for large-scale scenes.
    (c) Non-Planar Scene: Reconstructing non-planar structured scenes remains challenging for our method as well, and results are less accurate and less clean than those obtained for indoor scenes.
    (d) Initial mesh/depth: Using the same number of planar primitives for initialization, noisy or inaccurate initial meshes or depth maps can also degrade the reconstruction quality.

\end{document}